# Edge-Optimized Deep Learning & Pattern Recognition Techniques for Non-Intrusive Load Monitoring of Energy Time Series

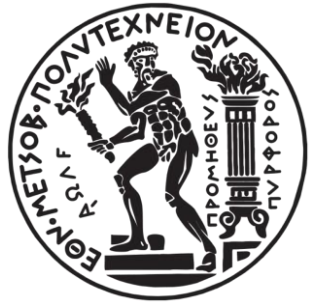

## Sotirios Athanasoulias

School of Rural, Surveying and Geoinformatics Engineering
National Technical University of Athens

This dissertation is submitted for the degree of
*Doctor of Philosophy*

May 2025

# Research Committee

*Examination Committee*

1. Nikolaos Doulamis, Professor · School of Rural, Surveying and Geoinformatics Engineering · National Technical University of Athens, Supervisor

2. Anastasios Doulamis, Professor · School of Rural, Surveying and Geoinformatics Engineering · National Technical University of Athens

3. Pavlos S. Georgilakis, Professor · School of Electrical and Computer Engineering · National Technical University of Athens

4. Emmanouel (Manos) Varvarigos, Professor · School of Electrical and Computer Engineering · National Technical University of Athens

5. Vassilis Gikas, Professor · School of Rural, Surveying and Geoinformatics Engineering · National Technical University of Athens

6. Stelios Tsafarakis, Associate Professor · School of Production Engineering and Management · Technical University of Crete

7. Eftychios Protopapadakis, Assistant Professor · School of Information Sciences · University of Macedonia

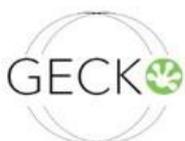
GECKO is funded by the European Commission under Horizon2020 MSCA-ITN-2020 Innovative Training Networks programme, Grant Agreement No 955422.
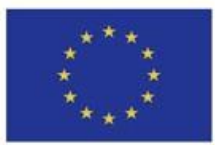
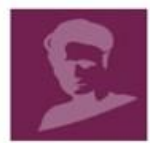
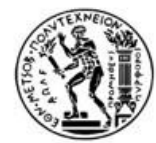

I would like to dedicate this thesis to Yoko

# Declaration

I hereby declare that except where specific reference is made to the work of others, the contents of this dissertation are original and have not been submitted in whole or in part for consideration for any other degree or qualification in this, or any other university.

<div align="right">
Sotirios Athanasoulias
May 2025
</div>

# Acknowledgements

I wish to express my deep gratitude to the many people who supported me throughout my PhD journey. First and foremost, I would like to thank my academic advisor, Professor Nikolaos Doulamis from the National Technical University of Athens, and my industrial advisor, Nikolaos Ipiotis from Plegma Labs. Professor Doulamis has been an exceptional supervisor, whose remarkable ethos and dedication have greatly influenced my development as a researcher. His support in building a research-oriented mindset and honing my analytical skills was invaluable.

I am equally grateful to Nikolaos Ipiotis, whose mentorship was pivotal in guiding me through the technical aspects of my PhD. As my industrial supervisor, he provided both a business-oriented perspective and a welcoming environment within his team. His mentorship helped me develop essential professional skills, and his constant support made the challenges of this journey far more manageable.

A special thanks to Professor Anastasios Doulamis, who introduced me to the fundamental principles of Machine Learning and Energy research. Throughout our collaboration on various research projects, he provided invaluable guidance, freedom, and encouragement, helping shape numerous aspects of this dissertation. His insights, feedback, and hands-on involvement were instrumental in my development.

My sincere appreciation also goes to Dr. Nikolaos Temenos, my postdoctoral advisor. His guidance was crucial in helping me achieve impressive results, often within challenging timelines.

I would also like to thank the esteemed members of my advisory committee, Prof. Anastasios Doulamis and Prof. Pavlos Georkilakis, for their insightful feedback and ongoing support throughout my PhD journey. Additionally, I am grateful to the distinguished members of the examination committee - Prof. Emmanouel (Manos) Varvarigos, Prof. Vassilis Gikas, Associate Prof. Stelios Tsafarakis and Assistant. Prof. Eftychios Protopapadakis - for their participation in the PhD defence process.

Finally, I extend my thanks to the entire GECKO Marie Curie ITN consortium, whose support was instrumental in enhancing my research skills, fostering meaningful collaborations, and nurturing an interdisciplinary approach to research.

# Abstract

The rapid growth in global energy demand, coupled with the pressing need for sustainability, necessitates innovative solutions to improve energy efficiency. While technological advancements have introduced numerous energy-saving systems, their standalone implementation often falls short of delivering meaningful impact. Providing users with feedback on their energy consumption behavior has proven to be a critical step toward fostering sustainable practices.

Non-Intrusive Load Monitoring (NILM) stands out as a promising approach for providing users with energy feedback by disaggregating the total household energy consumption, recorded by a central smart meter, into appliance-level insights. This empowers users with actionable information to optimize their energy use. NILM's potential has been significantly enhanced by advancements in artificial intelligence, the Internet of Things (IoT), and the extensive data generated by widespread smart meter rollouts.

Although AI-enabled NILM research is promising, its real-world deployment remains limited due to several critical challenges. The first challenge lies in the availability and representativeness of training datasets. Existing NILM datasets primarily originate from regions like the USA and UK, leaving other regions, such as the Mediterranean, underrepresented. This creates a gap in understanding unique energy consumption behaviors, including the use of appliances like air conditioners and electric water boilers, which constitute a significant share of energy consumption and present considerable potential for demand-side energy flexibility. Developing comprehensive datasets that capture these distinct patterns is crucial for training models that can generalize effectively across diverse scenarios. The second challenge pertains to the high computational demands of deep learning models and the limitations of centralized data processing. Deploying NILM systems often requires significant computational resources, typically relying on cloud infrastructure. This not only raises operational costs but also introduces privacy concerns associated with transferring household energy data to central servers. Finally, the reliance on centralized processing hinders scalability and excludes households with limited connectivity or access to cloud services, restricting the practical applicability of NILM solutions.

This thesis addresses these challenges through several key contributions. First, it introduces the development of an interoperable data collection framework that facilitates the creation of the Plegma Dataset, a high-frequency energy dataset specifically designed to capture the unique energy consumption behaviors of underrepresented regions like the Mediterranean. Second, it explores advanced deep neural network (DNN) approaches for NILM and investigates cutting-edge compression techniques to optimize these models for deployment on resource-constrained edge devices. Additionally, the thesis delves into the application of compression techniques for NILM models tailored to Mediterranean energy consumption scenarios, using the Plegma Dataset to validate their effectiveness and potential. By bridging the gap between theoretical innovation and practical application, this research aims to establish NILM as a scalable, efficient, and context-aware solution for advancing global energy efficiency.

# Contents

# List of Figures

# List of Tables

# Part I

# Introduction & Fundamentals

# Chapter 1

# Research Context & Direction

## 1.1 Introduction

The global energy demand is experiencing a significant upward trajectory, driven by population growth, rising living standards, and rapid industrialization. Projections from the International Energy Agency (IEA) indicate that global energy demand is set to double by 2050 [78]. A parallel trend can be observed in global electricity consumption, which has more than tripled between 1980 and 2023 [157]. Furthermore, the IEA forecasts that global electricity demand will grow at an accelerated rate over the next three years, averaging a 3.4% annual increase through 2026 [81].

In the European Union, the commitment to achieving net-zero emissions by 2050 has pushed the adoption of ambitious policies targeting emissions reductions, renewable energy expansion, electric vehicles (EVs), and heat pump deployment—factors that collectively drive an increase in electric power demand [121]. Additional contributors to this demand include long-term regulatory initiatives beyond 2030, such as the "Fit for 55" package, the declining costs of electrification technologies, and economic recovery following GDP contractions [43]. These dynamics underscore the interplay between policy, technology, and economic factors in shaping future energy and electricity demand trends.

Among the sectors driving this growth, the residential sector plays a pivotal role, with household electricity consumption representing a significant portion of total energy use. On a global scale, the residential sector contributes an average of roughly 25% to total energy consumption, with variations shaped by factors such as climate, economic conditions, and the implementation of energy efficiency measures. [55]. One key factor contributing to increased residential electricity demand is the growing emphasis on well-being and the expansion of household energy needs. According to the U.S. Energy Information Administration, residential electricity demand is expected to rise steadily, accounting for 32% of total final energy consumption by 2040 [169]. This underscores the crucial role of residential consumption in shaping overall energy demand patterns, further intensifying the global focus on sustainable energy solutions.

However, the rising demand for residential electricity, while improving the quality of life, imposes significant environmental and infrastructural costs. Residential electricity consumption is responsible for approximately 30% of global greenhouse gas emissions, highlighting its substantial contribution to ecosystem degradation [131]. Additionally, increased residential energy demand can place considerable stress on electricity grids, potentially resulting in large-scale power outages that disrupt entire regions and cause severe economic and public health crises [130]. In this context, addressing domestic energy consumption plays a crucial role in mitigating global energy overuse. As highlighted in [132], it can significantly contribute to achieving one of the seven stabilization wedges needed to reduce carbon emissions by 2054. Moreover, it holds substantial potential for unlocking demand-side grid flexibility, a key factor in building a resilient and efficient energy system [4].

To mitigate the adverse effects of rising energy demand, technological advancements have focused on improving the energy efficiency of appliances and systems to reduce overall consumption. However, the adoption of energy-efficient technologies, such as energy-efficient appliances, faces significant barriers due to their higher upfront costs, which often remain prohibitive for many consumers. This challenge is particularly acute for low-income households, where the initial investment in energy-efficient appliances is unfeasible despite their potential for long-term savings [73]. For instance, a study by the Kleinman Center for Energy Policy emphasizes that these high upfront costs limit access to energy-efficient technologies in low-income communities, restricting their ability to benefit from reduced energy consumption and lower utility bills [100].

Along with being accessible only to a small, affluent audience, energy-efficient technologies may also face another critical limitation: the potential to trigger the 'rebound effect.' The rebound effect expresses the extent of the energy saving produced by an efficiency investment that is taken back by consumers in the form of higher consumption, either in the form of more hours of use or a higher quality of energy service [72]. For instance, when we replace a 75 W incandescent bulb with an 18 W compact fluorescent bulb (CFL)—a reduction in power of about 75%—we could expect over time a 75% energy saving. However, this rarely happens. Many consumers, realizing that the light now costs less to run, are less concerned about switching it off, indeed they may leave it on all night, for example for increased safety or security resulting in consuming the same or even more energy [72]. This and similar examples presented in [147, 76] show that technological improvements in energy efficiency are often absorbed into increasing the standards of living thereby reducing their potential to achieve meaningful energy conservation.

According to numerous research studies, a crucial factor that can have a real impact in driving domestic energy efficiency is the consumer energy behavior, often surpassing the impact of buildings and energy-efficient technologies themselves [115]. Research in this field has largely concentrated on feedback mechanisms as a key strategy for fostering more sustainable energy habits. In the context of energy efficiency and consumer behavior, feedback involves providing individuals with detailed information about their energy consumption, helping to "rematerialize" energy use—making it more tangible and understandable [115] while simultaneously raising awareness and motivating individuals to adopt more efficient energy behaviors [59, 67]. Several field studies have demonstrated that feedback on energy consumption, particularly when it includes real-time and historical data, can lead to savings of up to 20% [93], with even greater impact in cases where feedback is enhanced by providing information on the consumption of individual appliances [31].

One approach to providing this feedback to consumers is through the installation of sub-metering devices on existing appliances or the deployment of smart appliances capable of transmitting their energy consumption data to a central gateway. However, both solutions face significant barriers to large-scale adoption due to their intrusive nature, which increases installation complexity and the cost associated with purchasing new equipment. Additionally, the development of interoperable systems that can seamlessly integrate diverse sensors and devices within a unified IoT framework presents another significant challenge, further limiting the widespread implementation of these solutions [11]. An alternative to address these challenges is Non-Intrusive Load Monitoring (NILM), also known as energy disaggregation, a method first introduced by Hart in the early 1980s [68]. NILM operates by analyzing the total household energy consumption recorded by a central meter to estimate the energy usage of individual appliances. This approach eliminates the need for submetering installations, relying exclusively on data collected from a single central meter.

With the advancement of artificial intelligence (AI) and the widespread deployment of smart meters in many countries worldwide [9], NILM has gained renewed interest in both research and industry, driving various academic and commercial efforts [90]. These developments enable NILM to achieve highly accurate energy disaggregation performance by leveraging the vast amounts of data generated by smart meters, enhancing its potential for precise appliance-level energy consumption analysis. This thesis will focus on identifying and addressing key challenges associated with AI-enabled NILM, particularly the context-aware nature of the problem, the need for extensive training data, and the practical difficulties of deploying these solutions due to their resource-intensive requirements.

## 1.2 Challenges and Motivation

The rapid advancement of artificial intelligence has transformed sectors such as manufacturing, finance, transportation, and healthcare by providing innovative solutions to complex challenges [141]. AI systems mimic human cognitive functions like learning, reasoning, and decision-making by processing large datasets, identifying patterns, and generating insights [107]. These advancements have significantly improved efficiency, accuracy, and automation. However, AI performance heavily depends on the quality, diversity, and volume of training data. Poor or biased datasets can impair model accuracy, reduce reliability, and limit generalizability, highlighting the critical need for robust data collection and preprocessing [197]. The extensive data generated by smart meters, combined with the advanced capabilities of AI methods, has significantly enhanced the potential for applying AI-enabled solutions to the energy sector, particularly in addressing the challenges of NILM. While numerous studies have demonstrated the effectiveness of deep neural network approaches for NILM, which significantly outperform traditional signal processing techniques, these methods have yet to achieve widespread adoption in households, despite growing interest from the industry.[12].

A key obstacle that hinders the adoption of AI-enabled NILM is its context-aware nature. NILM performance is influenced by factors such as the energy behavior of occupants, the specific appliances they use, as well as climatic and sociodemographic characteristics. So the real challenge lies in the creation of comprehensive datasets that capture these diverse patterns to enable the training of NILM models capable of effective generalization. While several open-access datasets exist, particularly from regions like the USA and UK, there has been limited focus on regions such as the Mediterranean. This region exhibits distinct consumption behaviors, often involving appliances like air conditioners and electric water boilers, which are not commonly represented in existing datasets and hold significant potential for energy flexibility. Furthermore, the process of data collection for NILM datasets presents additional complexities, as it requires the development of interoperable and efficient frameworks to gather aggregate and sub-metered data from participants' households, addressing data gaps and ensuring the reliability needed for DNN-NILM model training.

Another limitation of AI-enabled NILM approaches is their high demand for computational power and resources, necessitating central servers or cloud computing infrastructures [12], which drive up both operational costs and energy consumption. Additionally, the reliance on centralized data processing necessitates the data transferring of the recorded consumption data from the originating household to a central server, with the inferred disaggregated results then being transmitted back to the users. This not only raises data privacy concerns but also increases data storage expenses, further impeding the practical adoption of these solutions in real-world scenarios [160]. In an effort to alleviate these constraints, the NILM research community has increasingly focused on compression techniques to reduce the computational complexity of DNN-based NILM models, enabling their deployment on resource-constrained devices at the consumer's premises [12, 15]. This approach eliminates the need for centralized infrastructure and data transferring, as data processing and inference occur locally on the consumer's side [160].

The motivation of this thesis is to address the aforementioned challenges by advancing research that bridges the gap between theoretical advancements and practical implementation of edge-based NILM solutions. By focusing on enabling real-world adoption, this research aims to overcome barriers related to data availability, computational resource constraints, and privacy concerns. The ultimate goal is to develop scalable, efficient, and context-aware NILM methods that can operate seamlessly on resource-constrained devices, thereby facilitating widespread deployment and contributing to energy efficiency and sustainability efforts.

## 1.3   Research Questions

The development and deployment of NILM solutions face several critical challenges, as outlined in the preceding sections. Addressing these challenges is essential for enabling the practical adoption of NILM systems that can contribute to energy efficiency and sustainability efforts on a global scale. This section presents the key research questions that this thesis seeks to address. These questions are formulated to bridge gaps in existing research, particularly in the context of dataset limitations, computational constraints, and region-specific applicability. By tackling these issues, the aim is to advance the field of AI-enabled NILM and support its integration into real-world scenarios. The research questions addressed in this thesis are the following:

- **How can the challenge of NILM datasets be addressed by developing an interoperable data collection framework, and how can this framework be applied to bridge the gap in data availability for regions such as the Mediterranean?**
  This research question focuses on designing a robust and efficient data collection framework that captures diverse energy consumption behaviors. The goal is to ensure the inclusion of underrepresented regions, such as the Mediterranean, where distinct consumption patterns and appliance usage behaviors exist.

- **How can DNN-based NILM approaches and compression techniques be utilized to improve the computational efficiency and real-world applicability of NILM solutions?**

This question investigates the potential of leveraging deep neural networks and advanced compression strategies to enhance NILM model performance. It emphasizes optimizing computational requirements to enable deployment on resource-constrained devices.

- **How can compression techniques be applied to NILM models, and what insights can be gained from testing these techniques on Mediterranean-specific energy consumption scenarios?**
  This question examines the application of model compression methods to NILM solutions, with a focus on evaluating their effectiveness in addressing the unique energy consumption characteristics of the Mediterranean region. The aim is to ensure scalability and efficiency while maintaining accuracy in disaggregation tasks.

These research questions aim to address key gaps in the current state of NILM research, paving the way for practical, scalable, and region-specific solutions.

## 1.4  Originality and Contributions

The originality of this thesis lies in its interdisciplinary approach to addressing the critical challenges associated with AI-enabled NILM solutions. By integrating advancements in artificial intelligence, behavioral science for promoting energy-efficient practices, and IoT systems, this research seeks to bridge the gap between theoretical innovations and their practical implementation. The creation of interoperable data collection frameworks further enhances the potential for widespread adoption by ensuring data diversity, reliability, and accessibility. The focus on underrepresented regions, such as the Mediterranean area, adds a novel dimension by exploring unique energy consumption patterns and their implications for NILM development.

This thesis also emphasizes the development of edge-based NILM approaches, which aim to overcome the limitations of centralized data processing by enabling real-time, decentralized energy disaggregation. By developing state-of-the-art compression techniques, the research addresses the computational and resource constraints that hinder the scalability of DNN-based NILM models. The key contributions of this research are summarized as follows:

- Development of an interoperable data collection framework [11] to address dataset limitations in NILM, with a particular emphasis on underrepresented regions, especially the Mediterranean. This includes the creation of the Plegma Dataset[9], a comprehensive high-frequency dataset suitable for NILM and related applications.

- Design and evaluation of advanced DNN-based NILM approaches, including the development of a novel sequence-to-sequence (seq2seq) Graph Neural Network (GNN)-based NILM model [14], which utilizes a GNN as the encoder and a Transformer-based decoder.

- Introduction of an innovative pruning strategy to identify sub-optimal NILM neural networks prior-to-full training, significantly reducing computational costs during both the training and testing phases while improving disaggregation performance compared to conventional post-training pruning methods [12, 15].

- Proposal of a novel metric to determine the optimal pruning threshold, balancing model performance and computational efficiency, unlike traditional approaches that apply compression arbitrarily [12].

- Implementation of model compression strategies for NILM, leveraging insights from Mediterranean-specific energy consumption scenarios, derived from the Plegma Dataset, to develop context-aware and scalable solutions.

In conclusion, this thesis advances AI-enabled NILM by addressing key challenges in scalability, efficiency, and real-world applicability. By integrating edge-based techniques, advanced DNN models, and Mediterranean-specific energy insights, it offers practical, context-aware solutions that enhance performance while promoting energy efficiency and adoption in underrepresented regions.

## 1.5 Structure of the Dissertation

The remaining chapters of this thesis are organized into three distinct parts, each corresponding to one of the key research questions outlined earlier. These parts collectively address the challenges and objectives central to advancing NILM research and its real-world applicability.
**Part II: Interoperable Data Collection Approach to Address NILM Dataset Gaps**
This part focuses on answering the first research question: *How can the challenge of NILM datasets be addressed by developing an interoperable data collection framework, and how can this framework be applied to bridge the gap in data availability for regions such as the Mediterranean?*
Chapters in this part:

- **Chapter 2** proposes an Internet of Things (IoT) based framework for energy monitoring and analysis at a household level. The design of the system has taken into consideration the needs of smart-home users and has been adapted to focus on robust and real-time data collection. Our developed solution is based on Z-wave communication protocol, utilizing some of its key benefits such as interoperability and cost-effectiveness. The main goal of the proposed framework is to provide an end-to-end solution that would give accurate energy feedback through an intuitive graphical user interface. The proposed pipeline could also be used for the development of real-life living labs for high-granularity data collection as well as for the deployment and evaluation of Home Energy Management Systems (HEMS) and applications.

- **Chapter 3** presents the Plegma dataset, a high-frequency electricity measurements dataset addressing the challenges of real-world data collection for NILM and related applications. Collected over one year from 13 households in Greece, it includes whole-house loads, appliance-level consumption at 10-second intervals, and environmental data such as humidity and temperature. It also provides metadata on building characteristics, demographics, and user practices for comprehensive analysis. As the first dataset of its kind from Greece, Plegma captures Mediterranean-specific consumption patterns, including underrepresented devices like air conditioners and electric water boilers. With 218 million readings from 88 sensors, it is a critical resource for advancing NILM and Home Energy Management Systems (HEMS).

**Part III: Development and Optimization of DNN-Based NILM Architectures for Edge Deployment**

This part addresses the second research question: *How can DNN-based NILM approaches and compression techniques be utilized to improve the computational efficiency and real-world applicability of NILM solutions?*

Chapters in this part:

- **Chapter 4** introduces a novel seq2seq approach for Non-Intrusive Load Monitoring (NILM), utilizing Graph Neural Networks (GNNs) as an encoder and a Transformer-based decoder. This is the first known attempt to model NILM as a graph problem. Specifically, the proposed method leverages Graph Convolutional Networks (GCNs) to encode the aggregated signal into graph nodes based on the entropy of each appliance's signal, capturing its distinct operational states. This encoding integrates time-invariant dependencies within the aggregated signal, offering a holistic representation of the data. Experimental results demonstrate that this approach yields competitive or superior performance on multi-state appliances like washing machines compared to state-of-the-art methods. While its effectiveness is currently limited to multi-operational devices, this work introduces a promising graph-based perspective to NILM, paving the way for future research using GNNs in this domain.

- **Chapter 5** presents a novel pre-training model compression strategy for Deep Neural Networks (DNNs) in Non-Intrusive Load Monitoring (NILM). The proposed approach employs iterative magnitude pruning using $L_1$ norms to identify an optimized, compressed DNN that balances computational complexity with performance. This method reduces computational costs by up to 95% during training, achieving a significantly smaller model with negligible performance degradation compared to existing post-training compression techniques. Experimental results on the UK-DALE dataset highlight the effectiveness of this strategy, isolating sub-networks with only 5% of the original model's parameters while maintaining performance comparable to the full model. This compression approach is particularly suited for edge IoT deployment, enabling real-world NILM applications with reduced resource requirements.

- **Chapter 6** introduces OPT-NILM, a novel pruning strategy designed to optimize Non-Intrusive Load Monitoring (NILM) neural networks for both the training and deployment phases. Unlike conventional post-training compression methods, OPT-NILM identifies sub-optimal network structures before full training, significantly reducing computational costs while maintaining or improving disaggregation performance. The approach employs a metric that balances model performance with computational cost, enabling a systematic determination of pruning thresholds rather than relying on arbitrary compression. Experimental results on the UK-Dale dataset demonstrate that OPT-NILM can reduce trainable parameters by up to 95% with minimal performance degradation. To showcase real-world applicability, this chapter also presents a deployment scenario using a Raspberry Pi, illustrating the feasibility of running NILM applications on resource-constrained edge devices. By making NILM accessible for low-resource hardware, OPT-NILM enhances user adoption and opens new opportunities in the energy market.

**Part IV: Advanced Compression Techniques for NILM Architectures Evaluated in Mediterranean Scenarios**

This part addresses the third research question: *How can compression techniques be applied to NILM models, and what insights can be gained from testing these techniques on Mediterranean-specific energy consumption scenarios?*

Chapters in this part:

- **Chapter 7** presents an optimized structured pruning methodology for Non-Intrusive Load Monitoring (NILM), aimed at improving the efficiency and practicality of deploy-ing NILM models on edge devices. The methodology combines unstructured pruning to determine optimal sparsity ratios for each neural network layer with structured pruning to remove entire units based on these sparsity values. This dual-pruning approach ensures that critical feature information is preserved, avoiding arbitrary pruning thresholds and improving classification performance. Experimental results on the Plegma dataset—one of the first datasets from the Mediterranean region capturing local consumption patterns and devices—highlight the effectiveness of this approach. The methodology reduces the baseline model's MFLOPs by up to 48.85% while maintaining satisfactory disaggregation performance, outperforming traditional unstructured pruning, which does not reduce FLOPs. This structured pruning strategy underscores the potential for edge NILM applications in the Mediterranean, promoting flexibility, energy efficiency, and real-world adoption of NILM solutions.

- **Chapter 8** introduces a novel Dependency Graph (DG) structural pruning methodology to enhance the efficiency of deep learning models for Non-Intrusive Load Monitoring (NILM) in edge-enabled applications. Unlike conventional structured pruning, DG pruning captures inter-layer dependencies and prunes related parameters jointly, preserving model integrity and enabling more aggressive compression. Applied to the Plegma dataset, which reflects Mediterranean appliance usage patterns, this approach achieves up to 90% model size reduction and a $10\times$ improvement in computational efficiency with minimal impact on performance. These results highlight the potential of DG pruning to support scalable and energy-efficient NILM solutions tailored to the resource constraints of edge devices in real-world settings.

**Chapter 9** provides a comprehensive summary of the key findings and contributions of this dissertation. It discusses the research outcomes and accomplishments achieved during the course of this work, highlighting their impact on advancing the field of NILM. Finally, it outlines promising avenues for future research, emphasizing areas where further exploration could enhance the applicability and effectiveness of NILM technologies.

# Part II

# Interoperable Data Collection Approach to Address NILM Dataset Gaps

# Chapter 2

# An Interoperable and Cost-Effective IoT Sensing Architecture for Household Energy Monitoring and Analysis

## 2.1 Introduction

The continuous and increasing need for energy in relation to limited resources has led the research towards different ways to optimize our energy usage. Given that domestic energy consumption constitutes a substantial percentage of the total energy consumed [28], indicatively in the EU households represent almost the 30% of final energy consumption, there has been a lot of research work carried out in developing smart home solutions capable of helping users to moderate their electricity, gas, and water consumption in order to reduce costs and expenses. Energy monitoring and Home Energy Management Systems (HEMS) are effective ways to reduce energy consumption in households by providing real-time information on energy usage and offering automated solutions for energy conservation[65]. Using this technology, homeowners can identify areas where an excessive amount of energy is used and take action to reduce its usage.

As stated in [198], there are currently available HEMS in the market that offer various benefits and are designed for specific tasks. Different techniques have been utilized, such as wireless communication methods like radio frequency (RF), Bluetooth, and WIFI connectivity, as well as smart meters. These techniques have also been combined in various ways, as explained in [168]. Studies have shown that households using HEMS and energy monitoring systems can save up to 20% on their electricity consumption and bills by optimizing their usage. [134].

The use of the energy monitoring systems can be two-folded, as except from providing direct feedback and optimizing the energy consumption of the household users, they could also be utilized to collect energy data for the development and training of machine learning and deep learning-based applications that can also contribute a lot into the conservation of energy. These applications include Non-Intrusive-Load-Monitoring (NILM), Demand Forecasting, and Demand Response (DR) schemes, which have been proven to be very effective in energy-saving practices. Deep neural network (DNN) based NILM approaches such as [189],[14],[191], [89] DNN-based demand forecasting such as [35],[3], [98], and reinforcement learning demand response schemes [180], [173] which are considered the-state-of-the-art in this domain, need excessive amounts of training data that could be collected using such HEMS or energy monitoring systems and used to develop robust solutions the will be used to further improve the energy efficiency behavior of the households. Finally, these systems can also be used as a testbed for the deployment of edge-based solutions such as [160],[105], [196] that can run on a limited resource device of the corresponding energy monitoring system. However, according to [24], existing HEMS and energy monitoring systems have not been as successful as expected to be. The basic reasons for it are related to the high cost and the difficulties of installing these solutions in buildings that already exist. Thus, it is important to investigate new designs to develop energy monitoring systems that are cheap, easy to install, and versatile for different scenarios.

In this paper, we propose an IoT-based framework that provides real-time energy monitoring and feedback through an intuitive user interface. Our solution consists of smart meters and a Raspberry Pi, which is used as a gateway under the z-wave communication protocol. Our developed framework is specialized to provide energy feedback for the aggregate as well as the appliance level energy consumption and indoor environmental conditions. Furthermore, it could also be used to collect high-granularity energy data that are useful for historical analysis and NILM or DR applications. The basic contributions of our proposed framework are summarized below:

- **Proposing a cost-effective and easy-to-install IoT-based framework.** The proposed energy monitoring system consists of a smart meter for aggregate consumption, smart switches for appliance-level consumption, an environmental sensor, and a Raspberry Pi. The components of our system are much less expensive than similar solutions, and it is designed to be easily installed with no special wiring or technical expertise required.
- **Proposing an interoperable IoT-based system.** Our provided solution operates under the Z-Wave communication protocol, which encompasses more than 4,000 interoperable products, making our solution adaptive to various application scenarios.

- **Proposing a robust IoT-based system that does not require a stable internet connection.** The developed framework collects data through a Z-Wave communication protocol and uses a Raspberry Pi as a gateway to process and locally back up the data. This eliminates data gaps due to possible network instability and provides robust and accurate energy feedback.

The rest of the paper is organized as follows. Background on similar energy monitoring systems and the technologies used is presented in section 2. Section 3 describes the proposed framework, its basic technical components, developed software, and application. Section 4 discuss some limitations of the existing system and proposes future improvements.

## 2.2 Related Work

In this section, we provide an overview of the existing platforms as well as the basic wireless technologies used in the context of Connected Home Technology (CHT). The provided literature review helped us to identify drawbacks to the existing systems, decide on the technological aspect of our developed solution and fill the gap in the existing research area.

### 2.2.1 Existing Platforms and Systems

The current landscape of software platforms for IoT is vast, and many major vendors have already launched their own CHT platforms, such as for example Watson (IBM), HANA (SAP), Jasper (Cisco), AWS IoT (Amazon), Azure IoT (Microsoft), HomeKit (Apple), Brillo (Google), IoTivity (Intel) or AllJoyn (Qualcomm) [202]. Although all of the aforementioned systems meet the standards of the industry and provide robust and secure services, they offer their own unique IoT ecosystem, which can be difficult to connect with other platforms or integrate with unsupported devices. Thus, these systems have limited interoperability and customizability, and as a result, they cannot provide solutions that can be adapted to different use cases and scenarios.

A solution to this limitation is provided by open-source software CHT platforms which give more flexible and interoperable options to the users. Some of the most common open-source software are listed in the table below. This table provides information about the wireless technology used as well as some of their key features and characteristics. Domoticz is a lightweight and easy-to-use home automation platform that allows users to monitor and control their home devices using a web-based interface and uses a variety of wireless communication protocols. Kaa is an enterprise IoT platform that operates under different wireless protocols and is ideal for non-technical users. OpenHab is another open-source

CHT platform with the ability to integrate a multitude of devices and systems. Some of the basic contributions of this platform include the uniform and intuitive user interface as well as the large and active OpenHab community. Finally, HomeAssitant is another open-source platform that gives the ability to create custom scripts and supports many different protocols, such as WiFi, Z-Wave, and Zigbee.

| Open-source platforms | Common Technology Used | Key features |
| --- | --- | --- |
| Domoticz | Z-wave<br>ZigBee | - Push notification<br>- Definition of IFFTTT rules<br>- Extend functionality using custom scripts<br>- Lightweight |
| Kaa | Z-wave<br>ZigBee<br>Bluetooth | - SDK compatibility that allows the extension of the platform<br>- Data schema definition<br>- Freedom of deployment |
| OpenHab | ZigBee<br>Z-wave<br>Bluetooth<br>KNX | - Large community with numerous extensions<br>- Voice control<br>- Safety<br>- User friendly interface |
| HomeAssistant | WiFi<br>Z-wave<br>ZigBee | - Custom scripts compatibility<br>- Integration with virtual assistants<br>- Multiple languages<br>- High speed |

Table 2.1 Overview of the existing CHT open-source systems

All these software support wireless communication and provide an end-to-end (built-in) solution for the smart home ecosystem, including gateway software, client application, and prebuilt package. Although these built-in solutions are ideal for prototyping and building a system without great effort, they are not customizable in terms of their architecture and parameterization, limiting their applicability to a variety of different use cases. For this reason, this paper aims to present a custom architecture that will overcome the aforementioned shortcomings and provide an interoperable and flexible solution.

## 2.2.2 Wireless Technologies Used For Connected Home Technology on IoT

As previously discussed, various wireless technologies are employed in the domain of connected home technology and monitoring applications. The selection of a communication protocol represents the initial stage in developing such an application. This section presents a review of prominent wireless technologies, ultimately leading to our choice to utilize the Z-wave communication protocol. Commonly utilized wireless technologies in home automation systems comprise WiFi, Bluetooth Zigbee, and Z-Wave, as described in [144]. WiFi is a wireless technology that is used to CHT based on the IEEE 802.11 standards [46]. It is available worldwide since it operates both in 2.4GHz and 5GHz bands [77]. In CHT, WiFi is mostly used for wireless monitoring and energy management of home appliances. Its main disadvantage is that it cannot operate without an internet connection, which can result in many data gaps and inaccurate energy feedback. Bluetooth is a short-range wireless technology based on the IEEE 802.15.1 standard that can provide a low-budget home automation solution [46]. One of the main disadvantages of this wireless protocol is its communication range which does not exceed 10 meters. However, it is a very common CHT technology due to its low cost and compatibility with many devices [151]. ZigBee utilizes three radio bands - 868MHz, 915MHz, and 2.4GHz - to ensure low power consumption and a 100-meter range [46]. While it boasts the ability to connect up to 6000 devices and has a relatively low-cost installation, its compatibility with devices from various manufacturers remains limited [47] Z-Wave is another wireless communication protocol that was developed primarily for use in CHT. This protocol uses very low power and communicates in the frequency of 900MHz and a range of 30 meters [46]. Z-Wave also uses a mesh network topology, which allows devices to communicate with each other through a series of nodes. This means that each Z-Wave device in the network can act as a repeater, increasing the range and coverage of the network [186]. This also improves the reliability and stability of the network, as devices can find alternative paths to communicate if one path is blocked or unavailable [111].

| **Indices** | **Zigbee** | **Z-Wave** | **WiFi** | **Bluetooth** |
|---|---|---|---|---|
| Power Consumption | 100mw | 1mw | high | 10mw |
| Range | 100m | 100m | 1000m | 10m |
| Scalability | high | high | medium | low |
| Interoperability | same producer | different manufacturer | WiFi compatible devices | Bluetooth compatible devices |

Table 2.2 Comparison of Wireless CHT technologies

An overview of the wireless technologies analyzed above can be shown in Table 2.2, which presents a comparison between their most important features as presented in [46]. According to the literature review presented above, the most suitable wireless protocol for the purpose of our developed interoperable house energy monitoring framework is the Z-wave. In summary, Z-wave is an excellent choice for this kind of application due to its low consumption, mesh network topology, and scalability opportunities. According to [111], although Z-wave protocol is considered the most suitable technology for CHT applications, there have not been many studies about Z-Wave and their integration into smart home systems, which was another motivation for our selection.

## 2.3 Proposed Framework

This section describes the developed energy monitoring framework. Firstly, we present the proposed system requirements that guided the decision on the architecture of our solution. Afterward, we present the overview of the proposed architecture and give information about the technical equipment used to build our solution. Subsequently, we present the implemented gateway and its software stack, which is considered as the brain of our CHT system. Finally, we present the graphical user interface and the basic capabilities that give to its users. The primary objective of the developed framework is twofold: (a) to provide energy feedback to users through an intuitive user interface and (b) to collect high-granularity aggregate and appliance-level consumption data for the creation of a Non-Intrusive Load Monitoring (NILM) dataset. As such, the system is designed to offer a valuable resource for users seeking to optimize their energy consumption while also providing researchers with a comprehensive dataset for energy analysis. To achieve this, the system focuses on creating an interoperable and robust framework that eliminates data gaps and provides accurate energy feedback.

### 2.3.1 Proposed system requirements

In order to finalize the technical setup and architecture of the developed CHT framework, we specified both the functional and non-functional requirements. Functional requirements describe the specific features and functions that a system must have to satisfy its users' needs. Non-functional requirements specify the qualities that the system must possess [44].

The functional requirements of our developed system are:

- The system should be able to collect high-granularity aggregate and appliance-level energy consumption data to the gateway and send it to the cloud.

- The system should eliminate the data gaps by providing a local backup system on the gateway.

- The cloud should parse the collected data and store them in a central database.

- Users should be able to view the collected data through an intuitive application.

    Users should be able to interact with the application and create their own graphs and visualizations.

The non-functional requirements of the system indicate that emphasis is placed on interoperability, cost-effective scalability, and privacy. This highlights the system's ability to integrate with other systems, scale efficiently, and maintain user confidentiality.

*A. Interoperability*

The system should give the opportunity to devices from different manufacturers to communicate with each other regardless of their brand, providing seamless interoperability and flexibility for the users.

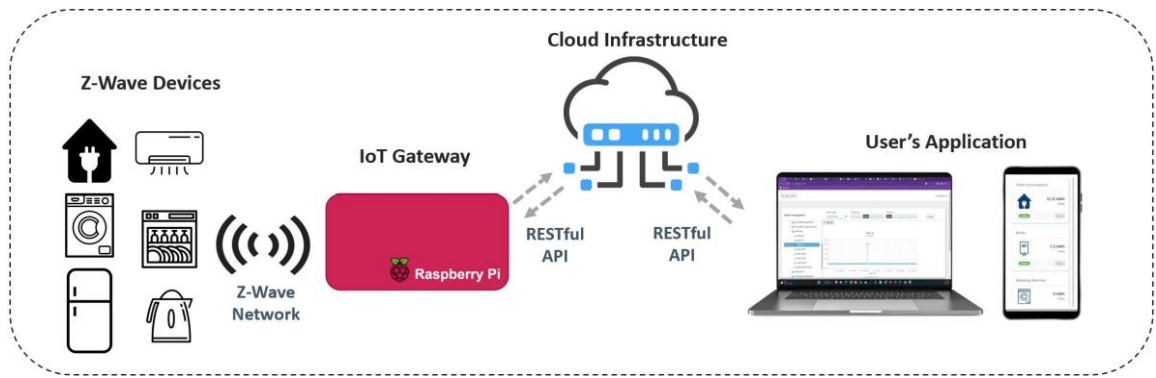

Figure 2.1 Overview of the proposed IoT architecture.

*B. Scalability - Cost*

The developed solution should be a scalable and cost-effective CHT monitoring system that could efficiently collect and provide energy feedback to different types of household users. The solution should be cost-effective in order to be accessible to a wide variety of users.

*C. Privacy*

The developed solution should provide different levels of security since it incorporates sensitive personal data. Access control using a proper authentication system, as well as proper anonymization techniques, will be applied to prevent illegitimate users from accessing sensitive information.

### 2.3.2 System Architecture and Technical Components

Based on the system's requirements analyzed in the previous section, we came up with the proposed CHT energy monitoring architecture. Given our aim of creating a flexible and interoperable IoT framework, we opted to utilize the Z-Wave protocol as the basis for our solution since it facilitates communication between devices manufactured by different companies, enhancing consumer flexibility and usability of the developed framework. The overview of the system architecture is presented in Fig. 2.1. The proposed framework facilitates a Z-Wave communication between the monitored devices and the IoT gateway developed for this purpose. The gathered data is subsequently transmitted to a cloud infrastructure for storage, employing a RESTful API. Ultimately, the end-user is able to access their consumption data via an application developed for this purpose, which communicates with the cloud infrastructure utilizing a RESTful API to retrieve the requested information.

The technical setup used to build our solution is depicted in Fig. 2.2. The developed framework includes the Aeotec Smart Switch 7, Nano Switch, Heavy-duty Smart Switch, and Multi-sensor 6 to collect appliance level and environmental data. For the total consumption data, we used the Aeotec Home Energy Meter Gen5, which can record up to 200 amps with 99% accuracy. Finally, the gateway, which is considered the most important component of the whole framework since it acts as a bridge between the devices and the cloud, is composed of a Raspberry Pi Model 4 equipped with an Aeotec Z-Pi7 Z-wave daughter card. The inclusion of the Z-wave daughter card facilitates the Raspberry Pi's Z-wave communication capabilities with other Z-wave devices, thereby enabling it to receive the collected consumption data. A detailed description of all the technical components can be found in table 2.3.

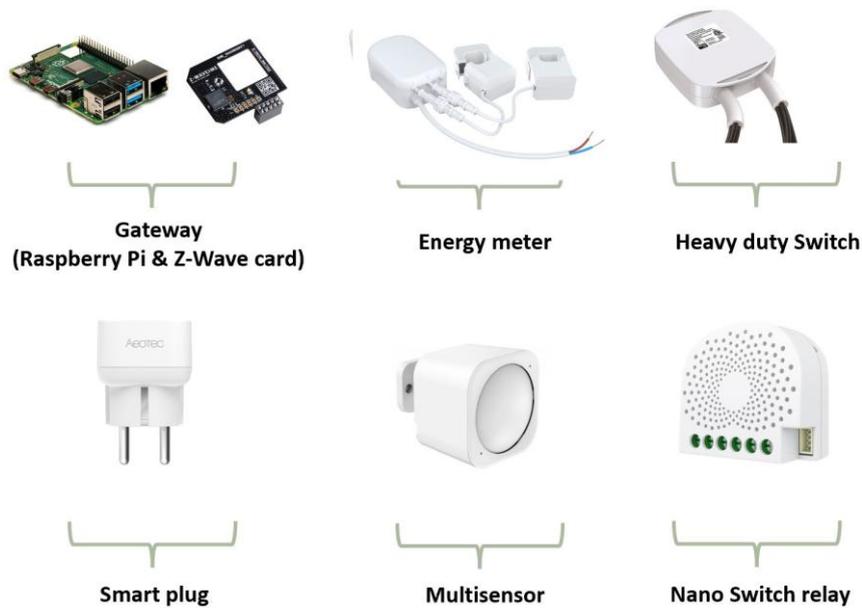

Figure 2.2 Technical components of the proposed CHT Home Energy Monitoring framework.

The used equipment is able to collect a variety of different information, including Amber (A), Volt (V), Watt (W), Kilowatt-hour (kWh), and Kilovarh (kVarh) for the energy monitoring equipment as well as temperature ($°C$), Lux, Ultraviolet index (UV) and humidity (%) for the multisensor. One additional benefit of the proposed solution is its cost-effectiveness in comparison to other paid services while simultaneously providing a greater degree of flexibility than the pre-existing built-in solutions. For instance, in the scenario of monitoring the overall energy consumption of a household as well as five individual appliances, the estimated cost would be approximately 300 euros. This cost is notably lower than that of analogous setups offered by industrial solutions. Nonetheless, the precise cost of this approach is subject to variation and is contingent upon factors such as the square footage of the residence, the quantity and type of appliances being monitored, and other relevant characteristics of the specific use case.

### 2.3.3 Gateway Software Stack

In this section, we describe the developed software components of the gateway that they were built to serve the data transmission process between the devices and the central server. The software stack of the gateway consists of 3 services, as shown in Fig. 2.3. The detailed roles and functionalities of the three services are described in the next subsections.

| Technical Equipment | Specification / Use | Cost Per Unit (euro) |
|---|---|---|
| Raspberry Pi 4 | Gateway | ≈ 60 |
| Z-Pi7 Z-wave daughter card | Gateway | ≈ 55 |
| Home Energy Meter Gen5 | - Measure aggregate<br>- max 200amp<br>- 150m wireless range | ≈ 80 |
| Smart Switch 7 | - Measure appliances<br>- max 13amp<br>- 35m wireless range | ≈ 40 |
| Nano Switch | - Measure switches<br>- max 15amp<br>- It powers automation and schedulling | ≈ 55 |
| Heavy Duty Smart Switch | -Measure heavy loads<br>- max 40amp<br>- Hot water systems, electric car chargers | ≈ 100 |
| Multosensor 6 | - Temperature<br>- Humidity<br>- Light level<br>- Motion - presence | ≈ 50 |

Table 2.3 Technical components.

**A. Z-wave JS UI service**

Z-wave JS is an open-source software library that allows developers to create applications that are based on the Z-wave communication protocol. Z-Wave JS comes with a user interface (UI) called Z-Wave JS UI, which is a web-based application that allows users to configure and manage their Z-Wave network. The UI is built on top of the Z-Wave JS server, which is responsible for communicating with the Z-Wave devices. An important aspect of Z-wave JS is that it is compatible with a wide range of platforms, including Linux, macOS, windows, and Docker. In our system, we used a dockerized version of this service which provides a secure and easy-to-deploy solution.

The main functionality of this service is to communicate with the Z-wave devices through the Z-wave communication protocol and forward the collected data to the Z-wave service through the MQTT protocol, which is widely used in IoT applications, where low power and low bandwidth requirements are essential [49].

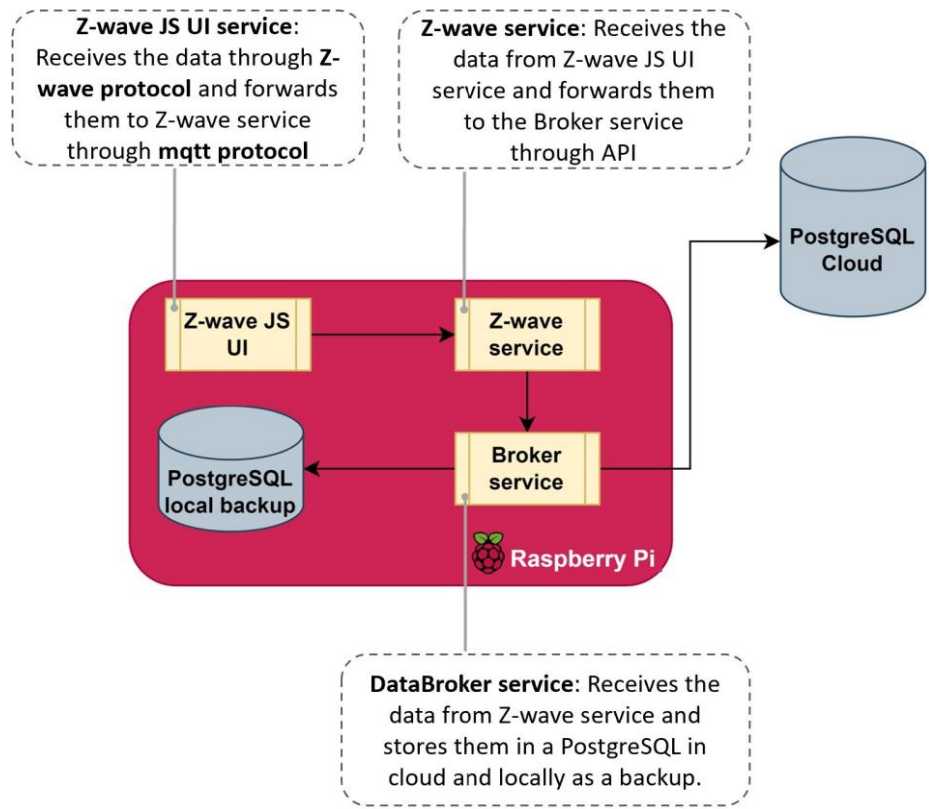

Figure 2.3 Gateway software services

**B. Z-wave service**

The Z-Wave service is a component developed using Node.js, which functions to obtain data from the ZwaveJS UI via a local MQTT broker. The service is responsible for collecting data and determining which data is to be sent to the Databroker service. Additionally, it manages a configuration file that contains information about the devices, which can be used to restore the system in the event of hardware failure. The Z Wave service is also responsible for handling actuation commands that are received from the cloud server by retrieving the commands and forwarding them to the appropriate device. Z-wave service can be considered as the system's 'brain' and plays an essential role in promoting system interoperability, providing users with parameterization capabilities, and enhancing overall system flexibility.

**C. DataBroker service**

The DataBroker service represents a local Node.js component, which is designed to fetch and verify the formatted data originating from the Z Wave service before its transmission to the cloud. Additionally, this service keeps local backups of the collected data to a local PostgreSQL, which is set on the gateway. This setup eliminates the data gaps even in the case of a network malfunction which will prevent the gateway from sending the data to the cloud. Upon network recovery, the Broker service proceeds to transmit all unsent data to the main PostgreSQL database, which is installed on the cloud side and stores all the collected data. The communication between the DataBroker service and the cloud is executed through RESTful API calls.

## 2.3.4 Graphical User Interface Platform

The concluding element of the framework supplied entailed the implementation of a graphical user interface that allowed for the retrieval of energy monitoring feedback and facilitated interaction with the system. This interface was constructed using JavaScript and Node.js as the primary technologies. JavaScript, a widely employed programming language, is well-suited for the creation of web applications that are interactive and dynamic. Node.js, an open-source, cross-platform environment, permits the execution of JavaScript code on the server side and thereby provides an optimal solution for the development of the backend of the residential monitoring application.

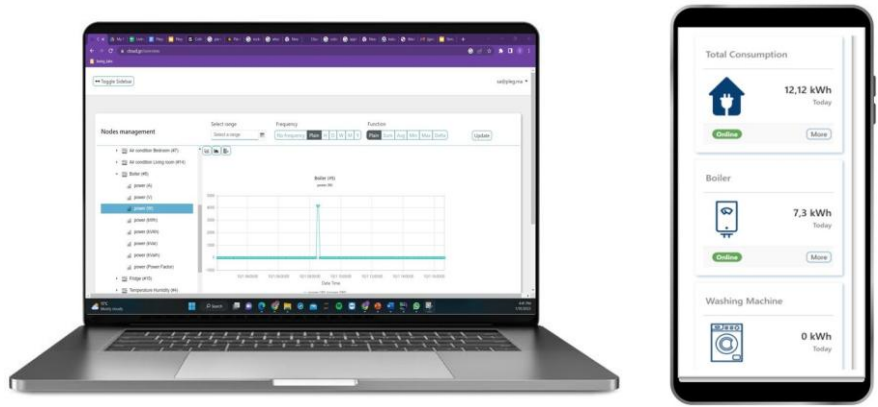

Figure 2.4 GUI of the developed desktop and mobile application.

The development of the application's user interface was facilitated through the utilization of Bootstrap, a widely adopted front-end development framework recognized for its ability to produce responsive and mobile-centric websites and web applications. Bootstrap, an open-source framework, furnishes developers with an assortment of preconfigured components, such as typography, forms, buttons, and navigation, among others, that can be readily personalized to conform to the design of the target website or application. The application's ability to deliver live energy feedback to users is enabled through the integration of the Socket.IO library. Socket.IO facilitates bidirectional, real-time communication between clients and servers by establishing a persistent connection between the two endpoints. This eliminates the need for clients to continually poll servers for updates and thus permits seamless real-time communication. Socket.IO is particularly suited for applications that necessitate instantaneous updates, such as chat applications, multiplayer games, and collaborative editing tools.

The developed application provides both a desktop and mobile version that enables users to interact and utilize its main functionalities. One of the main functionalities of the developed application is to provide energy consumption and cost monitoring feedback giving the user the opportunity to access their total and appliance level consumption for a specific period. Furthermore, the application provides data analytics and comparative diagrams by week, month, 3-month, or any other specified time frame. Finally, the application offers users a range of personalization options, including the ability to create their own visualizations and comparative diagrams tailored to their specific needs and preferences.

## 2.4 Discussion and Conclusions

In this research, we developed an end-to-end Home Energy Monitoring System utilizing Z-Wave technology for two purposes: a) offering energy feedback and b) gathering high-resolution consumption data. Our system presents an affordable and interoperable solution that does not rely on a stable internet connection, thereby preventing data gaps. However, it exhibits certain limitations that are primarily linked to the Z-Wave technology. Following the deployment of our solution in 15 households with varying characteristics and building types, we identified a fundamental limitation in the connectivity of some Z-Wave devices to the gateway. Specifically, we observed that certain devices that were situated far away from the gateway or obstructed by materials that weaken the signal strength, such as thick walls or metal surfaces, lost connection to the gateway. This was mainly due to the limited connectivity range of Z-Wave devices (30m) and the inability to establish a robust mesh network due to the building's structure. To overcome this problem, we installed WiFi smart plugs on specific appliances, which have a significantly larger range than Z-Wave devices, and enabled them to directly transmit the collected data to the cloud server. However, WiFi plugs were still preserving the limitation that they need stable and constant internet connectivity to transmit the data.

In future work, we will focus on the direction of integrating more wireless communication protocols within the developed system and extending the current software components to enable even more sophisticated features, such as house automation and remote control of the installed devices through the developed application. Finally, an intriguing direction for future research is the integration of energy monitoring into the field of Ambient Assisted Living (AAL), which has been growing rapidly in recent years [90]. Energy monitoring has the potential to contribute significantly to AAL by providing useful insights into various health features such as inactivity, sleep disorders, memory issues, variations in activity patterns, low activity routines, occupancy, and unhealthy living [148]. For example, according to [36], the usage pattern of the kettle and fridge during the night could be used to detect sleeping disorders. As a result, our future work will also explore ways to expand our existing platform to support AAL functionalities. This will involve the integration of additional sensors and algorithms to analyze the collected data and provide useful insights into the health and well-being of individuals living in the monitored environment. By leveraging the power of energy monitoring in AAL, we aim to develop a more comprehensive and effective system for supporting healthy living and enhancing the quality of life of individuals in need.

## Chapter 3

# The Plegma Dataset: Data Collection Methodology and Analysis of Domestic Appliance-Level and Aggregate Electricity Demand with Metadata from Greece

## 3.1 Introduction and Contextual Background

Over the past few years, there has been a growing adoption of smart meters on a worldwide scale. It is projected that the global market for smart meters will expand by 9% by the end of 2026, resulting in the replacement of conventional meters and contributing to the so-called smart grid transition [174]. For example, it is anticipated that 106 million smart electricity meters will be deployed in Europe between 2022 and 2027, primarily driven by large rollouts in the UK, Poland, Germany, and Greece, coupled with nationwide rollouts in various small and medium-sized European countries. As a result of these developments, it is expected that the installed base of smart meters in Europe will achieve 74% penetration by 2027 [85].

  Smart meters offer a range of benefits that can improve the energy system for both consumers and energy providers. For energy providers, smart meters offer automated and accurate meter readings, enabling streamlined billing processes and better management of energy demand during peak periods. Meanwhile, for consumers, smart meters provide real-time information about energy usage, which allows them to make more informed decisions about their energy consumption, potentially leading to cost savings. There is a wide range of problems that can be addressed using smart-meter data, as highlighted by [175, 178, 145]. For the residential sector, smart meter data enable a variety of new applications and services, including appliance-level energy feedback [30, 96] enabled by non-intrusive load monitoring (NILM) approaches, where the appliance-level consumption patterns are extracted by exclusively analyzing the aggregate household energy consumption [194, 90, 160, 14, 12, 103]; demand forecasting [137, 116]; home energy management systems (HEMS) for home automation and energy conservation [64, 140]; anomaly detection and retrofit recommendations, i.e., replacement of an energy expensive appliance [5, 32, 86, 142]; demand side flexibility and load shifting [40, 48, 149]; ambient assisted living (AAL), i.e., providing useful insights into various health features by analyzing occupants' consumption activity [61, 36]. On this basis, electricity consumption datasets are crucial for the development and evaluation of signal processing and machine learning algorithms of such applications. To ensure the reliability and effectiveness of these algorithms, it is imperative that datasets be obtained from real-world settings, where households conduct their regular activities without any interference, as opposed to laboratory conditions or synthetic datasets. Such a methodology enables more accurate and effective testing of these methods, ensuring their reliability and usefulness in real-world applications [125].

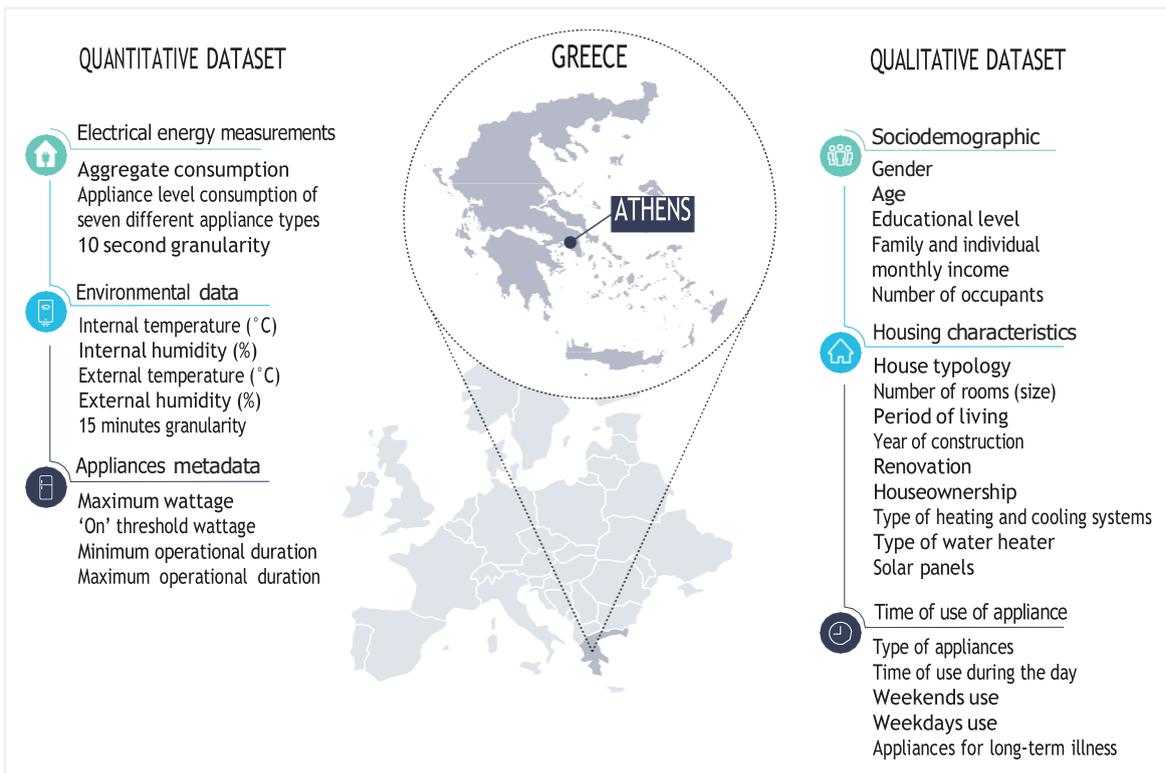

Figure 3.1 Plegma Dataset Overview.

There are a number of open-source residential datasets [82, 37] that provide electric consumption data with varying characteristics; some of the most popular are presented in Table 3.1. These characteristics include measured electrical quantities (e.g., current, voltage, active power, apparent power), sensor placement (e.g., single point sensing, circuit level, and appliance level), campaign duration, campaign location (e.g., country, municipality), and metadata (e.g., building properties, user demographics) [101]. Although all characteristics are important for determining the possible uses of each dataset, the presence of both aggregate and individual appliance measurements is crucial due to its vital impact on their potential usage, from NILM [74, 6] to appliance-level related applications such as understanding consumption patterns [124] in life cycle analysis studies [128] and mixed methods studies in relation to domestic practices [156].

The Plegma dataset [10] (Fig. 3.1), presented in this paper, is the only public dataset providing residential electricity consumption measurements at a 10-second temporal resolution in Greece, and one of the first of its kind in the Mediterranean area. The dataset captures consumption patterns that are typical to the local climate and lifestyle, offering insights into understanding the characteristics of regional energy use. These insights are important for a deeper understanding of regional energy dynamics, enabling comparisons with other areas and aiding in the evaluation of the transferability of energy-related applications across various settings. Comprised both quantitative and qualitative components, the Plegma dataset offers a dual perspective on energy use. The quantitative section encompasses aggregate household consumption and itemised appliance-level data from 13 distinct households. This includes data on specific appliances like air conditioners and electric water boilers, which are less commonly recorded in other datasets. Additionally, the Plegma dataset incorporates measurements of both internal and external environmental parameters, specifically temperature and humidity. The qualitative data cover details on sociodemographics, building characteristics, and patterns of appliance usage; crucial information for energy flexibility strategies. Data collection of the Plegma dataset [10] started in July 2022. The collected data are recorded at a sampling rate of 10s in order to be similar to the specifications of SMETS2 HAN [167] available recommended resolution and to ensure the real-world applicability of the developed solutions, which are based on our dataset. This sampling rate mirrors the granularity chosen by other prominent datasets, such as REFIT and UK-DALE, which are sampled at 8 seconds and 6 seconds, respectively, underlining a standardized approach to granularity in datasets of this nature. Other datasets, such as REDD, BLUED, and SustDataED2, have a high-frequency sampling rate (over 10 kHz), but they are recorded only for a few weeks and include a very limited amount of houses. Others, such as AMPds, IHEPCDS, IEDL, and ECD-UY, have been recorded at a lower temporal resolution of 1 min or less, limiting their usability for high-frequency applications [90].

Following the suggestions of [101], which provides recommendations for electricity consumption datasets collection, storage, and provision, we ensure the Plegma dataset's interoperability and comparability. Some of the most important recommendations that the Plegma dataset addresses are the *campaign duration*, which lasts for more than a full year, enabling capturing gradual changes in appliance usage patterns (e.g., due to seasonal changes or human behavior), *sensor placement* which follows a deployment of sub-metering devices providing consistent data for each of the monitored houses, *metadata* which provide a variety of demographic data and building characteristics, *file format* which provides a comma-separated values (CSV) format that is widely used to store similar data and is also compatible with NILM tools and energy analytics algorithms, and *accessibility of dataset* which is open access and easily accessible through Strathprints, the University of Strathclyde's institutional repository.

| Dataset | Loc. | Duration, Year | No. Houses | Sensor placement | Data | Granularity | Metadata |
|---|---|---|---|---|---|---|---|
| IHEPCDS [70] | FR | 47 months, 2006 | 1 | Agg., 3-Sub. | P,Q | 1 min | - |
| REDD [102] | USA | 1 month, 2011 | 6 | Agg., 9-24 App. | P,V | Agg. 15 KHz App. 1Hz | - |
| BLUED [58] | USA | 8 days, 2011 | 1 | Agg. | V, I | Agg. 12 KHz | ON-OFF transitions |
| UK-DALE [94] | UK | 655 days, 2015 | 5 | Agg., 5-54 App. | V, I | Agg 16 KHz App. 6 sec | Building properties Occupants attributes |
| APMds2 [118] | CA | 2 years, 2012 | 1 | 21 App. | V,I,f Pf,P,Q S,E | 1 min | Building properties Occupants attributes Ambient information |
| ECO [25] | CH | 8 months, 2014 | 6 | Agg., 6-10 App. | P,V,I, Q, Φ | 1Hz | Occupancy information |
| GREEND [122] | AT IT | 1 year, 2014 | 9 | 9 App. | P | 1Hz | - |
| REFIT [125] | UK | 2 years, 2015 | 20 | Agg., 9 App. | P | 8 sec | Building properties |
| DRED [170] | NL | 6 months, 2015 | 1 | Agg., 12 App. | P | 1 Hz | Building properties Ambient information Occupancy information |
| IDEAL [138] | UK | 23 months, 2020 | 255 | Agg., App. | P | 1 Hz | Ambient information Occupant information |
| IEDL [38] | IN | 1 year, 2022 | 1 | Agg., 5 App. | P | 1 min | Appliance information |
| SustDataED2 [135] | PT | 96 days, 2022 | 1 | Agg., 18App | P,Q, V,I | Agg. 13 KHz App. 0.5 Hz | ON-OFF transitions |
| ECD-UY [39] | UY | 21 days, 2022 | 110.953 | Agg., 9 App. | P | Agg. 15 min App. 1 min | Occupants information Building properties |

Table 3.1 An overview of NILM datasets. The table summarizes the country and year of the release for each dataset, the number of houses included, as well as the duration of the dataset, the measured variables, and the granularity and available metadata. Agg.=Aggregate, App.=Appliance, Sub.= Power circuit. Active Power (P), Reactive Power (Q), Apparent Power (S), Energy (E), Frequency (f), Power Factor (pf), Phase Angle ($\varphi$), Voltage (V) and Current (I).

In addition to the aforementioned recommendations, the Plegma dataset [10] also rigorously aligns with the FAIR data principles [181, 54], embodying the core research integrity values of honesty, cooperation, reliability, and accountability. Findability of the dataset is ensured through standard DOI identification and is easily discoverable through a standard search engine . Regarding Accessibility, the dataset is hosted in the University of Strathclyde's repository, guaranteeing open access along with essential documentation and open access code hosted on the project's GitHub page. The use of the CSV file format and the dataset's file organization, along with the available metadata description, strengthens Interoperability, allowing data exchange and reuse across various NILM tools and energy analytics algorithms. The inclusion of explicit licensing terms, such as Creative Commons, alongside comprehensive documentation on the required software for accessing and utilizing the datasets, enhances the dataset's Reusability aspect.

To address the ethical considerations and consent procedures associated with our data collection, we followed protocols that ensure the respect and privacy of all participating households. Participation in this study was entirely voluntary, with comprehensive informed consent obtained from each household prior to data collection. This process involved detailing the study's purpose, the nature of the data to be collected, and the use of such data for research and innovation within the scope of the European Commission-funded Marie Skłodowska-Curie Action GECKO project. Participants were informed that they could withdraw at any time without any consequences. All collected data were anonymized to protect participant privacy, with personal identifiers removed and stored securely within Plegma Labs' databases. This approach underscores our commitment to ethical research practices and the safeguarding of participant rights and privacy.

## 3.2   Methodology

**Selection Methodology**

The houses participating in data collection include households that are part of the Athenian community, which constitutes a newly established non-profit energy community in the municipality of Attika in Greece. The community was established as an initiative by a group of technology and engineering professionals with many years of experience in research and development. Thus, the selected sample of households included in the Plegma dataset consists of households that are familiar with information and communication technology (ICT) and cover a mix of demographics, from retired and working couples to families and single-person households.

To enhance user engagement in data collection, the developed energy monitoring system was extended to incorporate an intuitive graphical user interface. This application helps participants to monitor and visualize their energy consumption in real-time, providing a powerful incentive for their participation. Some of the homes were not included in the data collection process, primarily due to connectivity issues. These issues include underground utility meters, which make signal acquisition challenging, or architecture-related obstacles, such as thick walls or metal objects that weaken or block the Z-wave signals [186]. The selected houses' occupancy information, physical characteristics, and heating system are presented in Table 3.2.

| House | Occupancy | Dwelling age | Dwelling type | Size (#rooms) | # Monitored Appliances | Electric Phase | Heating type |
|---|---|---|---|---|---|---|---|
| 1 | 1 | 1970 | apartment | 3 | 5 | single-phase | Radiator oil |
| 2 | 1 | 1965 | apartment | 3 | 5 | single-phase | Electric heater |
| 3 | 1 | 1970 | apartment | 4 | 5 | single-phase | A/C |
| 4 | 3 | 2000 | apartment | 4 | 5 | single-phase | Electric heater |
| 5 | 3 | 1980 | detached house | 4 | 5 | three-phase | Radiator oil |
| 6 | 2 | 1960 | apartment | 4 | 3 | single-phase | Radiator gas |
| 7 | 1 | 1980 | apartment | 4 | 5 | three-phase | Air-to-air heat pump |
| 8 | 1 | 2010 | apartment | 3 | 4 | single-phase | Underfloor heating |
| 9 | 1 | 1960 | apartment | 3 | 3 | single-phase | Electric heater |
| 10 | 4 | 2000 | apartment | 3 | 6 | single-phase | Radiator oil |
| 11 | 3 | 1965 | apartment | 4 | 7 | single-phase | Radiator oil |
| 12 | 2 | 1985 | apartment | 5 | 6 | three-phase | Radiator oil |
| 13 | 1 | 1980 | apartment | 3 | 4 | single-phase | Radiator gas |

Table 3.2 An overview of the houses included in the study. The "Occupancy" column indicates the total number of individuals residing in the home during the observation period. The column labeled "Number of Monitored Appliances" displays the total count of electrical devices in the home from which data is being gathered. Additionally, the number of rooms provides information on the size of each residence. The primary heating system used in each house is detailed in the Heating type column. It's important to note that all the houses in our study utilize air conditioning (A/C) units for heating purposes, in addition to their main heating systems. However, there are a few exceptions: House 3 exclusively uses A/C for heating, while Houses 6 and 9 do not have A/C installed.

In each house, both the aggregate and appliance-level consumption were monitored. Appliance selection was motivated by incorporating appliances with relatively high electricity consumption that are also commonly used in Greece and other countries with similar environmental conditions. Some appliances included in the Plegma dataset are A/Cs, which many households in Greece use for heating and cooling, and electric water boilers, which cannot be found in many other datasets. The Plegma dataset also includes frequently used appliances such as refrigerators, washing machines, and dishwashers, which are also presented in the majority of the electric consumption datasets. Table 3.3 presents a detailed list of the appliances monitored in each household, while the dataset also provides information regarding their wattage (i.e. maximum power drawn by the appliance) and minimum threshold for an appliance to considered as 'on', the requisite minimum duration for an appliance to be categorized as active and the requisite minimum duration for an appliance to be categorized as inactive. All these information can be used as a guideline to help the dataset users assess the potential for transferability to their use cases and further refine their analysis by being able to determine the operational status of each specific appliance.

|  | House ID | | | | | | | | | | | | | |
|---|---|---|---|---|---|---|---|---|---|---|---|---|---|---|
| Appliance | 1 | 2 | 3 | 4 | 5 | 6 | 7 | 8 | 9 | 10 | 11 | 12 | 13 | total |
| Washing Machine | ✓ | ✓ | ✓ | ✓ | ✓ | ✓ | ✓ | ✓ | - | ✓ | ✓ | ✓ | ✓ | 12 |
| Dishwasher | - | - | - | - | ✓ | - | - | ✓ | - | ✓ | - | ✓ | - | 4 |
| Air Conditioner | 2✓ | ✓ | 2✓ | 2✓ | ✓ | - | 2✓ | ✓ | - | 2✓ | 3✓ | ✓ | ✓ | 18 |
| Fridge | ✓ | ✓ | ✓ | ✓ | ✓ | ✓ | ✓ | ✓ | ✓ | ✓ | ✓ | 2✓ | ✓ | 14 |
| Boiler | ✓ | ✓ | ✓ | ✓ | ✓ | ✓ | ✓ | - | ✓ | ✓ | ✓ | ✓ | ✓ | 12 |
| Washer Dryer | - | - | - | - | - | - | - | - | ✓ | - | - | - | - | 1 |
| Kettle | - | ✓ | - | - | - | - | - | - | - | - | - | - | - | 1 |

Table 3.3 Monitored appliances in each house. The final column shows the total number of the same type of appliances.

## Data collection setup

To facilitate the data-gathering process, an end-to-end energy monitoring and data collection framework was developed. An overview of the developed data acquisition system is shown in Fig. 3.2, whereby monitored devices communicate with the IoT gateway and transmit data every 10 seconds. To ensure dependable, scalable, and high-performing equipment for the data collection and monitoring system, commercially available hardware from reputable home automation companies such as Aeotec and Qubino was used. Raspberry Pi was selected as the IoT gateway for the data collection and monitoring system, as it is widely recognized for its suitability and versatility in home automation solutions [62, 150, 106].

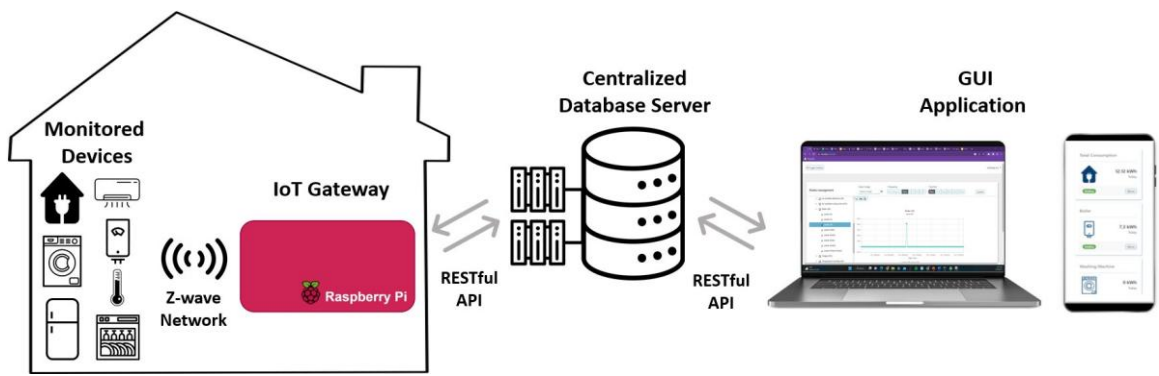

Figure 3.2 Overview of the developed energy monitoring and collection framework.

The overview of the developed energy monitoring framework can be depicted in Fig. 3.2, which comprehensively illustrates all the different stages of the data acquisition process. Firstly, The monitored devices communicate with the IoT gateway through the Z-Wave communication protocol. Subsequently, the gateway forwards the data to a central database server, AMD 2nd generation EPYC CPUs (8 VCPU, 16 GB RAM), and stores them in a PostgreSQL for secure preservation, leveraging a RESTful API. Finally, the collected data can be accessed through the developed GUI application, which directly communicates to the database server via RESTful API calls. Our solution utilizes the Z-Wave communication protocol, providing a stable interaction between the monitored devices and the gateway. The basic reason for selecting the Z-Wave protocol is that it was developed primarily for use in connected home technology, and it is also recommended for the SMETS2 ecosystem [111, 167]. Furthermore, compared to other wireless technologies like Bluetooth and Zigbee, it provides higher reliability and coverage [97, 47, 11]. Finally, Z-wave stands out for its power efficiency relative to WiFi devices, where they consume much more power and can pose serious drawbacks for battery-powered devices such as environmental sensors [46]. The majority of the advantages of this technology stem from Z-Wave's utilization of a mesh network topology [186], which allows devices to communicate through a series of nodes. Specifically, every Z-Wave device within the network can act as a repeater, extending the network's range and reliability as they enable them to find alternative paths to communicate if one path is blocked or unavailable [111].

**IoT Gateway**

One of the most important components of the developed data collection and monitoring framework is the IoT gateway, which acts as a bridge between the monitored devices and the central server. Specifically, it consists of a Raspberry Pi Model 4 that is equipped with an Aeotec Z-Pi7 Z-wave daughter card. The inclusion of the Z-wave daughter card enables the gateway's communication with the Z-wave devices, allowing it to receive and collect consumption and environmental data. These data are first stored in a local PostgreSQL database, and they are subsequently forwarded to the central server. This configuration eliminates the data gaps in the collection process since it allows our system to persist the data collection even in the case of a network disruption or internet connectivity failure, which would otherwise prevent the gateway from transmitting the data. To enable the data collection process described above, the appropriate software stack was developed, comprising three distinct modules as illustrated in Fig.3.3. A detailed description of the role and functionality of each service is provided below.

- **Z-wave JS UI service** [1] is an open-source software that enables developers to build IoT applications utilizing the Z-wave communication protocol. This service is also equipped with a user interface (UI), which is a web-based application that enables users to configure and administer their Z-wave network in an intuitive and easy way. The developed framework employs a dockerized version of the service, which provides a secure and easily deployable solution. The basic functionality of the Z-wave JS UI service is to communicate with the monitored devices utilizing the *Z-wave communication protocol* and forward the collected data to the Z-wave service.

- **Z-wave service** [2] is a service developed in Node.js and includes the main functionalities of the IoT gateway. Its main role is to obtain the collected data from the Z-wave JS UI service via *MQTT communication protocol* and forward them to the DataBroker service. MQTT is a lightweight communication protocol that has been widely utilized in many IoT solutions since it is intended to work under low bandwidth on low-power machines [80, 152]. Furthermore, this service is also responsible for 1)- determining which data are going to be forwarded to the DataBroker service, 2)- managing the configuration file of the monitored devices, which can be used to restore the system in case of hardware failure, and 3)- handling actuation commands to facilitate the communication between the server and the devices. Z-wave service promotes the framework's interoperability by providing system parameterization functionalities and enhances its flexibility and compatibility with other solutions.

---

[1]code available at https://github.com/zwave-js/zwave-js-ui

[2] code available at https://github.com/sathanasoulias/Plegma-Dataset/tree/main/data_collection

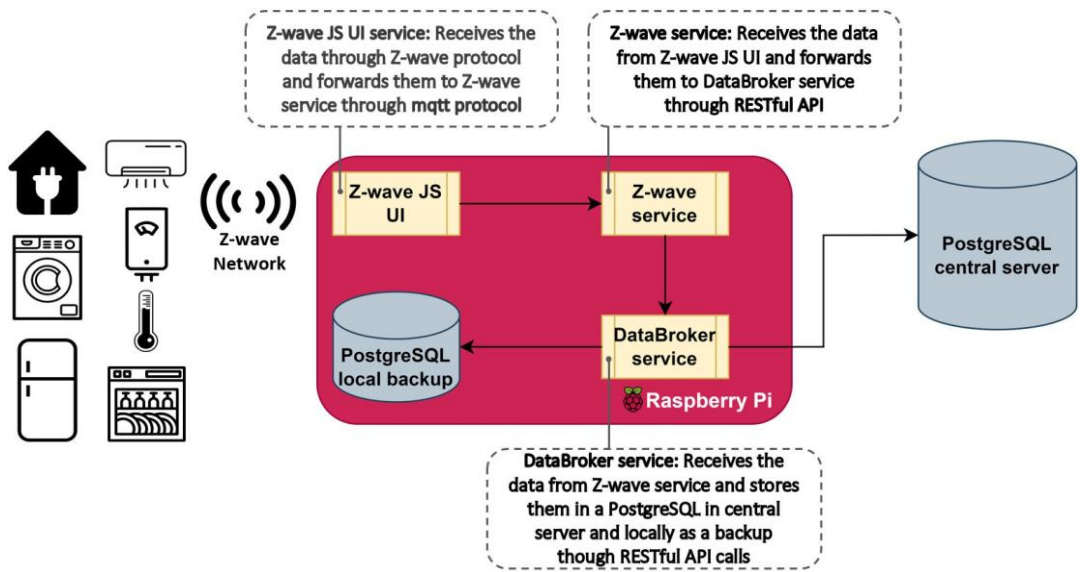

Figure 3.3 Overview of the developed IoT Gateway software services.

- **DataBroker service** [2] is a service developed in Node.js, and it is responsible for receiving the collected data from the Z-wave service and verifying their format structure before transmitting them to the central database server. DataBroker service is also responsible for keeping local backups of the collected data to a secondary PostgreSQL database located in the IoT gateway, avoiding data gaps in case of network failure. The communication between both the local PostgreSQL database and the central PostgreSQL database server is executed using *RESTful API calls*.

### Household aggregate consumption

Whole-house consumption data has significant value since it emulates the data that will be recorded by smart meters in the near future and can be used to build applications in various domains such as demand forecasting, demand response, detection of activities of daily living, or energy disaggregation.

The household aggregate consumption was measured using the Aeotec Home Energy Meter [3], which contains a single phase current clamp and a Z-Wave transmission module that transfers readings every 10 seconds using a Z-Wave frequency 868 MHz to the Z-wave JS UI service of the IoT gateway.

Aeotec Home Energy Meter has been successfully used in many IoT-based solutions [190, 112, 45] as it provides robust wireless connectivity with a range of 150 meters as well as secure broadcasting with AES-128 encryption protocol. Furthermore, it can record up to 200 amps with 99% accuracy in real-time, making it an ideal solution for home energy monitoring. Finally, another reason for our choice is that Aeotec Home Energy meters operate under the Z-wave communication protocol, which exhibits superior reliability and broader coverage compared to other wireless protocols [46]. The monitored measurement readings included Watt (W), Volt (V), Ampere (A), and kilowatt hour (kWh).

Three houses in the study had three-phase power. In these three cases, a three-phase version of the Aeotec Home Energy Meter was installed, which contains 3 clamps, one for each phase. The rest of the process follows the same configurations, collecting readings from the same measurements with a 10-second granularity.

**Individual appliance consumption**

For the purpose of the study, every household was equipped with a suitable number of smart plugs designed to gather consumption data at the appliance level. Each smart plug provided readings of the active power (W) of each selected appliance reported every 10 seconds.

The data collection process of domestic devices such as washing machines, dishwashers, A/Cs, and refrigerators was facilitated by Aeotec Smart Switch 6 plugs [4]. Smart Switch 6 plugs operate under the Z-wave communication protocol, are easy to install, and can experience an error of at most ± 1%. For the water boiler, which is a high load-consuming device and has never been included in any similar dataset, a Qubino smart meter [5] was used since it can operate up to 65 A in a Z-Wave frequency of 868.4 MHz.

The readings from all the monitored devices were transmitted to the IoT gateway utilizing the Z-wave communication protocol and the Z-Wave JS UI service. Subsequently, these data were forwarded to the central database server for storage and further pre-processing. The recording of the appliances started during the initial installation of the equipment, and recruited households were given specific instructions to avoid unplugging or repositioning the smart plugs during the data collection period.

---

[3]specification available at https://aeotec.com/products/aeotec-home-energy-meter/
[4]specification available at https://aeotec.com/products/aeotec-smart-switch-6/
[5]specification available at https://qubino.com/products/smart-meter/smart-meter-techold/

## Environmental data

The environmental data included in the Plegma dataset encompass both internal and external temperature (°C) and humidity (%). The data collection of the internal environmental data was conducted using the Multisensor 6 [6], which operates under the Z-wave communication protocol and presents an accuracy of ± 1°C for the temperature and ±3%RH for the humidity data. The indoor environmental data were recorded every 15 minutes and were forwarded to the central database server through the IoT gateway.

The outdoor environmental data were collected using the open API that provides historical weather data produced by MET Norway [7], and they are available under a Creative Commons license. The dataset includes both the external temperature and humidity with a 1-hour granularity.

## Server & database

The consumption data from all the monitored appliances, together with the whole-house consumption and the internal environmental data from all the houses participating in the survey, are stored in a PostgreSQL server hosted by Hetzner, in Nuremberg, Germany, with the following specifications: AMD 2nd generation EPYC CPUs (8 VCPU, 16 GB RAM), running on Ubuntu 20.04.5 LTS (GNULinux 5.4.0-131-generic x86_64) with PostgreSQL 14.5.

The operation of monitoring the health status and connectivity of the monitored devices was carried out using a developed Javascript script. This script evaluated the temporal difference between the last recorded measurements in the database and, in the case of a significant time gap, in excess of 30 minutes, an automatic email was sent to flag the issue for further investigation. The health status of the monitored devices was also available through the developed UI application.

## Sociodemographic, building characteristics and appliance usage metadata

A baseline questionnaire on participants' sociodemographic aspects and building characteristics was developed to provide an overview of social and building factors that may impact energy consumption, such as number of occupants, age, educational background, gender, income, house typology, size of the house, heating and cooling systems, among others. Furthermore, following Trotta et al. [166], questionnaire on the time of use of electric appliances, detailed data on participants' habits and routines in relation to the use of appliances were collected for further scrutiny of variation and patterns of such activities, which are related to energy demand and flexibility. Both questionnaires were applied to all 13 participants in the summer of 2023 via phone (n = 4) and face-to-face (n = 9), primarily in

---

[6]specification available at https://aeotec.com/products/aeotec-multi-sensor-6/
[7]documentation available at: https://api.met.no/weatherapi/

English but also in Greek when necessary. The first author conducted the data collection by asking the questions and filling out the questionnaire, a procedure that lasted, on average, 20 minutes.

## Data pre-processing pipeline

The collected electric and environmental data were pre-processed and cleansed in order to be used as a common baseline for researchers using the Plegma dataset [10]. The pre-processing was carried out using Python language version 3 and several libraries, including Pandas, Numpy, and Plotly for visualizations. The basic components of the proposed data processing pipeline are described in the following sections. The developed code for visualizing and preprocessing the dataset can be found in the project's GitHub page [8].

### Data synchronization

The installed monitoring devices (energy meters, smart switches, and environmental sensors) were only capable of broadcasting their readings, which resulted in the readings not being synchronized with each other as shown in Fig. 3.4. The timestamp assigned to each measurement was the UNIX timestamp when the corresponding data were received at the IoT gateway.

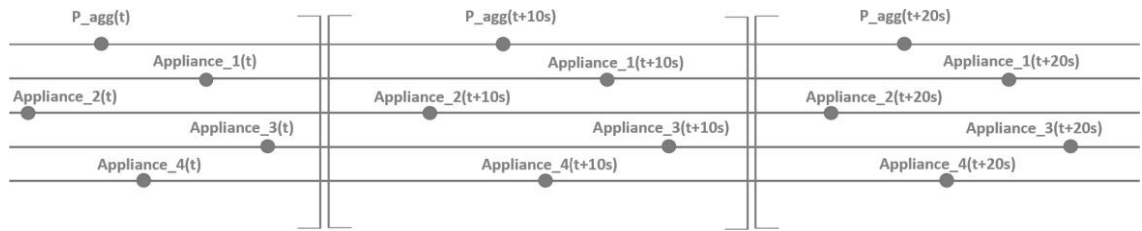

Figure 3.4 Overview of synchronization issue. Each line represents a sensor, P_agg being the whole-house consumption. The dots represent each recorded measurement, and t is the timestamp when the corresponding data point was received at the IoT gateway.

The synchronization of the different measurements was achieved through the technique of resampling, a common method used in temporal data analysis. Resampling involves altering the frequency of the data samples to align them in a common time vector [83]. In this context, resampling involved aligning the various electrical and environmental time series to a common time vector, thereby synchronizing their measurements at consistent 10-second and 15-minute intervals correspondingly. Let $X(\tau)$ denote the original time series where $\tau$ signifies the time index. Resampling aims to transform $X(\tau)$ to $X'(\tau')$ where $\tau'$ represents a regular grid of 10-second intervals. For every timestamp $\tau'$ in the new time vector, the corresponding value of $X'(\tau')$ was estimated based on the values $X(\tau)$ closest to $\tau'$. Thus, the relationship between the resampled and original time series values can be formally expressed as $X'(\tau') = X(\tau)$, where $\tau$ is the timestamp in the original time series that is closest to $\tau'$. This procedure resulted in a set of electrical (aggregate and appliance level) and environmental (temperature and humidity) time series that were harmonized to the same time vector, facilitating synchronized analysis on a uniform time grid.

---

[8]code available at https://github.com/sathanasoulias/Plegma-Dataset

**Abnormal measurements**

In the analysis of the collected data, sporadic discrepancies were identified. Specifically, these inconsistencies manifested as unexpected spikes and measurements that exceeded the standard power thresholds of the monitored devices and the ambient environmental conditions. Potential explanations for these inaccurate readings could be electrical interference since IoT devices can be affected by other electronic devices in close proximity, leading the sensor to log unexpected spikes or abnormal values. Additionally, software glitches in the IoT sensor or imprecise calibration and built-in offset inaccuracies can be some other contributing factors to observed erroneous measurements. Nonetheless, these wrong readings comprised a mere 1.65% of the total measurements. To address the issue of atypical measurements, the detected abnormal data points were replaced with the preceding normal reading (see Algorithm 1). A reading is deemed 'normal' if it is found within the expected range. For electrical data, these boundaries are provided in the `appliances_metadata.csv` file. For environmental metrics, acceptable temperature ranges are set from $-10$°C to $50$°C, and humidity values are confined between 0% and 100%.

---

**Algorithm 1** Pseudocode for cleaning data from abnormal values.

---

**Require:** data, range
**Ensure:** Cleaned data from abnormal values
1: normal ← *undefined*
2: **for** $n$ ← 1 to length(data) **do**
3:     **if** data($n$) is within range **then**
4:         normal ← data($n$)
5:     **else**
6:         data($n$) ← normal
7:     **end if**
8: **end for**

---

**Data gaps**

The dataset contains intermittent gaps in consumption and environmental readings, typical in large-scale data collection, presenting analytical challenges. Several potential reasons can underpin these discrepancies. Firstly, Z-Wave operates on specific frequency bands, which can occasionally face interference from other devices or electronic noise. Secondly, physical obstructions or the presence of certain materials in the communication path can attenuate the Z-Wave signals, leading to dropped packets. Additionally, network congestion due to a high number of devices communicating simultaneously or firmware issues in the Z-Wave devices themselves can result in data losses.

---
**Algorithm 2** Interpolate Consecutive Short Data Gaps
---
 1: **Input:** data, threshold
 2: **Output:** Interpolated data
 3: **for** $i = 0$ **to** length(data) **do**
 4:     *startGap* ← $i$
 5:     **while** $i <$ length(data) **and** data[$i$] is NaN **do**
 6:         $i \leftarrow i + 1$
 7:     **end while**
 8:     *endGap* ← $i$
 9:     *gapSize* ← *endGap* − *startGap*
10:     **if** *gapSize* < threshold **then**
11:         data[*startGap* : *endGap*] ← **interpolate**(data[*startGap* : *endGap*] )
12:     **else**
13:         data[*startGap* : *endGap*] remains with NaN values
14:     **end if**
15: **end for**
16: **return** data

---

To ensure the integrity of the subsequent analyses and make the dataset compatible with applications like NILM, where granularity and continuity of data are essential, a rigorous data imputation methodology was implemented. For the electrical energy measurments, gaps less than 30 seconds were interpolated, whereas in the environmental data, gaps were filled only if they were shorter than 1 hour. Interpolating short data gaps ensures accuracy due to the closeness and continuity of adjacent data points. These interpolated values thus likely reflect genuine readings, safeguarding data integrity. Conversely, larger gaps introduce greater uncertainties. Spanning a more extended period, their interpolation risks generating values that deviate from actual scenarios. Especially in applications like NILM, this can amplify errors during load disaggregation or pattern analysis. Hence, to preserve data reliability, these larger gaps were left unaltered.

## 3.3 Data Records

The Plegma dataset [10] is accessible in both raw and clean formats, and the collected data are provided as comma-separated values (CSV) files at Strathprints, the University of Strathclyde data repository. The unprocessed dataset encapsulates the originally collected data, unaltered by any preliminary processing, thereby encompassing abnormal values and data gaps attributed to equipment malfunctions. The primary rationale for offering this raw version of the dataset is to give potential users the opportunity to carry out their individual pre-processing techniques.

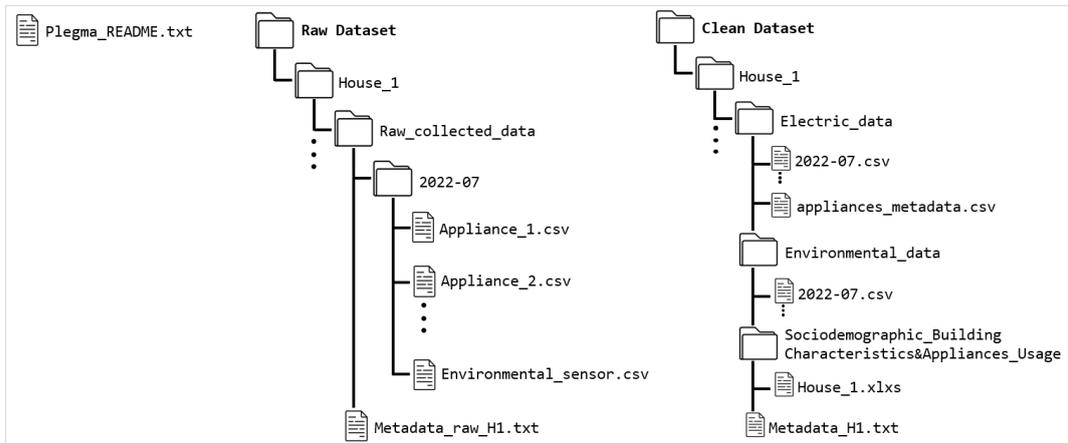

Figure 3.5 Overview of the dataset's folder structure.

The clean version of the dataset, in contrast, represents the dataset after the implementation of a proposed pre-processing pipeline. This pipeline addresses several key components, including i) the standardization of units, date formats, and column name conventions, ii) the handling of abnormal measurements and data gaps due to failure of the equipment or internet connectivity disruptions, iii) the data synchronization, and iv) the identification and flagging of known issues, timestamps where the appliance level reading exceed the monitored total consumption due to synchronization mismatch, instrumental accuracy or inductive and capacitive appliance-level loads. The folder structure of the Plegma dataset is presented in Fig. 3.5. Both `Raw Dataset` and `Clean Dataset` directories contain 13 sub-directories named `House_<i>`, where `i` is an integer between 1 and 13. In the `Raw Dataset`, each house sub-directory contains a `Metadata_raw_H<i>.txt` file with the house metadata (monitored appliances & recorded values) and the unprocessed collected data organized in each folder per month, which contains the collected data from the installed smart meter, smart plugs, and environmental sensor in CSV formats. In the `Clean_dataset` folder, each house directory contains three sub-folders for the `Electric_data`, `Environmental_data`, as well as `Sociodemographic_Building_Characteristics&Appliance_Usage` along with the house metadata `metadata_H<i>.txt`, which contains the same information as the ones presented in the raw version of the dataset.

**Electrical Energy Measurements**

Within the cleaned version of the Plegma dataset, the processed consumption metrics for each house are located within the `Electric_data` directory. Corresponding to each month, a CSV file is structured according to the nomenclature `year-month.csv` and encompasses the fields delineated in Table 3.4.

| Electric_data/year-month.csv | | |
|---|---|---|
| field | *type* | *description* |
| datetime | string | Datetime of the record in MM/DD/YYYY HH:MM:SS AM/PM format (UTC+3) |
| V | float | Instant voltage in Volts (V) |
| A | float | Instant current in Amperes (A) |
| P_agg | float | Instant aggregate power in watts (W) |
| appliance_1<br>.<br>.<br>appliance_n | float | Instant appliance power in watts (W) |
| issues | integer | Timestamps where the reading at the appliance level surpasses the total monitored consumption are marked as 1, while all other instances are flagged as 0. |

Table 3.4 Description of the records in files of the electric consumption data.

In addition to these electrical data points, The directory `appliances_metadata.csv` provides specifics for each appliance, including wattage, minimum active duration ("min_on"), and minimum inactive duration ("min_off").

**Environmental Data**

The processed environmental data of each house are available in the corresponding `Environmental_data` folder. For each month, there is a CSV file named using the `year-month.csv` convention, similar to the organization of the consumption data, and includes the fields described in Table 3.5.

| Environmental_data/year-month.csv | | |
|---|---|---|
| field | *type* | *description* |
| datetime | string | Datetime of the record in MM/DD/YYYY HH:MM:SS AM/PM format (UTC+3) |
| internal_temperature | float | Instant internal temperature in Celsius |
| internal_humidity | float | Instant internal humidity (%) |
| external_temperature | float | Instant external temperature in Celsius |
| external_humidity | float | Instant external humidity (%) |

Table 3.5 Description of the records in files of the monitored environmental data.

**Sociodemographic Building Characteristics & Appliances Usage**

The sociodemographic, building characteristics, and appliances usage data can be found in the corresponding folder of each house in a Microsoft Excel Worksheet (.xlsx) format, which is structured in three different subsections, one for each category. The sociodemographic section includes the gender, occupation, educational level, age, number of occupants, and income information; the building characteristics include information such as house topology, number of rooms, and construction year, while the time of use of appliances present information on how often and what time of day each appliance is being used according to the perception of the occupants that can be used as soft labels.

The data availability for all the included Plegma houses can be seen in Fig. 3.6. The vertical right edge shows the uptime per house, showing the percentage of the non-missing values for both electric and environmental data per house during their corresponding data collection campaign period. The average uptime across all houses was 93.44%, with House 5 having the lowest at 78.38% and

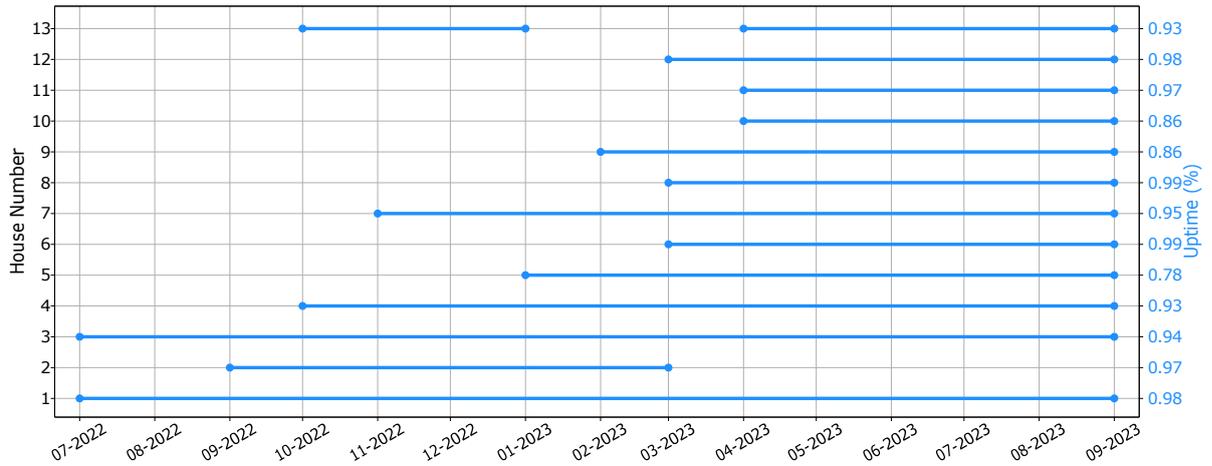

Figure 3.6 Plegma Data Availability. The left y-axis represents the corresponding house number, while the right y-axis represents the uptime of the corresponding data collection campaign.

House 6 the highest at 99.37%. Finally, a `Plegma_README.txt` file is also available in the root level of the dataset, providing information regarding the folder structure, granularity, naming convention and metadata,

## 3.4 Technical Validation

### Electric & Environmental Data

Over the course of the data collection campaign, a total of 218,410,245 readings marked with timestamps were recorded, encompassing both electrical and environmental measurements from each of the 13 houses involved in the project. Out of these, 6.86% were identified as Not a Number (NaN) values, indicative of instances where our system failed to retrieve the requested data successfully. While this percentage might initially appear substantial, it's actually a common occurrence in datasets of this nature [50] Similarly, the REFIT dataset [125], which is among the most referenced in this area, shows a close percentage of NaN values at 6.4%. This similarity suggests that the presence of NaN values is a standard aspect of data collection in such environments, arising from the complexities involved in data acquisition. [11, 177]In the electrical dataset, an 'issues' column has been incorporated to signify instances where the cumulative readings from individual appliances surpass the recorded aggregate consumption for a given sample. This condition is denoted by a value of '1'; conversely, a '0' indicates normal readings. However, the proportion of timestamps flagged with 1 in this manner within the dataset amounts to only 0.82%. As previously noted, the main factors causing this issue are related to synchronization mismatches and deviations pertaining to instrumental accuracy.

All the collected data, including electrical and environmental, has been visually inspected to verify the quality of their signatures. In all of the houses folders, the corresponding `metadata_H<i>.txt` file indicates the data availability in terms of installed sensors in each specific house. The quality of readings for certain appliances is impacted by their placement or disruption from nearby devices. This effect is particularly pronounced for appliances that are situated farther away from the IoT gateway, as well as for sensors that are installed behind large appliances like washing machines and fridges.

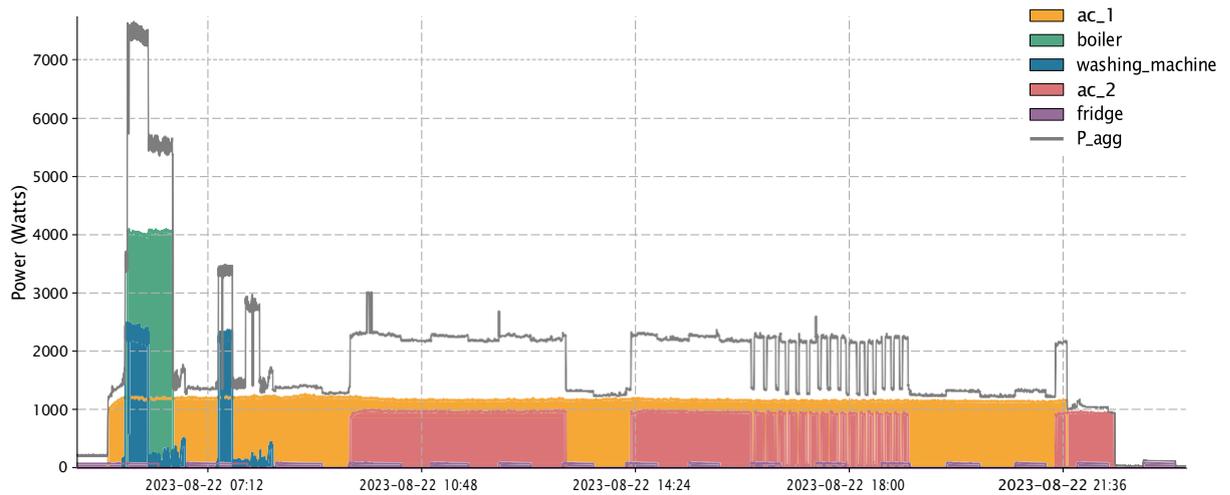

Figure 3.7 The power usage for House 1 on the 22nd of August 2023. The space between the appliance consumption plot and the total power consumption (P_agg) line illustrates the energy consumed by appliances that are not under monitoring.

Previously, we mentioned that the installed sensors did not report values synchronously with the smart meter of the aggregate consumption. Specifically, the appliance level measurements could differ by two or even three readings. However, after the synchronization pre-processing step described in the previous section, all the readings were harmonized under the same time vector. Fig 3.7 shows that even though this discrepancy occurs during the data collection process, the data signatures are well synchronized since the moments when appliances are turned on or off can be distinctly seen in the aggregate readings. Table 3.6 shows the percentage of total household consumption captured by sub-metering across the 13 houses involved in the project. According to this table, the average percentage of sub-metered consumption is 54%, while in some cases it can go up to 85% as happens in house 1. The low percentage of sub-metered energy consumption observed in some houses reflects the absence of monitoring for certain high-consumption appliances like ovens and stoves.

The main reason that we decided to exclude these types of appliances from our dataset was the increased complexity involved in the installation process of the monitoring equipment to built-in devices. Finally, Table 3.7 shows the amount of the total recorded consumption among the different appliance categories. This information could be important for applications such as NILM, where it is important to capture a large amount of appliance uses. In order to calculate the number of activations of each appliance, we utilized the `appliances_metadata.csv`, which contains appliance-specific parameters regarding their 1)-maximum wattage, 2)-the wattage threshold for being considered "on", the 3)-minimum and the 4)-maximum operational duration. Based on these data we were able to precisely determine their "on-off" status.

| Household Energy Consumption and Submetering Statistics | | | | | | | | | | | | | |
|---|---|---|---|---|---|---|---|---|---|---|---|---|---|
| *House ID* | *1* | *2* | *3* | *4* | *5* | *6* | *7* | *8* | *9* | *10* | *11* | *12* | *13* |
| Aggregate Consumption (kWh) | 2958 | 1415 | 2954 | 2818 | 3778 | 1512 | 2045 | 1222 | 1218 | 1544 | 2861 | 3664 | 3073 |
| Sub-metered consumption (kWh) | 2534 | 701 | 1557 | *1728* | *2356* | *744* | *633* | *574* | *470* | *919* | *2154* | *977* | *2128* |
| Percentage of sub-metered consumption (%) | 85 | 49 | 52 | *61* | *62* | *49* | *30* | *47* | *38* | *59* | *75* | *26* | *69* |

Table 3.6 Monitored appliances in each house. The final column shows the total number of the same type of appliances.

| Appliance Level Consumption Statistics | | | | | | |
|---|---|---|---|---|---|---|
| **Monitored appliance** | A/C | Water Boiler | Dishwasher | Fridge | Washing Machine | Kettle |
| **Total number (#) of monitored appliances** | 18 | 12 | 4 | 14 | 13 | 1 |
| **Total Consumption (kWh)** | 6656 | 5246 | 267 | 4251 | 1126 | 69 |
| **Total number (#) of appliance activations** | 2631 | 5042 | 346 | continuous | 1476 | 73 |

Table 3.7 Appliance level consumption statistics. The statistics include the number of monitored appliances, total consumption, and activations per appliance.

The environmental dataset includes measurements of indoor and outdoor temperature, as well as humidity levels. The environmental sensors were specifically placed in the living room of each house. This positioning implies that heating or cooling activities occurring in other rooms may not be recorded. Nevertheless, upon examining the respective data patterns, it becomes evident that they align with anticipated trends, effectively capturing both seasonal fluctuations as well as the heating and cooling behaviors of the residents.

An illustrative example is depicted in Fig. 3.8. In this example, the environmental data from House 3 is presented for different time periods, showcasing how these metrics vary with each season (summer and winter) and how heating and cooling practices influence them. On the left side of the chart which refers to the summer season, we observe that when the A/C is activated for cooling, it leads to a reduction in internal temperature and humidity as expected. Similarly, in the winter season chart, we observe an increase in internal temperature and a decrease in internal humidity during A/C activation since the A/C was used for heating purposes. This diagram not only validates the environmental data but also highlights their correlation with A/C usage, suggesting potential applications in various contexts.

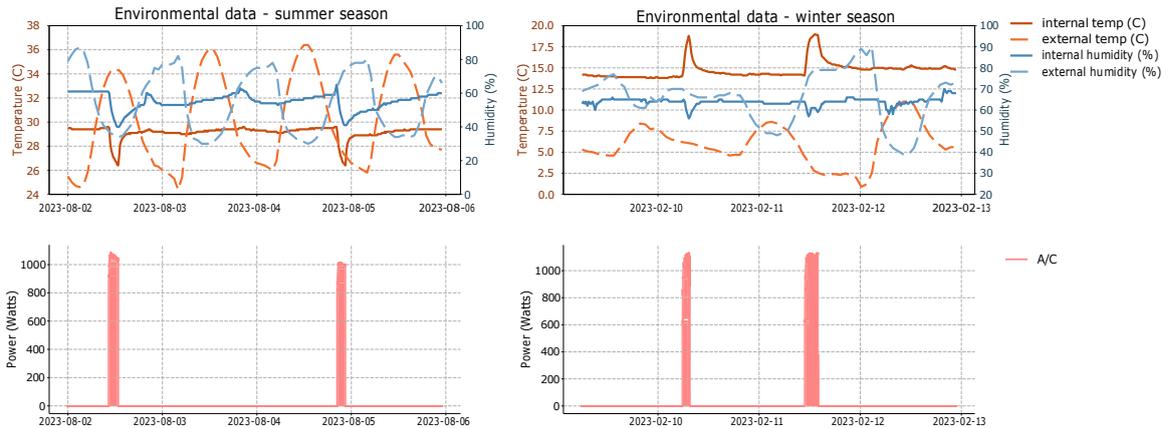

Figure 3.8 This visualization showcases environmental data collected from House 3 across different seasons, highlighting the heating and cooling practices employed by the residents. The left graph provides insights into environmental conditions in August 2023 and their variations as a result of air conditioning usage for cooling. On the right, we observe environmental data from February, revealing fluctuations driven by the use of air conditioning, which, in this instance, is employed for heating rather than cooling the house.

## Sociodemographic Data & Building Characteristics

At the individual level, 13 respondents answered questions about their gender, educational level and individual monthly income. 69.2% were men in contrast with 30.8% of women. 46.2% has bachelor's degree, 23.1% high school, 15.3% master's degree and 15.3% doctorate's degree. 23% of families earn up to €713.00 per month (up to one minimum wage), 46.1% of recruited households have a monthly income between €1,426.00 and €2,852.00 (equivalent to 2 to 4 minimum wages), and 15.3% of households earn between €2,852.00 and €4,278.00 per month (4 to 6 minimum wages)(see Fig. 3.9).

At the household level, 53.8% are young adults living alone, 15.6% are adults living with two more individuals, 15.3% are families with children, and 15.3% are elderly. 15.3% of households have a pet. The family's monthly income varies in accordance with the number of occupants in the same house; 46.1% earn between 4 and 6 wages, 38.4% between 2 and 4 wages, 7.7% earn up to 2 wages, and 7.7% earn up to 1 wage. The predominant house typology is apartments (92.3%) in contrast with only 7.7% detached houses. The number of rooms gives general information about the house size; 53.8% of households have 4 rooms, 30.7% 3 rooms, and 15.3% 5 rooms.

In terms of house ownership, 53.8% are owners, 38.4% are renters, and 7.7% have 'free use' also described as complimentary occupancy or rent-free living. Regarding the types of heating and cooling systems, most of the households have a combination of two different types. 92.3% have air-conditioner, 30.7% radiator oil, 23.1% electric heater (portable), 15.3% radiator gas, 7.7% air-to-air heat pump, 7.7% underfloor heating and 7.7% declare other options. The predominant water heater is electric boiler (84.6%), following by 30.7% of solar boilers, and 15.3% of gas boilers. No one has declared the ownership of solar panels. Finally, when it comes to year of housing construction, 46.1% were built between 1950 and 1970, 30.7% between 1970 and 1990, 15.3% between 1990 and 2000, and 7.7% post 2000.

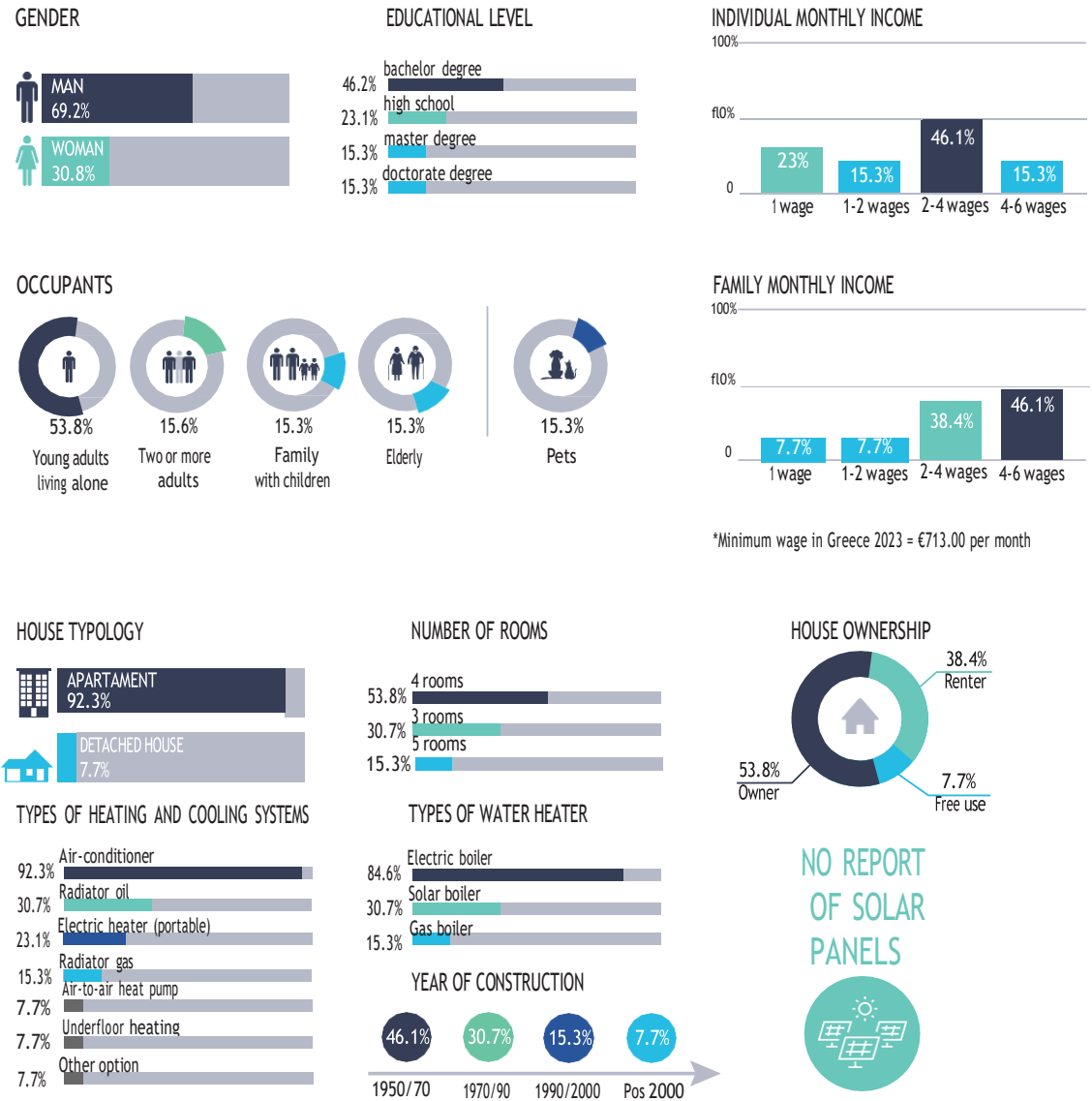

Figure 3.9 Overview of the sociodemographic and building characteristic data.

## 3.5  Discussion and Conclusions

The Plegma dataset represents a significant contribution to the field of energy monitoring and analysis, particularly within the Mediterranean context. It stands out as one of the first datasets in this region to provide detailed household-level data, encompassing electrical energy consumption, environmental metrics, and sociodemographic and building characteristics. This regional focus fills a critical gap in energy research, as Mediterranean climates, with their distinct seasonal energy consumption patterns, are often underrepresented in global datasets.

One of the dataset's unique features is its inclusion of appliances such as air conditioners (A/C) and boilers, which are significant energy consumers in Mediterranean households and are not widely recorded in other publicly available datasets. By capturing their detailed usage patterns, the Plegma dataset provides invaluable insights into the role of these appliances in overall energy consumption. Additionally, the inclusion of environmental data, such as indoor and outdoor temperature and humidity, further enriches the dataset, allowing for a deeper understanding of the interplay between energy use, seasonal variations, and environmental conditions. These insights are particularly relevant for applications like NILM, where appliance-specific usage patterns and their correlation with environmental factors are critical for accurate disaggregation and analysis.

The sociodemographic and building characteristic data add yet another layer of context, offering a holistic view of household energy consumption behaviors. Information about household composition, income levels, heating and cooling systems, and building typologies enables researchers to analyze how demographic and structural factors influence energy use. This multidimensional approach makes the Plegma dataset a valuable resource for energy efficiency studies, policy development, and behavioral modeling.

In conclusion, the Plegma dataset is a pioneering effort that provides a well-documented, multifaceted resource for advancing energy research in the Mediterranean region. Its combination of electrical, environmental, and sociodemographic data sets it apart from other datasets and makes it particularly suited for studying the unique energy dynamics of Mediterranean households.

# Part III

# Development and Optimization of DNN-Based NILM Architectures for Edge Deployment

# Chapter 4

# Graph Neural Network-Based Approach for Optimizing Non-Intrusive Load Monitoring

## 4.1 Introduction

Non-Intrusive Load Monitoring (NILM), or energy disaggregation, is an effective framework that can significantly reduce cost and time in home energy audit [27]. The idea of NILM describes the process of estimating the energy consumption of each individual appliance by relying exclusively on the total consumption of the building [89]. Thus, this approach reduces sensing infrastructure costs by relying on energy disaggregation techniques to monitor appliances' loads. NILM does not only inform the end-consumers about their potential energy savings, but it could also be used for demand-response management, or even to monitor daily activities of elderly people for security and self-empowerment. It also allows new fair pricing policies especially for green credits that benefits the entire electricity market [52].

In general, energy disaggregation approaches could be grouped into two distinct categories: (1) event-based approaches and (2) event-less approaches [199]. Event-based approaches seek to disaggregate the total consumption by means of detecting and labelling the 'on' and 'off' states of each appliance. These were the first kind of approaches to NILM and require a data collection step where a number of transitions (on-off states) from appliances of interest are collected and labelled to be used as training data [136]. On the other hand, event-less approaches, which is also the one presented within this paper, do not rely on event detection but they propose a sequence-to-sequence process [88]. Specifically, these approaches attempt to translate each sample of the aggregated signal to the corresponding signal of each appliance.

Many different approaches have been used towards NILM, from statistical approaches such as Bayesian methods [155] to probabilistic approaches using Markov Models [119]. However, with the recent success of deep learning approaches in many different domains, NILM research has been traversed from traditional signal processing approaches to deep learning structures. Deep learning models including Long Short Term Memory (LSTM), Gated Recurrent Unit (GRU) and Convolutional Neural Networks (CNN) are some of the methods that are being considered as the state-of-the-art approaches towards NILM [79]. In this context, multi-channel recurrent convolutional neural networks have been considered as another superior approach to NILM problem [87]. All these approaches are good at capturing sequential and neighboring dependencies across the aggregated signal [53]. Specifically, all the aforementioned methods model non-linear similarity metrics and relationships [51], between the aggregate (input) and the appliance specific consumption pattern (output).

In this study, we introduce a new method for energy disaggregation using Graph Neural Networks as encoders. In contrast with the state of the art methods mentioned above, GNN provide the ability of capturing non-sequential dependencies across the data via message passing between the nodes of the graphs [195]. Our approach consists of two parts: *(i) the encoding part*, which is a pre-training process and includes the transformation of the aggregate signal into a graph structure based on the entropy of the corresponding appliance's signal in order to construct enriched embeddings through the neural message passing process *(ii) the decoding part*, which is a supervised training process of a transformer-based decoder to predict the electrical consumption of a specific domestic appliance given the corresponding graph embedding. Experimental results indicate that our approach is specialized in multi-operational state appliances such as washing machine, where it produces satisfactory or even better results than the state of the art approaches. However, at this stage, our approach could only be used to disaggregate known datasets as it provides the limitation of knowing the entropy of the appliance's signal even on the testing data. Despite this limitation, this is the first GNN-NILM approach proposing a new time-invariant direction and problem modelling utilizing the entropy metric though a graph structure.

### 4.1.1 Contribution

Within this paper, we propose the first GCN approach to address the NILM problem. The scope of this study is to enrich the current knowledge and state-of-the-art methodologies for seq2seq applied in NILM by presenting a time-invariant approach. The advantages of our approach are summarized below:

- **Proposing a GCN method that is capable of learning time invariant dependencies along the power signal.** In contrast with the state of the art approaches, the proposed approach utilizes the graph structure (encoder) as well as the attention mechanisms (decoder) and identifies time–invariant dependencies across the input and output time series.

- **Proposing an architecture that is capable of disaggregating multi-operational state appliances.** Our model has a very robust performance minimizing the false positive activations and produce better results from the state of the art methods on multi–state appliances like washing machine.

## 4.2 Related Work

The enormous success of deep learning in many domains such as computer vision and natural language processing, led to the adoption of many DNN approaches applied for NILM [79]. Recurrent neural networks (RNN), long short-term memory (LSTM), bidirectional (bi)LSTM, gated recurrent unit (GRU), (bi)GRU and convolutional neural networks (CNN) are some of the methods that are considered as the state-of-the-art techniques for NILM [89]. All these techniques take the advantage of recurrent and convolutional mechanisms to identify temporal patterns in power consumption sequences.

RNN approaches utilize the feedback connections to capture temporal information in 'memory' and are well suited to identify temporal dependencies in sequential data such as power signals. In this context, auto-regressive mechanisms in deep learning have been also proposed [18]. However their basic limitation is that they cannot learn long-range dependencies[161]. LSTM is a recurrent neural network variant that its architecture can deal with the vanishing gradient problem, and thus it can learn long-range dependencies easier [75]. Despite their ability to overcome this issue, LSTMs are less efficient in comparison to GRUs, another category of RNNs, which have widely been proposed in NILM as they can be trained faster or need less training data to generalise [129]. CNN based architectures, is another successful approach that has been used towards capturing long range temporal dependencies in time series but require long filters and very deep structures in order to achieve it. Finally, some alternative approaches such as [187],[164] suggest hybrid architectures which combines recurrent and convolutional structures in order to benefit from the advantages of both methods [79].

The approaches described above have also a great success in sequence-to-sequence (seq2seq) problems such as machine translation, where word sequences are translated from on language to another [164]. Sequence to sequence architectures include two basic components, an encoder and a decoder [159]. Encoders read the input sequence and summarizes the information in internal state vectors encapsulating the information for all input elements. Decoders take as input the encoded information and predicts the desired outcome. Event-less NILM could also be considered as a seq2seq problem as the aggregated sequence is translated through a seq2seq model into the power consumption of a specific domestic appliance.

Another approach that has great success in the area of seq2seq machine translation is graph neural networks and especially graph convolutional networks (GCN). In addition to the RNN-CNN seq2seq approaches, GCN approaches represent the document as a graph that connects relevant contexts regardless of their distances creating a holistic representation of the data [16]. In the context of natural language processing the relations between the graphs could represent adjacency, syntactic dependency, lexical consistency and co-reference [184], [120], [20]. The incorporation of these contextual information injects a semantic bias into sentence encoders achieving better performance in comparison with the linguistic and syntax agnostic approaches [120].

Inspired by the GCN approaches used on machine translation, we aim to attempt modelling the problem of energy disaggregation using graph neural networks. Since we do not have a domain knowledge on energy signals as happens in language, we decided to model our graph using the entropy metric which represents the different operational states of the domestic appliances. Specifically, each node of the graph represents a different operational state while the edges represent the transition probability from one state to another. We hypothesize that using graph encoders that incorporate operational state transitions will lead to more informative representations of the aggregate time series, when used as context vectors by the decoder, especially for the multi-state operational devices like washing machine.

## 4.3 Proposed Methodology

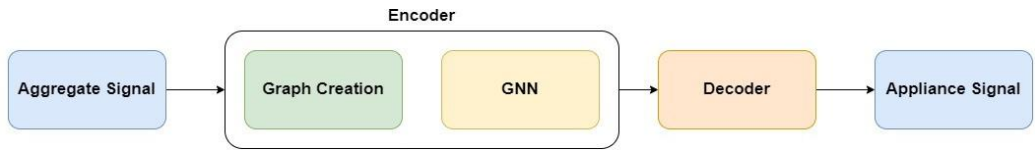

Figure 4.1 High level overview of the proposed seq2seq GCN approach.

### 4.3.1 NILM Problem Formulation

The concept of non-intrusive load monitoring was first introduced by George W. Hart in 1992 [68]. According to their proposed problem formulation, the aggregate active power of a set of measured appliances M at a time t can be formally defined as:

$$x^a_t = \sum_{m=1}^{M} (y^m_i) + \sum_{k=1}^{K} (w^k_t + \varepsilon_t) \tag{4.1}$$

where $y^m_i$ are the contributions of individual appliances m that have been metered, the second sum term represents the contribution of K further appliances that they have not sub-metered while the error term is the noise term originating from the measurement equipment [79]. The goal of energy disaggregation is to solve the inverse problem and determine the individual consumption $y^m_i$ of appliance m based exclusively on the measurement of the aggregate signal.

### 4.3.2 High–level Architecture

Our approach proposes a seq2seq model using a GCN encoder and a Transformer based decoder. The overall process is depicted in Figure 4.1. Inspired by [184], and the application of Graph Neural

Networks on machine translation problem, our approach introduces a seq2seq NILM model which uses GCN as encoders.

The proposed seq2seq model architecture consists of two parts, the encoder and the decoder. First, the aggregated time series windows are clustered based on the entropy of the corresponding appliance consumption. These clusters will represent the nodes of the graph. Using Markov Chain processes, we identify the transition probability between clusters, creating a directed weighted graph representation. Subsequently, using Graph Convolutional Networks and neural message passing we create a latent representation of the initial aggregated time series which incorporates non-sequential dependencies based on the entropy of the appliances' operational states. Finally, the decoder which is a Transformer based model, takes as input the enriched graph embeddings and through a supervised learning process predicts the corresponding appliance's consumption signal. From a data-flow perspective our proposed module can be summarized as:

$$X \dashrightarrow Encoder \dashrightarrow X' \dashrightarrow Decoder \dashrightarrow Y \qquad (4.2)$$

where X is the aggregated signal, X' is the enriched embeddings from the GCN and Y is the corresponding appliance's signal. Although sequence to sequence models are commonly used in NILM [79], this is the first seq2seq approach which models the problem using graphs. Using Graph Convolutional Networks the aggregated signal is encoded in a latent space capturing time invariant relations across the time series.

### 4.3.3 Encoder

The encoder's task is to create an enriched latent representation of the aggregate signal which incorporates time invariant relationships across the whole time series based on the operational state of the disaggregated device. The encoding process, which is depicted in Figure 4.2, involves the modelling of the aggregated signal within of a graph structure and the creation of the enriched graph embeddings.

Specifically, as it is shown in the above, the encoding process starts with the transformation of the aggregated signal into a weighted directed graph and concludes to the creation of the enriched graph embeddings through the neural message passing method. The whole process can be described in the next few subsections.

**Widows Clustering – Node Embeddings**

The aggregated signal is split into smaller windows. In order to create the node embeddings of the graph, these windows are being assigned to specific clusters based on the entropy metric of the corresponding appliance's consumption windows. The cluster boundaries are calculated using Lloyd's algorithm, also known as Voronoi iteration which is an algorithm for estimating evenly spaced sets of points in subsets of Euclidean spaces constructing uniformly sized convex cells [114]. Thus, each cluster contains the aggregated windows of the same operational mode of each appliance and represents the corresponding nodes of the graph. The graph embeddings are initialized as the mean of the assigned aggregated windows.

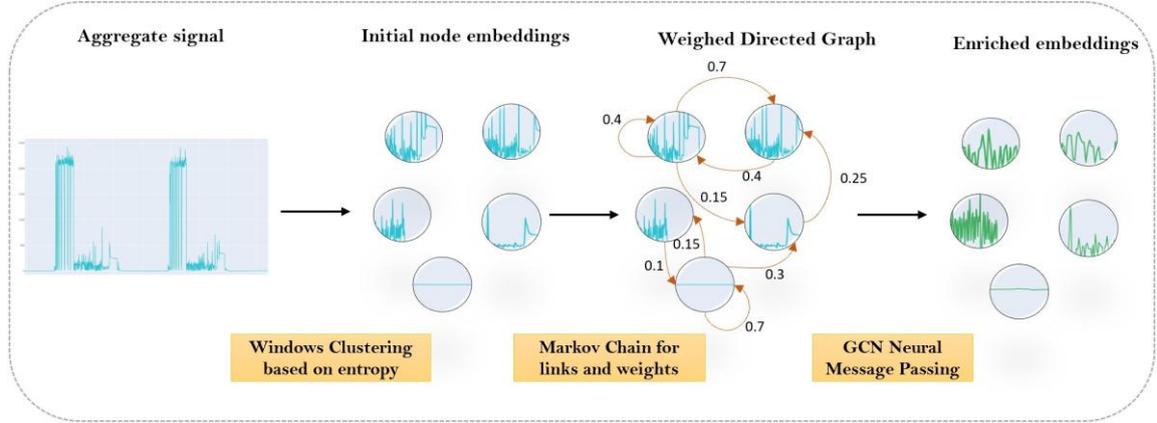

Figure 4.2 Encoder Pipeline

$$E_{c_i} = \frac{1}{N}\sum_{k}^{N}(X_k) \tag{4.3}$$

Where $E_{c_i}$ is the embedding of the ith cluster, $N$ in the total number of the assigned aggregated windows and $X_k$ is the kth aggregated windows that has been assigned in this cluster.

**Markov Chain Processes for links and weights identification**

Following the process described in the previous section, we managed to create the initial graph embeddings which represent the different operational states of the disaggregated appliance. In order to find the links as well as the weights between these nodes we use Markov Chain Processes. A Markov chain or Markov process is a stochastic model describing a sequence of possible events in which the probability of each event depends on the state
attained in the previous event [146]. The matrix describing the Markov chain is called transition matrix and it is considered the most important tool for analyzing Markov chains. The transition matrix is usually given the symbol $P = (p_{ij})$.

The entries of this matrix represent the conditional probability that the next state is $j$ given that the current state is $i$. The transition matrix $P$ lists all the possible states in the state space $S$ and is represented by a square matrix ($N \times N$) where $N$ is the number of the different states on state space $N$. Finally, the rows of $P$ should always sum to 1.

$$\sum_{j=1}^{N}(p_{ij}) = \sum_{j=1}^{N}(P(X_{t+1}=j|X_t=i)) = \sum_{j=1}^{N}(P_{X_t=i}(X_{t+1}=j)) = 1 \tag{4.4}$$

Concluding to the formal definition that let $\{X0, X1, X2, ...\}$ be a Markov chain with state space S with a size of $N$, the transition probabilities of the Markov chain are:

$$p_{ij} = P(X_{t+1} = j | X_t = i)) \; for \; i, j \in S, t = 0, 1, 2, 3, ... \tag{4.5}$$

Transforming this definition to our problem, the $\{X0, X1, X2, ...\}$ values represent the appliances entropy values of the corresponding aggregated windows and the state space S represents the different operational modes of the device as found in the previous step. Thus, by calculating the transition probability matrix of this Markov chain process, we could find the links and their corresponding weights from node to node creating a fully connected- weighted directed graph.

**GCN Neural Message Passing**

The GCN's task is to receive the weighted directed graph from the previous steps and create the enriched embeddings using the neural message passing process . Graph convolutional networks as introduced by Kipf and Welling in 2017, they were shown to be very effective for the node classification task [99]. The number of convolution layers used within a GCN model define their ability to capture information between the neighboring nodes. Specifically, if only one convolution layer is provided the model can capture information only between immediate neighbors.

More formally, considered a graph $G = (V, E)$ where $V$ and $E$ are the set of nodes and edges respectively. Each node $u \in V$ is represented by a feature vector $x_u$. The GCN induces a new representation of the $V$ nodes relying on the representation of their respective neighbors. For a one layer GCN the new node representations are computed using the following equation:

$$h_u = \rho(\sum_{u \in N(u)} (Wx_u + b)) \tag{4.6}$$

where $W$ is a weight matrix, $b$ a bias vector, $\rho$ represents an activation function and $N(u)$ is the set of neighbors of $u$ including itself. Stacking more the 1 GCN layers the previous equation can be generalized in the form of:

$$h_u^{(j+1)} = \rho(\sum_{u \in N(u)} (W^{(j)} x_u^{(j)} + b^{(j)})) \tag{4.7}$$

Our proposed GCN model consists of 8 GCN layers with a ReLu activation function. The last layer uses a sigmoid activation instead. Using the process described above on our constructed graph we create enriched embeddings of the aggregate signal which incorporate time-invariant information from their neighboring nodes.

## 4.3.4 Decoder

The decoder's task is to receive the corresponding aggregated enriched embedding from the GCN and predict the appliance's signal. The architecture of the decoder is a Transformer based model which is consists of 2 Transformer layers, which assign importance weights to the extracted GCN features and a decoding layer, consisted of a 1D deconvolutional layer two linear layers, which generates the desired output.

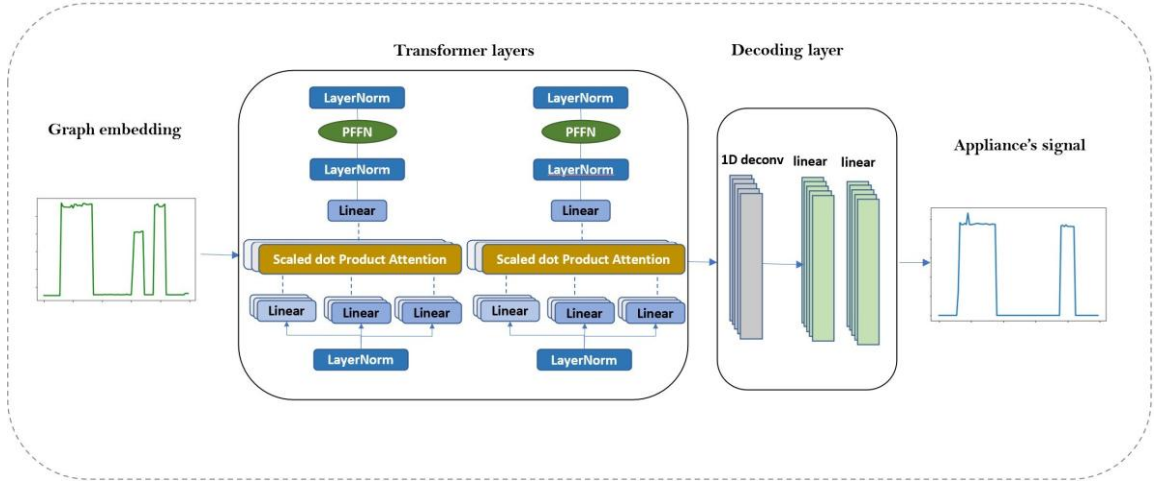

Figure 4.3 Decoder Pipeline

The major advantage of the proposed architecture is that Transformer layers do not process data sequentially. Instead, utilizing attention mechanisms, they assign different weights at each part of the sequence and learn long-range dependencies among the whole aggregated time series [161]. The Transformer layers consists of two major blocks, the multi-head attention module and a position feed-forward networks (PFFN). Transformers implement attention mechanism in a form of a Query-Key-Value model. Specifically, a single-head dot-product attention mechanism applies linear transformations to the input signal to form Q, K and V matrices and then implements a matrix multiplication of these matrices after a scaling and a softmax operation [172].

$$A(Q, K, V) = softmax\left(\frac{QK^T}{\sqrt{d_k}}\right)V \quad (4.8)$$

where $d_k$ is the input length of $K$ matrix and it has the same size with the $d_q$ and $d_v$. The used model deploys a multi-head attention mechanism by extending the single head attention to h dimensions. This can be achieved by concatenating the single head attention outputs, followed by a liner layer.

$$MHA = Concat(A(Q_i, K_i, V_i)) \quad \forall i \in 1,...,h \quad (4.9)$$

PFFN module takes as input the normalized attention scores and performs linear transformations with GELU activation function. PFFN is applies to each position separately and identically meaning that while the linear transformations are the same across different positions, they use different parameters from layer to layer [172].

## 4.4 Experimental Results

### 4.4.1 Dataset description & Experimental setup

The dataset used for the proposed model evaluation is UK-Dale [95]. UK-Dale is one of the most well-known datasets for energy disaggregation, and it includes electricity measurements from 5 houses in the United Kingdom. We focus on low frequency data and multi state operational devices, specifically on washing machine. The aggregate and appliance consumption signals were tested at 1/6 Hz freq and time gaps with less than 3 minutes difference were forward filled. The input signals were not processed to balance their on-off states, as we wanted to use minimal pre-processing and test the performance of our model on real world scenario when the on-off states of the devices are not balanced. The training set signals were split in windows of 480 samples (48 minutes) with a stride of 240.

To validate the performance of our approach, we utilized several state of the art models which process the signals sequentially such as LSTM, GRU and convolutional seq2seq network [88],[87]. Comparing the performance of our time-invariant GCN based approach with these models under the same evaluation set up helped us to better understand the potential of GCN approaches in NILM problem.

### 4.4.2 Training and Performance Evaluation

The metrics used for the performance evaluation of our methodology are the Mean Relative Error (MRE), Mean Absolute Error (MAE), accuracy and F1-score. MRE and MAE are calculated using the ground truth appliance's consumption and estimated signal. These metrics expresses the model's ability to infer the individual appliance consumption levels.

$$MRE = \frac{1}{max(y)} \sum_{i=1}^{n} |\hat{y}_i - y_i|, \quad MAE = \frac{1}{n} \sum_{i=1}^{n} |\hat{y}_i - y_i| \quad (4.10)$$

The accuracy is calculated using the correctly identified timepoints over the sequence length, while F1-score is calculated based on the equation 6.7 .

$$F1 = \frac{TP}{TP + \frac{1}{2}(FP + FN)} \quad (4.11)$$

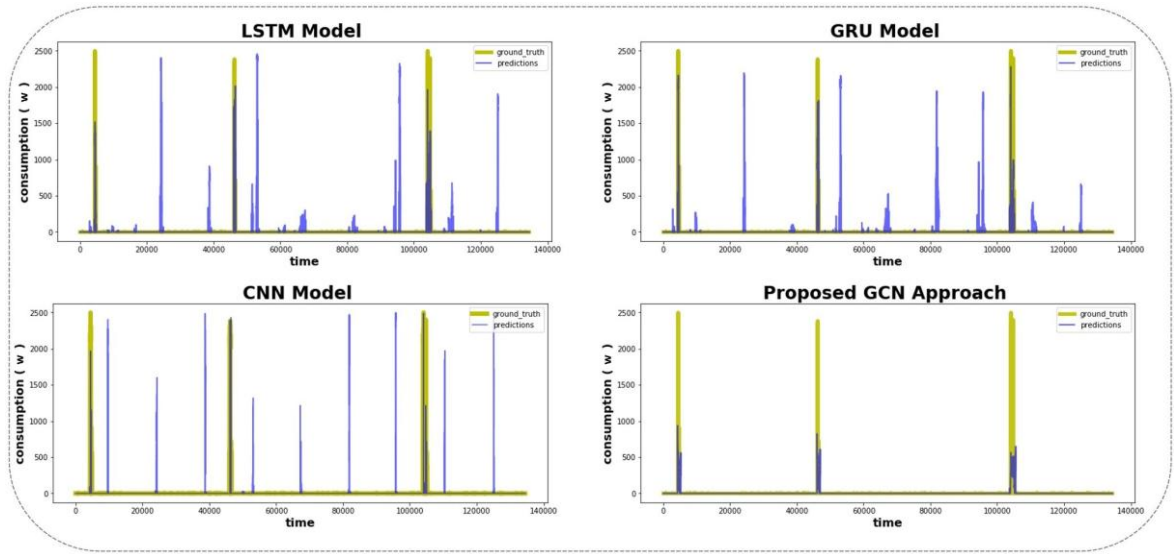

Figure 4.4 Washing machine results, yellow represents ground truth while blue the predicted values.

F1–score expresses the model's ability to detect the appliance activations in imbalanced data and it is considered one of the most important metrics when imbalanced data are provided.

## 4.4.3   Evaluation

Our proposed model was evaluated on different appliances as well as in comparison with the other state-of-the-art methods on the washing machine device.

Table 4.1 GCN Approach results UK Dale

| Appliance | MRE | MAE | Accuracy | F1-Score |
|---|---|---|---|---|
| Kettle | 0.00976 | 28.53128 | 0.99024 | 0.00000 |
| Washer | 0.0209 | 10.43168 | 0.99258 | 0.78053 |
| Microwave | 0.0135 | 6.57303 | 0.98642 | 0.00000 |
| Dishwasher | 0.0465 | 42.21961 | 0.96667 | 0.51522 |

According to the table 1, the GCN approach seems to be more suitable to multi-operational devices such as washing machine and dishwasher. On the other hand, it does not manage to successfully recover the consumption signal of devices like kettle, microwave and fridge where they present shorter duration activations and less operational states.

Table 4.2 Comparative results on washing machine appliance

| Appliance | MRE | MAE | Accuracy | F1-Score |
|---|---|---|---|---|
| CNN [191] | 0.02282 | 15.40796 | 0.98351 | 0.72610 |
| LSTM [139] | 0.05485 | 23.42228 | 0.95041 | 0.22098 |
| GRU [139] | 0.05551 | 21.90249 | 0.95803 | 0.24901 |
| **GCN** | **0.02094** | **10.43168** | **0.99258** | **0.78053** |

Comparing the performance of our approach with the state of the art models on washing machine appliance, the proposed architecture seems to produce comparable or even better results based on the selected evaluation metrics. Specifically, our approach seems to produce significant better F1-score, managing to detect the appliance activations and minimize the false positive predictions, as it is also shown in the comparative diagrams above.

## 4.5 Conclusions

The proposed GCN methodology is the first approach utilizing the graph neural networks in energy disaggregation of multi operational devices, where it seems to outperform the state of the art sequential models without requiring any data balancing. Although the proposed model works well only with multi operational devices, it leads the way and proposes an interesting concept for incorporating more GNN approaches on this domain.

# Chapter 5

# Pre-Training Pruning Strategy for Lightweight Non-Intrusive Load Monitoring at the Edge

## 5.1 Introduction

Non-Intrusive Load Monitoring (NILM), also known as Energy Disaggregation, is an emerging area of interest within the signal processing community primarily due to its role in advancing energy conservation initiatives as well as its interest as a challenging engineering problem [92]. NILM can be perceived as a blind source separation problem, comparable to an instance of Independent Component Analysis (ICA), where the main objective is to decompose the aggregate power signal into its additive appliances'-level signals [91].

Lately, research in NILM has shifted towards employing Deep Neural Network (DNN) methodologies [91, 191, 14, 161], as they have achieved significant state-of-the-art results compared to traditional signal processing approaches [155, 182]. A challenging task for the DNNs is their inherent complexity, necessitating substantial computational resources during both the training and inference phases, making it difficult to deploy these advanced NILM techniques on resource-constrained edge devices [2]. The prevailing solution relies on centralized data processing, which not only elevates the overall cost and energy consumption for running such services, but also raises privacy issues due to the transfer of sensitive data to external sources [160].

Recently, the NILM community has shifted its attention toward strategies aimed at simplifying the computational complexity of deep learning structures in an effort to address the previously mentioned constraints [22, 105, 2, 19]. One widely used method for compressing NILM models is pruning, a deep learning technique that helps identify smaller, more computationally efficient neural networks by removing less important values in the weight tensors of the fully trained models. However, a common drawback of existing compression frameworks is that they are applied after the model has been fully trained [105, 2, 19]. This implies that existing edge-NILM solutions fail to tackle the core problem of reliance on centralized data processing, as the initial training of the model still demands substantial computational resources and infrastructure.

In this work, we present an iterative pre-training pruning strategy to identify an optimal, compressed network in terms of trainable parameters, occurring before the model's training. Adopting this approach, eliminates the needs for extensive training resources and results in streamlined, lightweight models capable of operating on devices with limited computational resources, while at the same time helps in resolving challenges related to centralized data processing, strengthening the practical deployment of NILM. To showcase the effectiveness of the proposed pre-training pruning strategy, we compare to the traditional Magnitude-Based pruning approach [105, 2], all evaluated on the widely used UK-DALE dataset.

In summary, the contributions of this work are as follows: (i) We present a novel pre-training pruning strategy for edge-computed NILM, (ii) We enhance the disaggregation accuracy relative to traditional Magnitude-Based pruning methods, and (iii) We propose a metric based on Euclidean distance to identify the optimal balance between performance degradation and model compression.

## 5.2 Methodology

### 5.2.1 Preliminaries

To explain the principles of NILM let us proceed with some definitions. Assuming that there exist $M$ devices within a network, NILM is defined as

$$P(t) = \sum_{m=1}^{M} P_m(t) + P_{\text{noise}}(t), \tag{5.1}$$

where $t = 1, \ldots, T$ is the time index, $P_m(t)$ is the power of the $m$-th device and $P_{\text{noise}}(t)$ is the unwanted signal power, i.e. the noise. The aim of power disaggregation is to measure $P_m(t)$, given exclusively $P(t)$ so as to isolate the energy usage of each individual appliance $m$. Unarguably, solving (5.1) with traditional algorithms can be challenging, thus research is shifted towards DNNs to transform the problem into a non-convex optimization task. While these networks excel as universal function approximators suitable for NILM challenges, their complexity necessitates a larger neuron count and higher computational resources, making their compression necessary.

### 5.2.2 Baseline Approach: Magnitude-Based Pruning

A typical strategy DNN model compression is magnitude-based pruning (MBP) [66]. Assuming that the weights of the DNN are $w \in \mathbb{R}^N$, magnitude pruning reduces their total number from $N$ to $\hat{N}$ with $N < \hat{N}$, by removing connections with minimal contribution to the network's performance, resulting in a new representation of weights $\hat{w} \in \mathbb{R}^{\hat{N}}$. The identification determining which connections are less important relies on criteria like the $L_1$ or the $L_2$ norm. A wide array of deep-learning pruning techniques exists, which can be generally categorized based on when and how they are applied. Some methods initiate pruning post-training, while others integrate the process iteratively as the network trains [113, 153, 56].

In our study, we employ a post-training, $L_1$-norm-based pruning strategy as a reference point, which is in line with previous edge-computing NILM research [105, 2]. This technique discards the least significant connections in the model, defined by a predetermined sparsity threshold $p_t$. Assuming a dataset $D = \left\{ P(t), \sum_{M=1}^{M} P_m(t) \right\}_{t=1}^{T}$ and a NN with initial weight values $w_0$, the pruning of a NN is formulated as the following constrained optimization problem

$$\min_{w_0 \in \mathbb{R}^N} L(D; w_0)$$
$$\text{s.t.} \quad \|w_0\|_1 \leq p_t \tag{5.2}$$

where $L(\cdot)$ is the defined loss function, n is the total number of structural sets, and $\|\cdot\|_1$ is the standard $L_1$ norm. Therefore, the pruned model would only keep the weights with the highest percentage $1 - p_t$ while the rest will be discarded. Note that this pruning technique is executed post-training, which implies that there is no improvement in computational resource requirements during the initial training phase.

### 5.2.3 Proposed Pre-training pruning scheme

The proposed pre-training pruning methodology addresses the previously noted shortcomings of traditional MBP by implementing the pruning process during the model's initialization phase. The foundational hypothesis underpinning our approach posits that within a randomly initialized neural network exists a sub-network with an initialization configuration that allows it to achieve equivalent test set accuracy to the full network when subjected to the same training iterations [60].

---

**Algorithm 3** Proposed Pre-training pruning scheme

    Initialize a neural network $f(P(t)_{t=1}^T; w_0)$

    **while** $r <= R$ **do**                                                          ▷ R: total number of rounds

- Train the network for 1 epoch, arriving at $w_r$
- Prune $p_t^{1/r}$% of the weights $\hat{w}_r = \mu_r \odot w_r$
- Replace non-zero index values of $\hat{w}_r$ with those of $w_0$

    **end while**
    Fully train the sub-network $f(P(t)_{t=1}^T; \hat{w}_R)$

    $\mu$: Binary mask $\mu_r \in \{0, 1\}^N$
    $\odot$ : Hadamard product

---

Building on this conceptual framework, the proposed method leverages an iterative magnitude pruning technique based on the $L_1$ norm. Specifically, let $f(P(t)_{t=1}^{T}; w_0)$ describe the operation of a DNN with initial weight values $w_0$ during a time window $t = 1, \ldots, T$. The network is trained for 1 round (epoch) and weight values $w_1$ are obtained. Subsequently, the $p_t^{1/r}$ % of smallest magnitude weights are being pruned by applying the corresponding binary mask $\mu_1 \in \{0, 1\}^N$ resulting in $\hat{w}_1 = \mu_1 \odot w_1$.

Once the pruning is completed, non-zero pruned weight values are reset back to their initial ones $w_0$. This process is repeated for $R$ rounds. Note that the pre-training process is a very small proportion compared to the full-training process, i.e. $R << K$, where $K$ corresponds to the full-training time. The identified optimal sub-network $f(P(t)_{t=1}^{T}; \hat{w}_R)$ could then be fully trained, employing much fewer computational resources compared to the original uncompressed model. The above are summarized in Algorithm (3).

## 5.3 Experimental results

### 5.3.1 Experimental setup

**Dataset and Model architecture**

The proposed approach was tested on the UK-DALE dataset [95], which includes aggregate and appliance-level electricity consumption data resampled at 1/6 Hz. The selection of the UK-DALE dataset was mainly due to its widespread use as a benchmark in evaluating NILM application algorithms [92]. The models were trained on houses 1,3,4 and 5 and tested on unseen data from house 2. To assess the efficacy of the proposed methodology, we conducted experiments employing a sequence-to-sequence Convolutional Neural Network (seq2seq CNN) model. It is a computationally efficient network capable of generating whole-window predictions instead of time-point ones, compared to the traditionally used seq2point CNN architecture [191].

**Evaluation Metrics**

To evaluate the model's performance, the following three metrics were adopted: the Mean Absolute Error (MAE) and the Symmetric Mean Absolute Percentage Error (SMAPE) for assessing the energy efficacy of the model as well as the F1 score for assessing 3the model's performance in classifying correctly the 'on/off' states. Moreover, the computational complexity of the model was evaluated using the number of trainable parameters and Multiply-and-Accumulate (MAC) operations. A key shortcoming of previous NILM compression studies is the arbitrary selection of the optimal pruning threshold $\hat{p}_t$ without considering model performance. This paper introduces a metric to address this gap, balancing model complexity with disaggregation performance. The metric accounts for both performance loss in pruned models and gains in parameter reduction. The calculation of optimal pruning threshold $\hat{p}_t$ follows a two-step procedure as follows. First the Euclidean distance for all $p_t$ is calculated given an 'F1 score - MACs' diagram as:

$$\text{dist}(F1, MACs(p_t)) = \sqrt{(1 - F_1)^2 + (0 - MACs(p_t))^2} \quad (5.3)$$

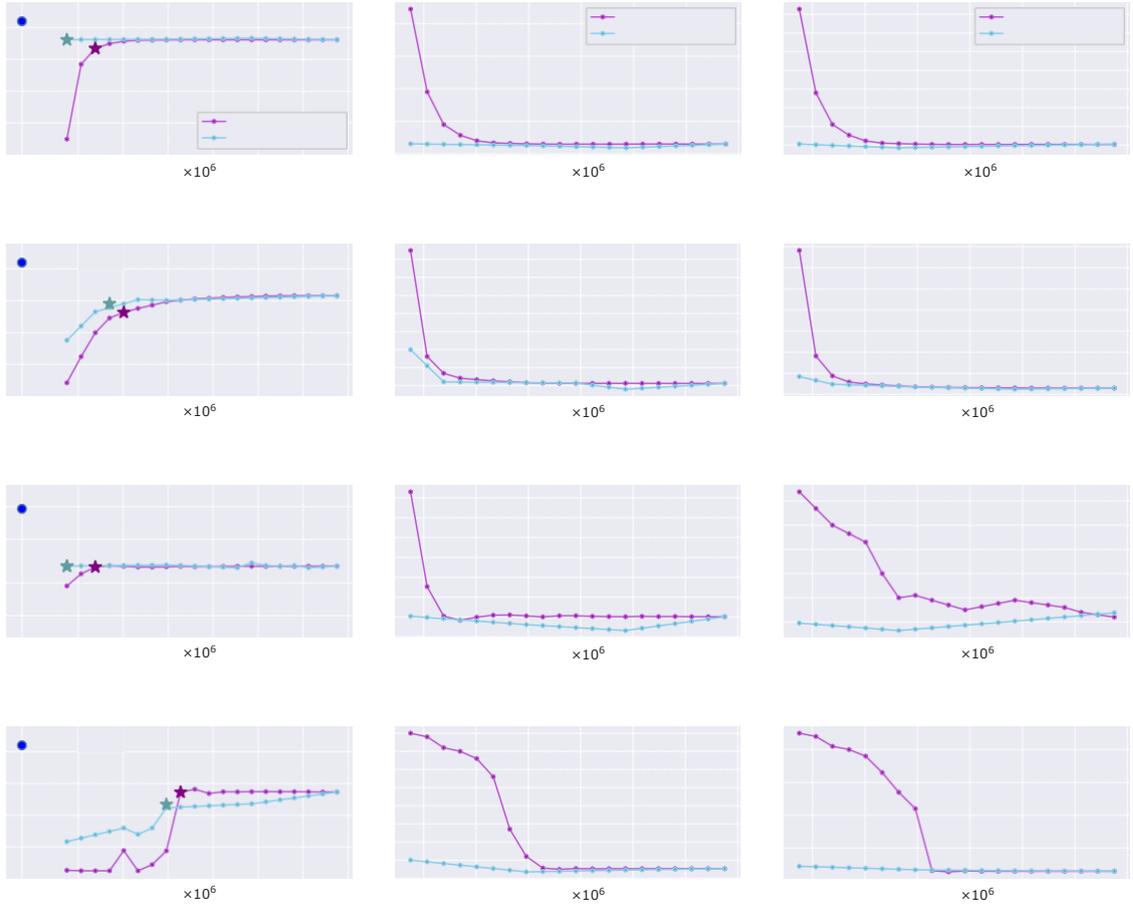

Figure 5.1 Diagrams comparing computational complexity (MACs) for different pruning thresholds (0-95%) to performance degradation (F1, MAE, SMAPE). Stars in the MAC-F1 plots mark the $\hat{p}_t$ for both proposed and after training MBP, as determined by Eq. (5.4) while blue dot represents the 'ideal' point.

Then, $\hat{p}_t$ is selected as the $p_t$ that results in the minimum Euclidean distance among the ones calculated using (5.3), as

$$\hat{p}_t = \arg\min_{p_t \in (0, 0.95)} (\text{dist}(F1, MACs(p_t))). \quad (5.4)$$

### 5.3.2 Evaluation and Findings

Fig. 5.1 provides a side-by-side comparison between the proposed pre-training pruning strategy and the conventional after-training MBP. The figure evaluates both methods using the selected performance metrics for different levels of compression, which is reflected in varying MACs. The 'performance vs compression' curves reveal that the proposed method significantly outshines the conventional one by a large margin, particularly at high sparsity - low MACs levels, observed across all the appliances tested in the study. In the case of appliances like kettles, which have simple consumption signals, the new method hardly affects performance, even when reduced to 7.95 MACs, which is only 5% of the original model's parameters.

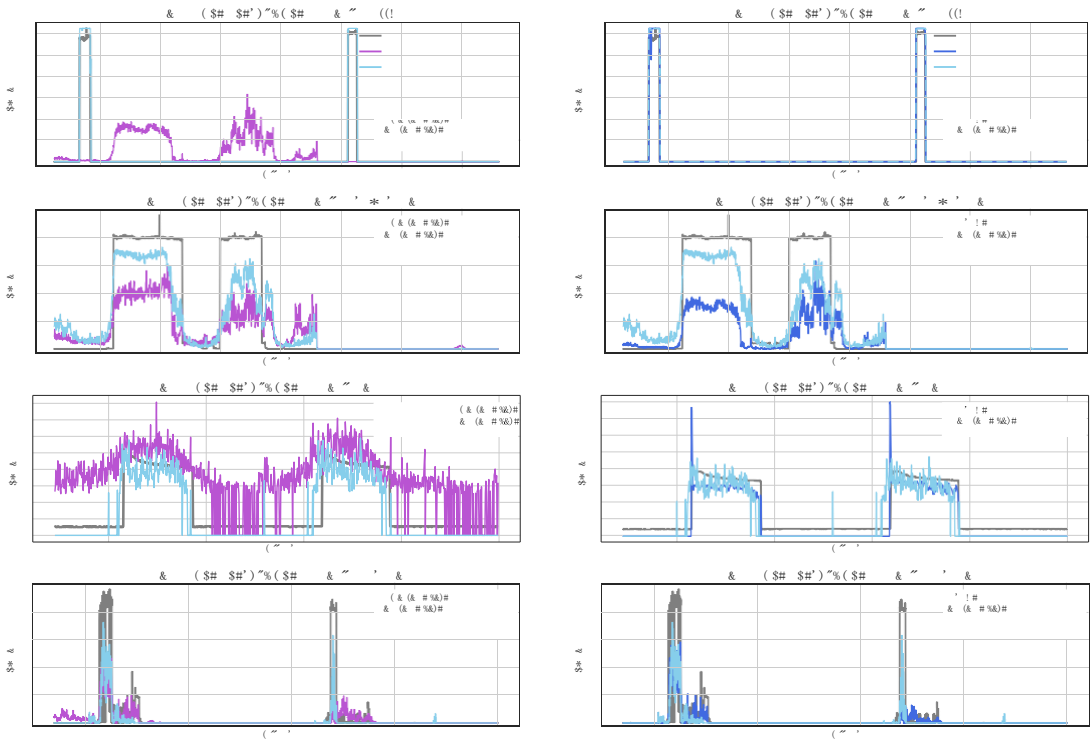

Figure 5.2 Prediction consumption diagrams using the proposed, the after-training MBP scheme and the the baseline model. The pruning thresholds were set equal to the $\hat{p}_t$ of the proposed approach: 95% for the kettle, 80% for the dishwasher, 85% for the fridge and 60% for the washing machine.

As indicated, the proposed approach markedly excels over conventional MBP for all the tested appliances. A similar pattern is observed when comparing the performance of our approach and the baseline model, where the performance of the pruned model frequently matches or even exceeds the baseline model's disaggregation efficacy.

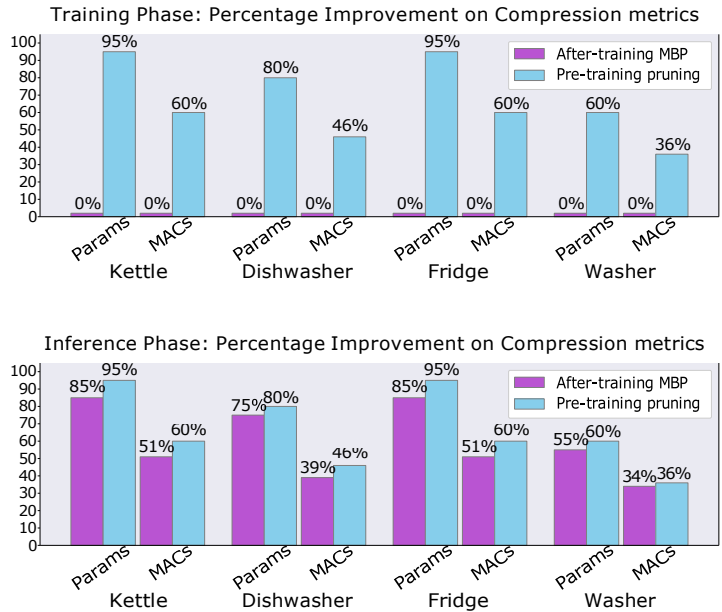

Figure 5.3 Percentage improvement on compression metrics.

The advantages of the proposed methodology are further substantiated by examining the consumption prediction diagrams in Fig. 5.2. The pruning thresholds were set equal to the $\hat{p}_t$ for both after-training MBP and the proposed pre-pruning scheme, in order to evaluate their performance under the same conditions.

Finally, Fig. 5.3 shows that the proposed methodology not only enhances the performance but also the computational efficacy of the NILM models. Specifically for the training phase, our approach adeptly identifies sub-networks that optimize both trainable parameters and MACs, achieving reductions of up to 90% and 60%, whereas the conventional MBP yields no computational improvements. A similar trend emerges in the testing phase, where the proposed approach consistently excels over after-training MBP in both computational metrics.

## 5.4 Conclusions

This work addresses the critical challenges imposed by the resource-intensive nature of deep learning models in NILM. Unlike existing compression techniques, which primarily focus on enhancing computational efficiency only during the deployment phase, the proposed pre-training pruning strategy significantly optimizes both the training and testing phases. Experimental results from the UK-DALE dataset demonstrates that our approach not only reduces computational overhead but also delivers superior disaggregation performance when compared to standard Magnitude-Based pruning schemes promoting the real-world adoption of such solutions.

## Chapter 6

# OPT-NILM: An Iterative Prior-to-Full-Training Pruning Approach for Cost-Effective User-Side Energy Disaggregation at the Edge

## 6.1 Introduction

Electricity load monitoring for appliances is a significant task in light of current economic and ecological trends. It complements home energy management systems (HEMSs) and ambient assisted living (AAL) technologies, contributing to efficient and cost-effective energy management [29], [84]. Additionally, electricity load monitoring serves as a tool for detecting malfunctioning appliances, such as identifying issues like frosting cycles in fridges with damaged seals, among other possibilities. Promoting sustainable living requires householders to adopt energy-related behavior changes. Energy monitoring plays a pivotal role in effective energy management by enabling the monitoring of power consumption of individual appliances, thus informing the planning of technical measures to minimize energy usage. Energy disaggregation techniques can be leveraged to enable granular monitoring of power consumption at the appliance level.

    Non-Intrusive Load Monitoring (NILM) or energy disaggregation algorithms aim to infer the energy consumption patterns of domestic appliances by decomposing the aggregated household energy consumption signal into the individual power signals of its corresponding appliances [90]. Recently, there is a significant number of publications for NILM using deep learning models ([154], [119], [14]). Due to many limitations, NILM approaches have not been widely used in households despite the interest from the industry. Specifically, the training process of such NILM models requires a lot of computational power and resources, so they cannot be deployed on the user side, i.e., on the edge. Instead, they require central servers or cloud computing infrastructures, which increase the cost and

energy of running such a service. The current concept implies data transfer between the data source and a central server, which creates privacy problems and data storage costs [108]. Deploying deep learning algorithms on the edge - at consumers' homes equipped with smart meters and low-power devices - could be a viable solution. In order to make this transition from central data processing to user-side energy disaggregation, many different edge-NILM solutions have been proposed. The main goal of all these solutions is to compress and optimize the models' structures to be able to operate with limited computational resources. One of the most common techniques used for NILM model compression is pruning [105], [19].

Pruning is a technique in deep learning that aids in the development of smaller and more efficient neural networks by eliminating unnecessary values in the final trained models' weight tensors based on their contribution to their predictions. The weights and neurons contributions can be determined by local measures such as their magnitude and L1-norm [171]. However, the existing compression frameworks share the basic limitation that they are being applied to a fully trained model, and *they cannot be executed before full training*. Thus, the proposed edge-NILM solutions do not solve the core issues of central data processing in the sense that the network has to initially be fully trained centrally by allocating all the demands of the central server and cloud computing infrastructures. As a result, current compression schemes only provide a solution to the testing phase of NILM algorithms on the edge, which is the least computationally heavy task of the whole process. However, in the machine learning community, there is an increasing interest in a new training trend according to which we achieve training acceleration that embraces the promising training-on-the-edge paradigm.

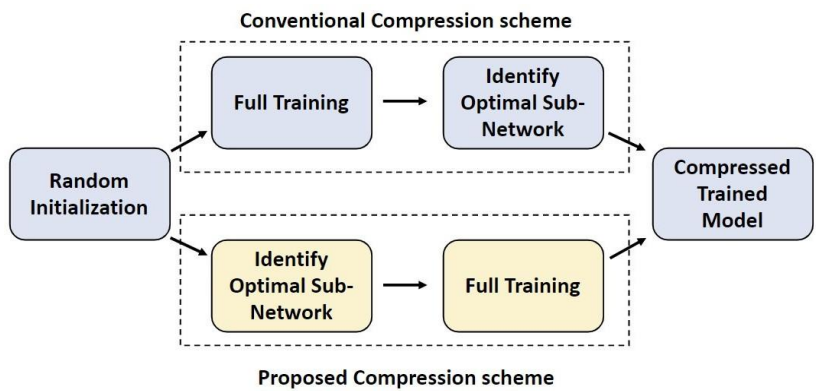

Figure 6.1 The comparison of the conventional pruning process (upper) and the proposed OPT-NILM (lower).

Here, we propose a prior-to-full-training NILM compression scheme, which allows for the identification of optimal sub-deep NILM networks without first requiring full training of the selected model. Following such a scheme would aid in dealing with central data processing issues and NILM real-world deployment since we are able to identify efficient sub-NILM models at their initialization stage, eliminating the training resources and creating efficient, lightweight models that would be able to run in limited resource devices. While the training phase of our framework necessitates data transmission to a central node in order to train the identified sub-network, it's essential to underscore that since the deployment is taking place in houses not included in the training set, the testing phase functions without any subsequent data transmission. All inferencing occurs directly on the edge side, bolstering data privacy and promoting user adoption, given that there's no necessity for users to dispatch their consumption data to an external entity.

These pruned models will be trained using fewer computational resources than the corresponding uncompressed ones and at the same time tested on the edge, utilizing the user's limited resource devices. The main goal of the OPT-NILM is to provide such a framework to optimally identify sub-deep networks before training for a cost-effective user-side NILM. The main contribution of this work is summarized below:

- **Proposing a computationally efficient before-full-training pruning scheme for edge computed NILM**. In contrast with the conventional pruning approaches, the proposed approach identifies optimal sub-deep NILM networks prior to full training. The proposed framework not only identifies sub-deep-neural-network structures that can be easily deployed in a limited resource device, but it also reduces the computational resources needed for the training phase of the NILM models promoting the real-world deployment and adoption of NILM applications.

- **OPT-NILM identifies optimal sub-networks that achieve better disaggregation performance compared to the conventional after-training pruning schemes.** Deep neural networks (DNN) are known to be over-parameterized. Thus, a trained DNN for NILM contains many ineffectual parameters that can be safely pruned or zeroed out with a small or no effect on its performance. In our scheme, where these parameters are pruned before the full training, our sub-deep neural network structures are less overparameterized during the full training, reducing the computational resources needed and preserving a better trade-off between disaggregation performance and reduction in the number of trainable parameters.

- **Proposing a model optimization metric to determine the ideal balance between the model's disaggregation performance and compression.** In NILM applications, the trade-off between accuracy and efficiency is critical. Assuming that we set a high pruning percentage, this results in a significant accuracy drop since the pruned model will not have enough representation power. OPT-NILM is both a resource-efficient and performance-effective technique and introduces an objective model optimization metric for NILM that describes the trade-off between the performance and the model complexity by equally weighting both these factors.

Although the proposed prior-to-full training pruning scheme was inspired by the [60], this work is a pioneering application within the NILM domain. Additionally, this study offers a comprehensive comparison to other compression methods and introduces a novel metric tailored to the unique needs of NILM. From a technical standpoint, the primary contribution of this paper is the introduction of a cost-effective and interoperable deployment strategy for the proposed OPT-NILM inference phase. Our solution is anchored on a Raspberry Pi device and leverages the Z-Wave communication protocol. Originally developed for use in connected home technology, this protocol ensures reliable and robust data transmission between monitored devices and their respective gateways [11]. The paper is structured as follows: Section 2 covers the background on low-frequency NILM with deep learning and compression methods. Section 3 delves into NILM problem formulation and its deep neural network modeling. The proposed solution is detailed in Section 4, while Sections 5 and 6 present and discuss results. Section 7 concludes and outlines future directions.

## 6.2 Related Work

In this section, we provide a brief background on deep learning energy disaggregation approaches and a review of the compression approaches used in deep learning, as well as in deep-NILM models specifically.

### 6.2.1 Deep Learning Models for NILM

Deep learning has achieved enormous success in domains such as natural language processing, time-series analysis, and computer vision [107]. Over the last few years, numerous deep learning approaches have been proposed for NILM as it has been proved that they achieve a superior performance [79], including Convolutional neural networks (CNN), recurrent neural networks (RNN), long short-term memory (LSTM), bidirectional (bi)LSTM, gated recurrent unit (GRU) - (bi)GRU, and Transformer models [90]. RNN approaches, such as LSTM and GRU, use feedback connections to capture temporal dependencies within the power signals [161]. Both LSTM and GRU architectures have been widely proposed in NILM [87], [23] since they converge fast and provide a good disaggregation performance. CNN-based architectures capture long-range temporal dependencies in time-series data, making them a successful NILM technique [193]. This strategy requires large model depth and extensive filters, which increases computational complexity. NILM techniques like [34] propose hybrid recurrent-convolutional architectures, benefiting from the advantages of both types of layers. Transformer-based architectures have become another widespread approach for NILM [161, 189, 176] due to their ability to adopt self-attention mechanism and process data in an order-invariant way. However, all the aforementioned deep learning NILM approaches suffer from computational complexity issues, which increase the training cost and limit their applicability in a real-world deployment on edge.

### 6.2.2 DNN Compression Methods for NILM

Recent developments have driven the adoption of NILM and related energy applications on edge devices. The basic reason for that is that deploying such applications on the edge eliminates the need to transfer data between the users and a central data source, addressing the challenges tied to central data processing and privacy. The landscape of research in edge computed NILM is broad and includes different approaches, from deep learning models on edge devices [19, 41, 191, 2, 105] to feature extraction [163, 162, 185], federated learning [192] and hardware-specific optimizations such as Field-Programmable Gate Arrays (FPGAs) [71] and e-Sense device [63]. Since NILM research has mainly been traversed to deep learning techniques, there is a growing interest in works that deal with NILM inference on edge devices to be deployed as part of Home Energy Management Systems [183]. This trend significantly influenced our decision in this paper to delve deeper into the realm of deploying deep learning architectures on resource-constrained devices and explore the existing and new compression methodologies. However, research on compression methodologies on edge-computed NILM models remains limited. In [19], multiple pruning techniques, including magnitude, relative threshold, and entropy-based pruning, are being investigated and applied on NILM CNN sequence-to-point (seq2point) proposed in [191]. These methods are tested on the kettle and dishwasher appliances from the Refit dataset [126]. The application of a quantization approach has also been proposed in [2]. In this work, the same seq2point CNN architecture is being modified from 32-bit float to 8-bit integer model weights. [105] proposes a model compression scheme of a multi-class seq2point CNN using pruning and tensor decomposition. This approach is evaluated on 3 different appliances from UK-DALE, and REDD datasets [94], [102]. In [160], a performance-aware NILM compression technique is proposed, incorporating an after-pruning approach (PAOP) and an after-pruning approach combined with quantization (PAOPQ) tested across four different architectures. Lastly, in [192], the authors introduced a cloud model compression technique suitable for edge implementation of FedNILM. This was achieved by employing filter pruning within the convolutional layers of the chosen deep-learning model. Although the aforementioned works lead the way toward edge inference in NILM, they provide some significant limitations. The basic drawback of these approaches is that the existing compression schemes are applied to already trained models. Thus, the proposed approaches do not overcome the issues raised by the high computational demand of the training phase and provide a solution only for the testing phase of the NILM models. Another limitation is that [105],[2] and [19] are being employed in a seq2point CNN architecture, which is a computationally inefficient approach since it provides only a midpoint prediction for each window. Since seq2point models are trained to predict the output signal only at the midpoint of the window, they employ a sliding window approach to construct the entire consumption signal, which increases the number of forward passes and, consequently, the computational resources required for inference compared to seq2seq models that predict the entire sequence at once [191]. Finally, in both [19], and [105], compression is applied in an arbitrary way, and there is no framework that evaluates the trade-off between model complexity and performance degradation to define the optimal pruning level.

## 6.3 Problem Formulation

This section presents NILM problem formulation as well as its modeling using deep neural networks. It also discusses some deep learning-related issues that hinder the real word edge deployment of such an application.

### 6.3.1 NILM Problem Formulation

The concept of non-intrusive load monitoring was first introduced by George W. Hart in 1992 [68]. According to their proposed problem formulation, the aggregate active power of a number of measured appliances $m = 1,\ldots,M$ at time $t = 1,\ldots,T$ can be formally defined as:

$$x(t) = \sum_{m=1}^{M} y_m(t) + \varepsilon_{noise}(t), \qquad (6.1)$$

where $y_m(t)$ expresses the power consumption of the $m$-th appliance and $\varepsilon_{noise}(t)$ describes the noise originating from the measurement equipment and the appliances that are not sub-metered during the measurement campaign. [79]. The goal of energy disaggregation is to solve the inverse problem in (6.1) and determine the individual consumption $y_m(t)$ of a selected appliance $m$ at time $t$ based exclusively on the measurement of the aggregate signal $x(t)$.

NILM is considered as a very challenging problem, as power signals do not present any linearity, and the use of each appliance depends on the contextual characteristics of each household. The diverse energy consumption patterns make the implementation of robust NILM algorithms with good generalization behavior even more challenging. Finally, another challenge that NILM models should deal with is the dataset imbalance since every appliance is used with different frequencies and duration.

### 6.3.2 Deep Learning Modelling of NILM

Deep learning for NILM was first introduced in 2015 by Jack Kelly, with major progress on disaggregation performance and generalization capability compared to conventional approaches such as [154], [119]. Solving energy disaggregation using deep neural networks is translated into a non-convex optimization problem. Specifically, learning in deep neural networks describes the process of calculating the weights of the parameters associated with the various regressions throughout the network. In order to find the parameters that give the best approximation, an objective is needed. Assuming a training set of $v = 1, \ldots, V$ values, the objective function $J(\cdot)$ quantifies the distance between the ground truth consumption values, $y_n$, and the predicted ones, $\hat{y}_n$, as:

$$J(\theta) = \frac{1}{2}\sum_{v=1}^{V} L(\hat{y}_v, y_v) \qquad (6.2)$$

where $\theta$ are the model parameters (or weights) and $L(\cdot)$ is the cost function. Note that in (6.2) we omit the subscript $m$ as we describe the optimization function of a single device. The minimization process of $J(\cdot)$ takes place through the back-propagation step [26], where gradient descent is applied to update the parameters of the model. Deep neural networks are universal function approximators that are capable of approximating very complicated functions. However, the trade-off of this capability is the number of neurons needed. Specifically, in order to approximate a non-convex function, as it is needed to do in NILM, which is considered a very challenging problem, it requires to use of high-complexity deep learning models with many parameters [89]. Although these models are considered as state-of-the-art approaches toward NILM, they increase the computational complexity and resources required to tackle this problem.

## 6.4 Methodology

In this part, we describe the suggested OPT-NILM compression strategy as well as the standard after-training magnitude pruning, which has already been employed in [105] [19] as a way to reduce the complexity of NILM deep learning models towards edge inference. In addition, we discuss the methods and benefits of the suggested scheme, highlighting its key contributions to the acceptance and implementation of an edge NILM application in the real world. Lastly, we define a trade-off metric for approximating the optimal pruning threshold in relation to the model's performance.

### 6.4.1 Magnitude Pruning

One of the most common methodologies for optimizing DNN structures is magnitude pruning. The origin of idea of pruning in artificial neural networks derives from synaptic pruning in the human brain, where axons and dendrites decay and die off, resulting in synapse elimination that occurs between early childhood and the onset of puberty [69]. In analogy, deep learning pruning removes redundant parameters or neurons that do not significantly contribute to the model's predictions. Subsequently, model pruning is a technique that reduces the number of the model's weights, $\theta \in \mathrm{R}^K$, to a lower dimensional representation, $\hat{\theta} \in \mathrm{R}^{\hat{K}}$ in which $\hat{K} < K$, by removing non-informative model connections.

Many deep-learning pruning variations have been proposed. Specifically, pruning can either be applied after training or iteratively during the training process [66] [33]. The removal of connections is performed either in an unstructured way by eliminating specific weights from each layer or in a structured one by removing larger structures such as neurons or convolutional filters [104, 109, 113, 153]. Finally, pruning approaches remove weights based on different metrics such as weights magnitude, gradients magnitude, layer-wise mutual information, or learned threshold via gradient descent [17, 113, 56].

In this work, we implement a post-training pruning based on $L_1$-norm metric as a baseline approach since it has also been used for edge computed NILM in [105] [19]. This approach removes the model's connections with the smallest contribution to its output according to a specified threshold $p_{thrs}$. Given a dataset $D = \{(x(t), y_m(t))\}_{t=1}^{T}$ corresponding to a time window $t = 1, \ldots, T$ of measured signal powers and a desired sparsity level $p_{thrs}$ (i.e. the percentage of removed parameters) neural network structural pruning can be formulated as the following constrained optimization problem:

$$\min_{\theta_0 \in \mathbb{R}^N} L(D; \theta_0) \\ \text{s.t.} \quad \|\theta_0\|_1 \leq p_{thrs} \tag{6.3}$$

Here, $L(\cdot)$ is the defined loss function, $\theta_0$ are the initial weight values and $\|\cdot\|_1$ is the standard $L_1$-norm. Thus, after magnitude pruning, the pruned model would only keep the weights with the highest $(1 - p_{thrs})\%$ while the rest will be discarded.

### 6.4.2 OPT-NILM approach

Magnitude pruning removes a percentage of a model's lowest $L_1$-norm connections according to a specified $p_{thrs}$. However, the whole pruning procedure is being applied to an already trained model, meaning that excessive computational resources and data transmission to a central server are required for the training process. The proposed OPT-NILM pruning approach, which deals with the aforementioned limitations, is mainly inspired by the Lottery Ticket Hypothesis paper [60]. According to this work, a randomly-initialized neural network contains a sub-network that is initialized such that - when trained in isolation - it can match the test set accuracy of the original network after training for at most the same number of iterations. The key characteristic of this approach is that pruning is being performed before full training rather than after training, as it is proposed in the existing edge NILM frameworks. Based on this idea, our proposed prior-to-full training pruning technique prunes the NILM networks at the initialization stage. The first step of the proposed approach is to initialize the NILM neural network and train it for a couple of iterations while also keeping track of its initial weights parameters $\theta_0$. In contrast with full training, where the model should become as accurate as possible, in this stage, we are trying to determine which of the initialized parameters lends themselves to the task. In order to achieve this, the model should only be trained for a couple of iterations, which are significantly less compared to the full training. Subsequently, this slightly trained model is pruned using the same techniques that are used to prune a fully trained model. In this work, the L1-norm pruning technique is used to remove the parameters which are not helpful to the task. Since the model is not trained for a long time this technique gives an indication of not only the current parameters but also of their initialization. Thus, if a parameter is currently ineffective, its initialization is probably not part of the optimal sub-network. The final step is to reset the parameters that were not pruned back to their initialization $\theta_0$.

The process of training, pruning, and resetting is repeated for $\hat{N} \ll N$, where $\hat{N}$ stands for the epochs of the pretraining cycle and $N$ stands for the epochs of the full training till the desired pruning level has been achieved. Once the optimal sub-network has been found, this network can be trained fully. Figure 6.2 provides a visual illustration of the process described above.

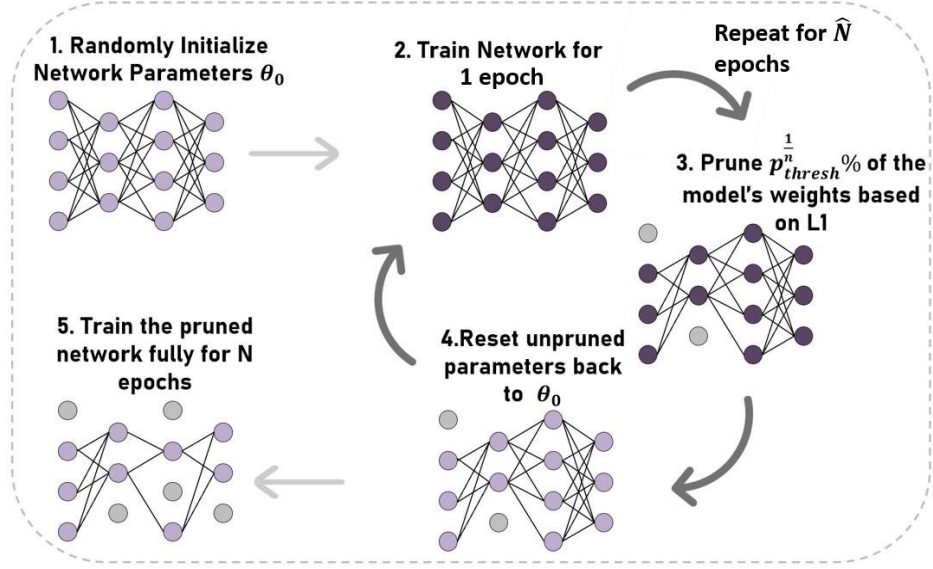

Figure 6.2 Overview of the OPT-NILM pruning scheme. Steps 2,3 and 4 consist the pre-training process of finding the optimal sub-networks, and they are repeated till the desired pruning level has been achieved..

From a more mathematical perspective, let $f(x(t)_{t=1}^T; \theta_0)$ be a deep neural network with initial parameters $\theta_0$. The procedure of the pretraining process is as follows: initially, the network is trained for $\hat{n} = 1, \ldots, \hat{N}$ iterations until the fist desired $\theta_1^{Tr}$ is obtained, where the superscript Tr denotes the training state. This can be described as:

$$\text{Train}: \theta_{\hat{n}}^{Tr} = F_{\text{train}}\ \theta_{\hat{n}-1}^{Rst}\ , \tag{6.4}$$

where $F_{\text{train}}(\cdot)$ is a function describing the training procedure of the network and $\theta_{\hat{n}-1}^{(Rst)}$ are weights obtained from the network after the reset state. Afterwards, $p_{thrs}^{1/\hat{n}}$ % of smallest magnitude weights are being pruned by applying a binary mask $\mu \in \{0,1\}^K$ such that its initialization is $\theta_1^{Pr} = \mu_1 \odot \theta_1^{Tr}$, where $\odot$ denotes the Hadamard (point-wise) multiplication. This is described as:

$$\text{Prune}: \theta_{\hat{n}}^{Pr} = \mu_{\hat{n}} \odot \theta_{\hat{n}}^{Tr}, \tag{6.5}$$

where $\theta_{\hat{n}}^{Pr}$ are weights obtained from the network after pruning. Then, the remaining weights are reset back to $\theta_0$ as

$$\text{Reset}: \theta_{\hat{n}}^{Rst} = F_{\text{rst}}(\theta_0, \theta_{\hat{n}}^{Pr}), \tag{6.6}$$

where $F_{\text{rst}}(\cdot)$ is a function that replaces the non-zero index values of the pruned network with those of $\theta_0$. Note that the above described process is repeated for all the $\hat{N}$ epochs. The identified optimal sub-network $f(x(t)_{t=1}^{T}; \hat{\theta})$ could then be fully trained, employing much fewer computational resources compared to the original uncompressed model. The proposed OPT-NILM pruning scheme is compactly described in the Algorithm (4).

---
**Algorithm 4** OPT-NILM Compression Scheme
---
Initialize a neural network $f(x(t)_{t=1}^{T}; \theta_0)$

**while** $\hat{n} <= \hat{N}$ **do**

- Train the network for 1 epoch to obtain $\theta_{\hat{n}}^{\text{Tr}}$
- Prune $p_{\text{thrs}}^{1/\hat{n}}$ % of the $\theta_{\hat{n}}^{\text{Tr}}$ by creating a binary mask $\mu_{\hat{n}}$
- Reset the remaining weights back to $\theta_0$, $F_{\text{rst}}(\theta_0, \theta_{\hat{n}}^{\text{Pr}})$

**end while**

Fully train the obtained sub-network $f(x(t)_{t=1}^{T}; \hat{\theta})$

---

The proposed pre-training process is able to find optimal computational light sub-networks that could be deployed on a limited resource device and trained using much fewer computational resources, providing a cost-effective embedded NILM solution for the consumers. Furthermore, experimental results show that the proposed OPT-NILM scheme manages to achieve better performance by identifying even smaller sub-deep NILM networks than the conventional pruning scheme. Last but not least, this approach could increase the efficiency and enhance the design of the network by providing information about what an optimal sub-network architecture would look like in terms of layers' importance and the number of initial parameters.

### 6.4.3 Optimal pruning threshold estimation

A basic limitation of the aforementioned works on NILM compression is that $\hat{p}_{\text{thrs}}^{\text{opt}}$ is selected in an arbitrary way without taking into account the performance of the models. This paper proposes a metric that fills this gap and identifies the optimal pruning threshold $\hat{p}_{\text{thrs}}^{\text{opt}}$ for NILM models by equally weighting the trade-off between model complexity and disaggregation performance. This metric incorporates both the performance degradation of the pruned model as well as the gain in terms of parameter reduction. The metric that is being used to find the $\hat{p}_{\text{thrs}}^{\text{opt}}$ is the F1-score as presented in (6.7).

$$F1 = \frac{TP}{TP + \frac{1}{2}(FP + FN)} \qquad (6.7)$$

where TP, FP and FN stand for the True Positive, False Positive and False Negative classified time instances in the predicted signature. The reason that F1-score was the selected measure for evaluating the disaggregation performance of the pruning-performance trade-off metric is its ability to assess if the model can properly identify the appliances' activations and address the class imbalance problem of NILM.

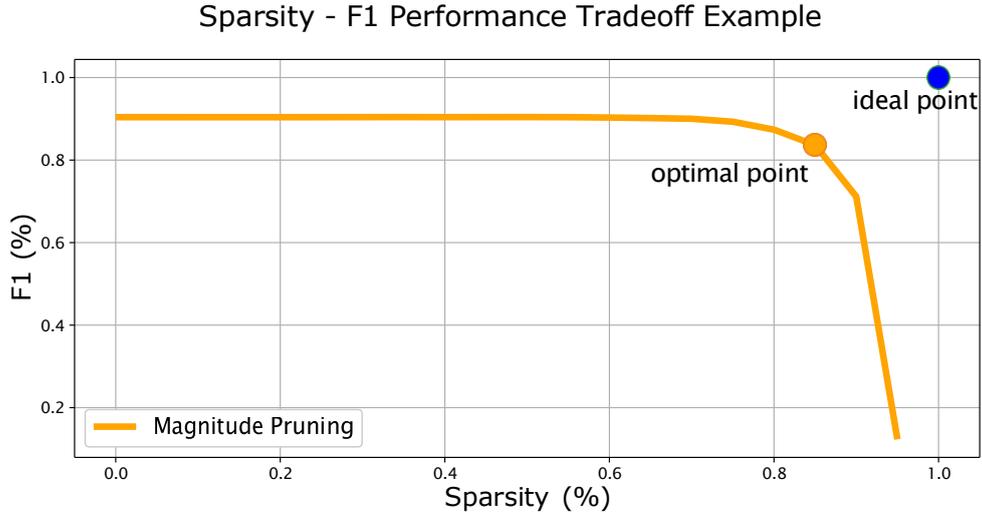

Figure 6.3 Example of the proposed trade-off metric. The blue dot denotes the ideal point (sparsity=1.0, $F_1 = 1$ while the orange dot denotes the optimal point of the performance-sparsity curve.

Pruning results are presented as an achieved performance against the pruning percentage with values $p_{thrs} \in (0, 0.95)$. The optimal point of such a curve is computed as the point that has the minimal Euclidean distance from the 'ideal' points and whose coordinates are F1-score equal to 1 and pruning percentage equal to 1. This metric using:

$$\acute{p}_{thrs}^{opt} = \arg\min_{p_{thrs} \in (0, 0.95)} (\text{dist}(F1, p_{thrs})) \tag{6.8}$$

and

$$\text{dist}(F1, p_i) = \sqrt{(1 - F_1)^2 + (1 - p_i)^2} \tag{6.9}$$

where $p_i \in (0, 0.95)$. A visual representation of the proposed trade-off metric is depicted in Figure 6.3.

Utilizing the performance-sparsity trade-off metric, we are now able to identify the optimal pruning threshold of each pruning technique and use it as a baseline to compare the conventional NILM magnitude pruning with the proposed before-full training NILM pruning scheme. Although different trade-off metrics, such as the performance-sparsity rate of change, could have been used to select the optimal pruning level, the major advantage of the proposed metric is that since the performance and the sparsity axis are in the same scale, it equally weights the performance and the model complexity factors concluding to a fair trade-off metric.

### 6.4.4 Deployment of OPT-NILM to consumer's side

The objective of this paper is to introduce a cutting-edge and cost-effective framework for NILM compression. However, to ensure practical usability and consumer benefits, a deployment scenario is essential. In this regard, we propose a decentralized solution that eliminates data transmission requirements for the inference phase and addresses privacy concerns of the consumers. The developed solution is based on the Z-wave communication protocol, which is ideal for smart home solutions due to its ability to create a mesh network topology, which allows devices to communicate with each other ensuring the reliability and stability of the network as well as better coverage and communication range [108, 186].

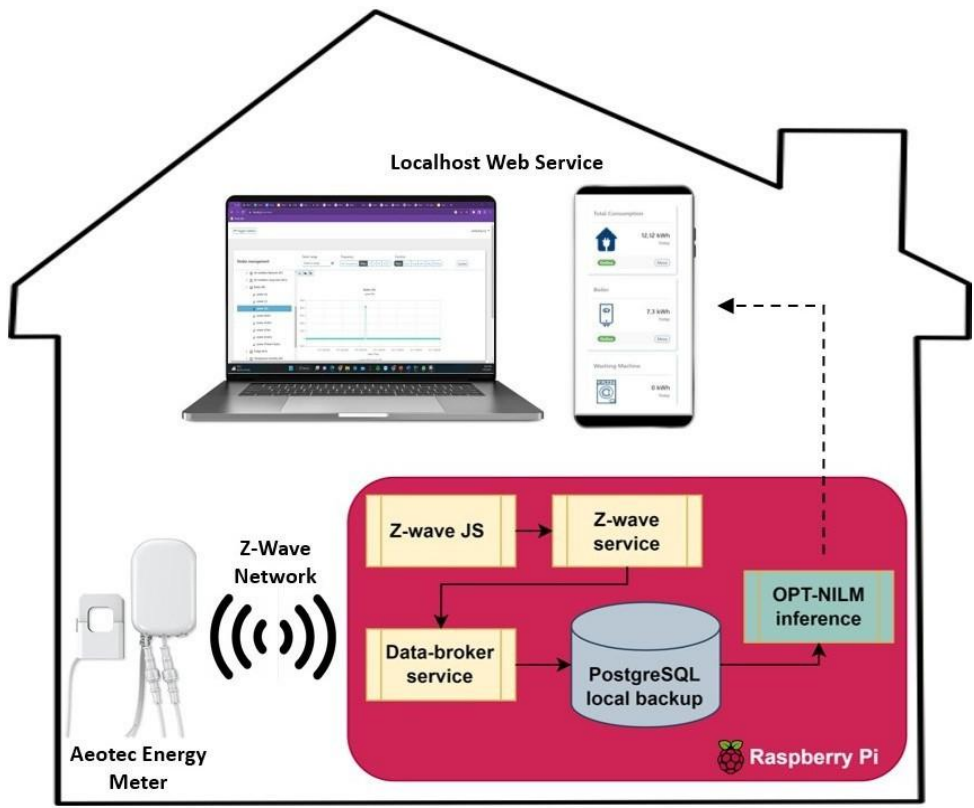

Figure 6.4 Proposed deployment architecture based on Z-wave communication protocol.

To implement our solution, several integral components are employed. We utilize a Z-Wave energy meter, specifically the Aeotec Home Energy Meter Gen 5 [1], which is capable of recording up to 200 amps with an impressive 99% accuracy, in order to monitor and transmit the aggregated consumption data to the OPT-NILM inference service. As a gateway to collect this data and execute the OPT-NILM inference service, the Raspberry Pi Model 4 [143] was used, due to its cost-efficiency, compact design for easy installation, and competency in facilitating Z-Wave communication using the Z-Wave daughter card [200] To ensure users can conveniently access the appliance-level consumption predictions while safeguarding data security, we've set up a local host web service, negating the need for transmitting data externally. A comprehensive visual layout of the proposed OPT-NILM inference deployment strategy is depicted in Figure 6.4. The proposed solution comprises four distinct services all developed and deployed on the edge side. These services are tasked with gathering the aggregate consumption data and producing the disaggregated results.

- **Z-Wave JS:** This is an open-source dockerized service that interfaces with the aggregate consumption smart meter via the Z-Wave protocol. It then transmits the collected data to the Z-Wave service through the MQTT (Message Queuing Telemetry Transport) protocol.

- **Z-Wave service:** This custom service receives the collected data from the Z-Wave JS UI via the MQTT protocol, and subsequently forwards it to the DataBroker service through an API (Application Programming Interface).

- **Data-broker service:** This service is responsible for receiving the data collected by the Z-Wave service and communicates with a local PostgreSQL database. Additionally, the Data-Broker service is tasked with updating (saving and deleting) the collected data in the existing database.

- **OPT NILM inference service:** This service is deployed in a Docker container that runs continuously on the edge device. This service communicates directly with the PostgreSQL database at specified intervals to generate the disaggregation results that they will visualised through the developed localhost web service.

The demonstrated deployment scenario underscores the practical applicability of our OPT-NILM approach, illustrating its real-world operation. This strategy addresses privacy concerns by keeping all data transmissions confined to the user's side, eliminating the need for external exchanges during the whole inference phase.

## 6.5 Experimental Setup

In this section, we give details related to the experimental setup. Specifically, we give a brief description of the dataset, the selected evaluation metrics as well as the seq2seq model architecture that was used to run our experiments and assess the performance of the proposed pruning scheme.

### 6.5.1 Dataset

A publicly available electrical load measurement dataset - UK-DALE [94] was used to showcase the proposed pruning methodology. UK-Dale consists of aggregate consumption and appliance-level energy consumption measurements from five different houses in the United Kingdom.The dataset was built at the sample rate of 1 Hz or one measurement per second for whole-house and 1/6 Hz or one measurement every six seconds for individual appliance consumption.UK-Dale has been widely used for bench-marking NILM algorithms as it is one of the first open-access datasets at this temporal resolution. In this paper, the appliances used to evaluate and test our algorithms include the kettle, the dishwasher, the washing machine, and the fridge due to their high frequency of use, high consumption, and presence in most houses. Furthermore, another reason for selecting these devices is their different consumption patterns, as the kettle provides an on-off consumption signal, the dishwasher and washing machine have different operational states, leading to a more complicated consumption pattern, and the fridge operates continuously. The aggregate signal was resampled to match the frequency of the appliance-level signals at 1/6 Hz. The models were trained using the data from houses 1,3,4 and 5, and they were tested on unseen data from house 2.

## 6.5.2 Model architecture

To evaluate and test the proposed prior-to-full training pruning scheme, we conducted experiments using a seq2seq CNN model. The model's architecture was inspired by the seq2point CNN, which was proposed in [191], and it was also used by the aforementioned NILM compression approaches. The basic reason that we decided to modify this architecture and use a seq2seq model is that seq2point models are less computationally efficient since they produce only one time-point prediction instead of whole windows requiring much more forwards-pass iterations. The proposed model architecture employs 5 1-D convolutional layers with rectified linear activation functions (ReLU) followed by two linear layers with ReLU and Sigmoid activations correspondingly. The CNN architecture is shown in Figure 6.5. The foundational model outlined possesses 22,146,000 trainable parameters and takes up 84 MB of memory. While each model in this study was tailored for a particular appliance, the model's minimal memory footprint posed no issues, especially since it was deployed on a Raspberry Pi 4 with 4GB RAM and a storage capacity of 16 GB. The parameters of the model that were adjusted for optimal training cost include the weights of the convolutional and linear layers of the model architecture described above. Although the proposed pruning technique is designed to be agnostic to specific model architecture, its practical implementation might necessitate some modifications depending on the specific architecture. Our choice of a CNN structure for this work was motivated by the robust compatibility of PyTorch's pruning module with the layers present in our proposed model.

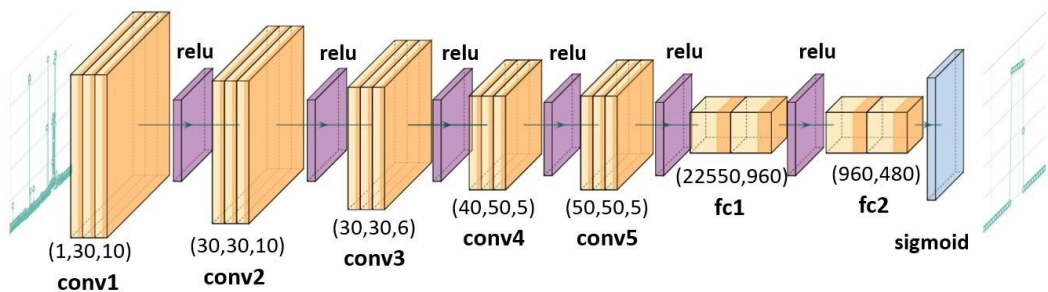

Figure 6.5 The proposed CNN seq2seq architecture. The values in CNN layers represents (in_channels,out_channels,kernel_size) while the values in linear layers represents (in_features,out_features).

### 6.5.3 Evaluation Metrics

We record three widely used metrics to evaluate model performance. Mean Absolute Error (MAE), Symmetric Mean Absolute Percentage Error (SMAPE) equations (6.10) and (6.11), were calculated using the ground truth, $y_t$, and estimated appliance signature, $\hat{y}_t$, providing an evaluation of the NILM model regression performance under a specific time window $t = 1, \ldots, T$ as

$$\text{MAE} = \frac{1}{T} \sum_{t=1}^{T} |\hat{y}_t - y_t| \tag{6.10}$$

and

$$\text{SMAPE} = \frac{2}{T} \sum_{t=1}^{T} \max \frac{|y_t - \hat{y}_t|}{|y_t| + |\hat{y}_t|, \varepsilon} \tag{6.11}$$

Moreover, $F_1$ score (6.7) was also used to assess the model's classification performance. The on-off activations of the appliances were computed by comparing the appliance consumption pattern with the requirements of Table 1. In this study, $F_1$ score is considered the most important metric, as it captures the model's ability to address the class imbalance, identify the appliances' activations and minimize the false positives. This was also the reason that $F_1$ score was selected to be used for the $\hat{p}_{thrs}^{opt}$ calculation.

## 6.6 Results

The conducted experiments presented in this section compare the after-training pruning, which has been used in the previous compression NILM frameworks [105, 19, 160] with the proposed OPT-NILM scheme. The results focus on the performance of each technique as well as on the reduction of the model's trainable parameters. It is worth noting that the OPT-NILM approach requires multiple iterations in order to identify the optimal sub-network, which may seem to extend the cumulative training duration. To delve into details, for the conducted experiments, the identification of the sub-networks took 10 cycles of a single epoch each, amounting to 10% of the full training duration that consisted of 100 epochs. Although this might seem a significant time commitment, the results are compelling. Namely, given that the proposed compression scheme prunes the model's parameters before training, it establishes itself as an efficient NILM compression framework. This is attributed to its dual benefit: it not only produces optimized models tailored for seamless deployment on edge devices with limited resources, but it also mitigates the computational burden during the initial training phase, given that training is executed on the identified sub-optimal model, thereby diminishing computational expenses.

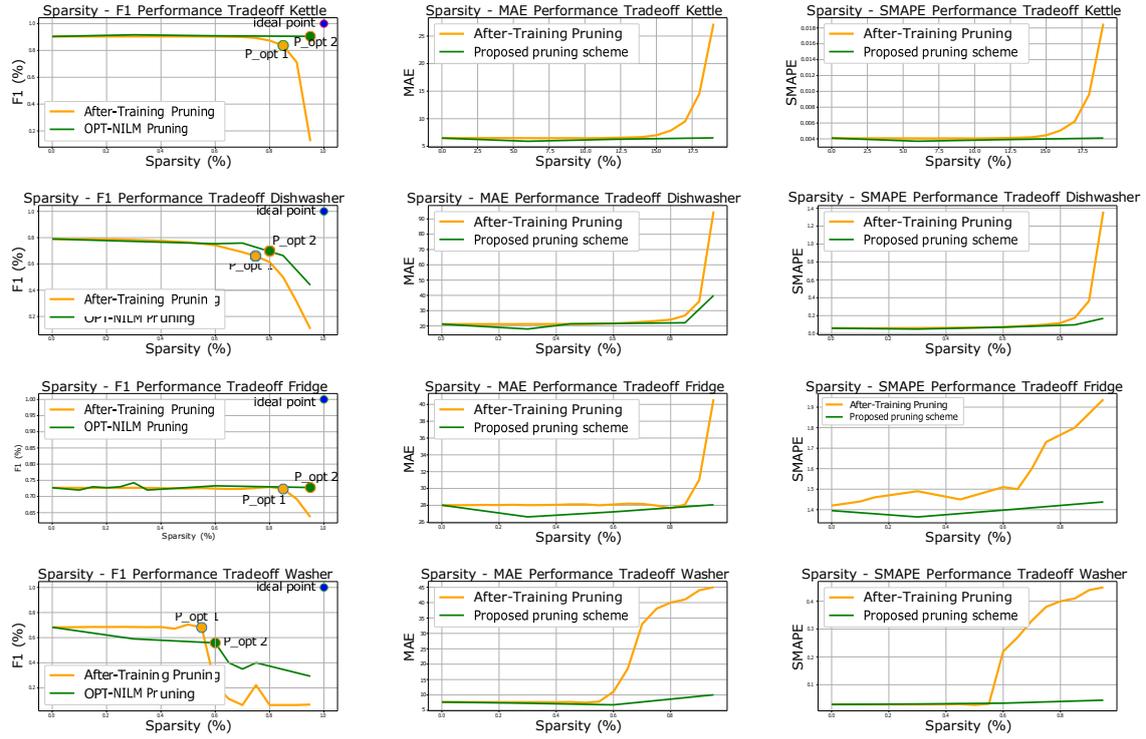

Figure 6.6 Pruning threshold vs Performance degradation diagrams. The blue dot indicates the ideal point, while the green and orange dots represent the optimal points based on the proposed trade-off metric.

As can be seen in Figure 6.6, the performance-pruning level curves indicate that the proposed prior-to-full training pruning can achieve a significantly better disaggregation performance with much fewer trainable parameters than the conventional approach. Specifically, for kettle appliance that only presents an 'on' and an 'off' state, the performance degradation, when using the proposed pruning scheme, is indiscernible even when the model only presents 5% of the initial weights. On the other side, the impact of parameter pruning is more severe on the dishwasher and washing machine, which have a more complicated consumption signal with more operational states. Finally, OPT-NILM showcases its superiority in fridge appliance where it also manages to sustain a better performance-compression trade-off for all the selected evaluation metrics. To sum up, in all of the tested cases, the proposed pruning technique seems to perform significantly better than the conventional after-training pruning since, for the same pruning levels, it manages to achieve significantly higher performance. This assumption could also be confirmed by looking at the consumption prediction diagrams in Figure 6.7, which present the inferred consumption pattern of each appliance for a pruning threshold set to $p_{thresh}^{opt}$ of the OPT-NILM approach and compares them with the baseline and after-training pruning approach.

For the kettle appliance, our proposed pruning scheme showcases a superior disaggregation capability even with the pruning level set to 95%, as it manages to infer the corresponding consumption pattern. On the other hand; conventional magnitude pruning does not manage to detect the kettle's activation function at all, providing a very poor disaggregation performance for the same pruning threshold. Comparing the results of the proposed pruning scheme and the baseline model, we could observe that both prediction curves are very similar to each other even though the pruned model uses only 5% of the parameters of the baseline one. Specifically, the OPT-NILM method surpasses the baseline model, yielding a MAE error of 140 compared to the baseline's MAE error of 153. For the dishwasher appliance, both techniques manage to infer the appliance's consumption pattern. However, the conventional after-full training pruning provides many false positive activations contrary to the proposed technique, which successfully predicts both 'on' and 'off' states.

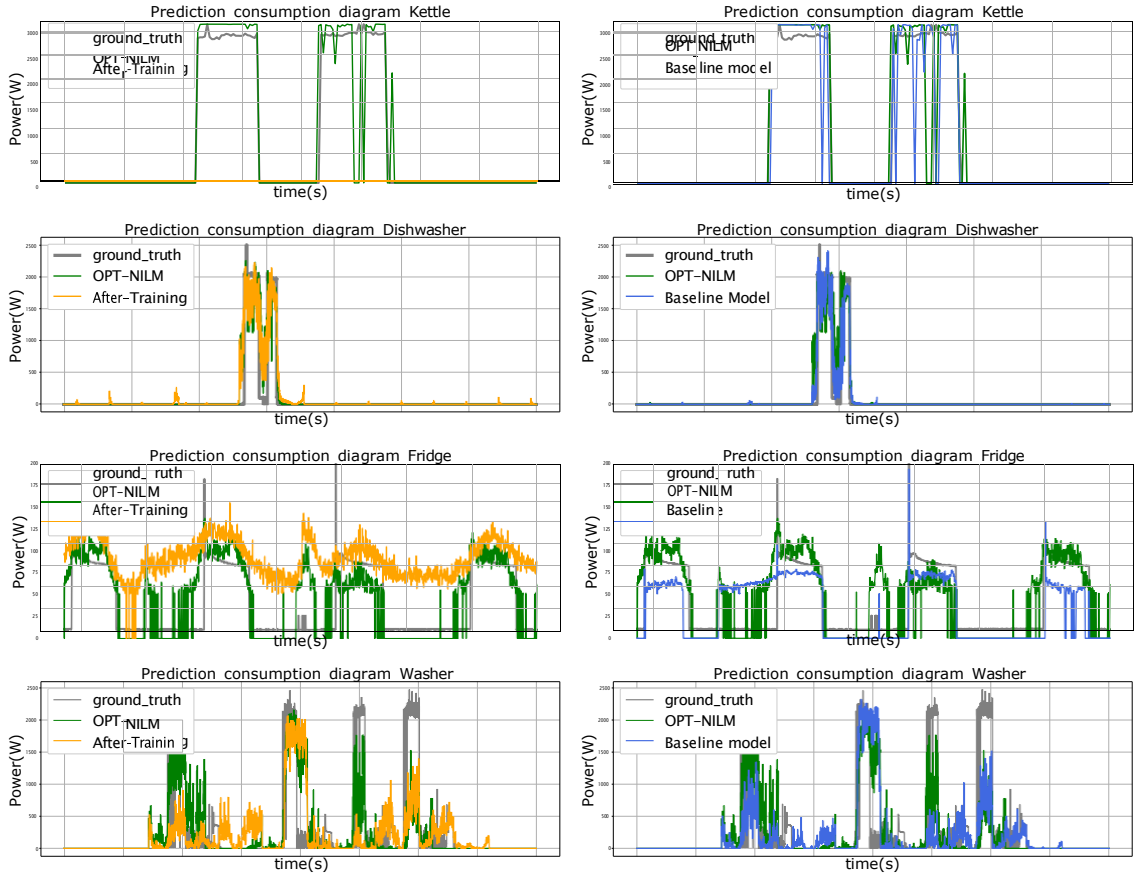

Figure 6.7 Prediction consumption diagrams using the OPT-NILM vs the after-training pruning scheme and the OPT-NILM vs the baseline model. The pruning thresholds were set equal to the $p^{\hat{o}pt}_{thrs}$ of the OPT-NILM approach both for the OPT-NILM and after-training approaches, 95% for the kettle, 80% for the dishwasher, 85% for the fridge and %60 for the washing machine.

Comparing the identified sub-network for the dishwasher appliances between the proposed pruning approach and the baseline model, we observe a similar pattern with the kettle appliance, with prediction curves being very similar to each other even though the pruned network uses only 20% of the baseline's parameters. Notably, the OPT-NILM model achieves a MAE of 41.5, whereas the post-training pruning yields a MAE of 42.4, further demonstrating the former's superior performance. Similar behavior is also observed in the fridge and washing machine appliances, with the OPT-NILM approach managing to perform significantly better than the after-training approach and inferring a consumption pattern very similar to the baseline model for a pruning threshold set to 85% and 60% correspondingly. Based on the prediction consumption diagrams for the washing machine, the OPT-NILM achieved a MAE of 21.7, markedly better than the after-training's 542.1 and the baseline's 22.2. A similar trend was observed for the washer appliance, where the OPT-NILM registered a MAE of 214, surpassing the baseline's 232 and the after-training's 262. The hypothesis that the suggested pruning approach could result in enhanced disaggregation effectiveness identifying more computationally efficient NILM models compared to traditional after-training pruning is also confirmed by looking at the Table 6.1, which presents the disaggregation performance in regard to the model's compression. Specifically, according to this table, the proposed technique achieves a better performance-compression trade-off (i.e. high pruning threshold and low-performance degradation) for all the tested appliances. Overall, the proposed OPT-NILM methodology consistently outperforms the traditional after-training pruning techniques and frequently produces comparable or even better disaggregation results than the baseline model. This enhanced performance is attributed to the fact that the proposed pruning approach identifies an optimal sub-structure within the initial network before the training stage, manifesting an augmented generalization capability on unseen data. This stands in contrast to the baseline model, which, due to potential overparameterization, may incorporate extraneous noise that undermines its performance. Conventional after-training pruning, on the other hand, operates under the assumption that low-magnitude weights are inconsequential and, therefore, dispensable. This assumption, however, is not always correct. Some of these low-magnitude weights remain pivotal to the model's core functionality. Their removal can, hence, significantly impair performance, rendering OPT-NILM a more efficacious alternative.

Table 6.1 Comparative evaluation results - Disaggregation performance with respect to compression threshold

| Appliance | Approach | Compression metrics | | | Performance metrics | | |
|---|---|---|---|---|---|---|---|
| | | Pruning (%) $p = p_{thrs}^{opt}$ | Trainable Params | MFLOPs | F1 | MAE | SMAPE |
| Kettle | Baseline | 0 | 22146000 | 39.27 | 0.90 | 6.49 | 0.004 |
| | After-Training | 85 | 3321900 | 19.07 | 0.84 | 9.81 | 0.006 |
| | OPT-NILM | 95 | 1107300 | 15.86 | 0.90 | 6.58 | 0.004 |
| Dishasher | Baseline | 0 | 22146000 | 39.27 | 0.79 | 21.19 | 0.059 |
| | After-Training | 75 | 5536500 | 23.86 | 0.62 | 25.65 | 0.127 |
| | OPT-NILM | 80 | 4429200 | 21.13 | 0.66 | 22.07 | 0.097 |
| Fridge | Baseline | 0 | 2214600 | 39.27 | 0.72 | 28.01 | 1.39 |
| | After-Training | 85 | 3321900 | 19.09 | 0.68 | 28.31 | 1.78 |
| | OPT-NILM | 95 | 1107300 | 15.86 | 0.71 | 28.30 | 1.46 |
| Washer | Baseline | 0 | 22146000 | 39.27 | 0.68 | 7.69 | 0.029 |
| | After-Training | 55 | 9965700 | 25.01 | 0.67 | 7.85 | 0.031 |
| | OPT-NILM | 60 | 8858400 | 25.89 | 0.60 | 7.51 | 0.030 |

Table 6.2 Percentage Improvement in compression metrics during training and inference phase of the final pruned model.

| Phase | Appliance | Approach | Improvement (%) Compression metrics | |
|---|---|---|---|---|
| | | | Num of Train Params | MFLOPs |
| Training | Kettle | After-Training Pruning | 0 % | 0 % |
| | | OPT-NILM | 95 % | 60% |
| | Dishwasher | After-Training | 0 % | 0 % |
| | | OPT-NILM | 80 % | 46% |
| | Fridge | After-Training | 0 % | 0 % |
| | | OPT-NILM | 95 % | 60 % |
| | Washer | After-Training | 0 % | 0 % |
| | | OPT-NILM | 60 % | 34% |
| Inference | Kettle | After-Training | 85 % | 51 % |
| | | OPT-NILM | 95 % | 60 % |
| | Dishwasher | After-Training | 75 % | 39% |
| | | OPT-NILM | 80 % | 46% |
| | Fridge | After-Training | 85 % | 51% |
| | | OPT-NILM | 95 % | 60 % |
| | Washer | After-Training | 55 % | 36% |
| | | OPT-NILM | 60 % | 34% |

However, despite the improvement in disaggregation performance, the main contribution of the proposed pruning scheme is the fact the model's parameters are removed before the full training of the model. This concludes with a more efficient model initialization since the identified sub-network would need much fewer computational resources to be fully trained. Thus, the model's training cost and computational resources will be dramatically reduced, promoting the real-world deployment and adoption of such a system. The reduction in the model's complexity is evaluated using the number of trainable parameters as well as the number of floating point operations (FLOPs) required to perform a forward pass. In order to highlight the contribution of the proposed technique, we evaluate the complexity of the pruned model both before the full training and testing phase.

Table 6.2 indicates that the proposed pruning method leads to a noteworthy enhancement in computational efficiency. Specifically, the optimal sub-network for the kettle appliance retains just 5% of the initial number of trainable parameters, while the one for the dishwasher appliance retains 20% of the initial number of trainable parameters. For the fridge appliance, the optimal sub-network retains 5% of the initial parameters, and for the washer appliance, it retains 40% of the original model parameters. Similar behavior is also observed in FLOPs parameters, where they also present a significant drop. On the contrary, the conventional magnitude pruning approach does not improve the computational efficiency of the model during the training phase nor on FLOPs or model parameters. Evaluating the computational complexity of the pruned NILM models for the testing phase, we observe that the proposed pruning technique is also superior in comparison to the standard after-training pruning. In terms of both the number of trainable parameters and FLOPs, the proposed pruning scheme seems to identify more computationally efficient networks that would be able to be deployed in a limited resource device and produce better disaggregation performance.

Table 6.3 Comparative results with other works among all tested appliances

| Approach | Percentage Change Compression | | Percentage Change Performance | | |
|---|---|---|---|---|---|
| | Trainable parameters | MFLOPS | F1 | MAE | SMAPE |
| Edge-NILM[105] | -75% | -44% | -8.8% | 18% | 27% |
| PAOP[160] | -37.5% | -25% | -7.01% | 8.3% | 15.2% |
| PAOPQ[160] | -27.5% | -18.1% | -17.7% | 6.5% | 16.2% |
| OPT-NILM | -82.5% | -49.8% | -7.4% | 1.05% | 18% |

The superiority of our approach in terms of both computational complexity and disaggregation performance is also demonstrated by comparing OPT-NILM's overall performance across all tested appliances against two other works,[105] and [160] which employ after-training compression techniques on the same model architecture. The comparative results presented in Table 6.3 indicate that our approach achieves a better trade-off between compression and disaggregation performance, surpassing the capabilities of current edge NILM solutions and offering a more dependable and computationally effective framework for potential consumers.

## 6.7 Conclusions and Future Work

In this work, we have proposed an efficient prior-to-full training pruning scheme for edge deployment of NILM that produces significantly better results than the conventional after-training pruning approach and reduces the computational resources for both the training and testing phase. The proposed pruning scheme not only identifies sub-optimal networks with better disaggregation performance but also assumes a cost-effective NILM deployment since the sub-network structures are identified before the training phase. Finally, we also introduced a trade-off metric to identify the optimal pruning threshold of a NILM model and use it to define a comparable ground between the proposed pruning scheme and the ones that have been used in past edge NILM research works. The experimental findings confirm that the proposed methodology outperforms conventional after-training pruning techniques, not only in terms of disaggregation performance but also in eliminating the computational costs of both training and testing phases, providing a framework for a cost-effective, secure and reliable embedded solution with high potential for the consumer's side. Additionally, OPT-NILM demonstrates an overall superior trade-off between disaggregation performance and compression when compared to other works, further underscoring the effectiveness of the approach. Therefore, the proposed solution presents a cutting-edge approach to edge-based NILM area that holds significant promise for real-world deployment and provides numerous advantages for consumers.

In our future research, we plan to explore additional pruning techniques, such as gradient-based magnitude pruning and information-based pruning, along with evaluating the efficacy of structured pruning.Additionally, we also plan to utilize the versatility of the developed pruning scheme and extend it to other architectures prominent in the NILM domain, like Transformers, LSTM and GRU. Finally, we aim to deploy our solution in real-world settings at a larger scale to assess the replicability of our simulation experiments under real-world conditions.

# Part IV

# Advanced Compression Techniques for NILM Architectures Evaluated in Mediterranean Scenarios

# Chapter 7

# Towards Edge-Computed NILM: Insights from a Mediterranean Use Case Using a Structured Pruning Approach

## 7.1 Introduction

To address climate change, the European Commission has implemented policy initiatives aimed at achieving climate neutrality by 2050, with an intermediate target of reducing net greenhouse gas emissions by 55% by 2030 compared to 1990 levels [42]. Central to this objective is enhancing energy efficiency and optimizing energy consumption within buildings, which are responsible for approximately 37% of global energy-related CO2 emissions [42]. Advanced metering infrastructure plays a pivotal role in this effort by enabling bidirectional communication between utilities and consumers, thereby facilitating the remote management of electricity usage [11]. With smart meter deployment anticipated to reach 80% of European consumers by 2025 [42], the implementation of detailed energy monitoring and techniques becomes essential. These techniques support real-time energy management, identification of malfunctioning appliances, and promote more efficient participation in sustainable energy practices, such as demand response schemes, targeted energy feedback, and tailored pricing policies.

An increasingly popular technique is non-intrusive-load-monitoring (NILM) or energy disaggregation, which identifies the ON–OFF states of appliances and estimates their power consumption based solely on the building's total meter readings. Recently, NILM research has focused on using Deep Neural Networks (DNNs) [191, 14, 161], which have achieved impressive results compared to methods based on traditional signal processing and statistics [155, 182]. However, the complexity of DNNs necessitates a centralized data processing scheme due to their high computational demands for both the training and inference phases, leading to increased costs and privacy concerns due to the transfer of sensitive data to external servers. To address these issues and facilitate the shift from centralized data processing to decentralized approaches on edge, the NILM community is now focusing on reducing the computational complexity of deep learning models using various compression techniques.[22, 105, 2, 19, 15].

Another important point about NILM is that it is inherently a context-aware problem, as the energy consumption patterns of appliances depend on various external and internal factors. These include environmental conditions like weather and seasonality and appliance-specific characteristics such as operational modes and technology types. Although extensive research has been conducted in regions like Northern Europe, the UK, and the USA, there is a lack of research focused on the Mediterranean region [9, 94, 102, 25, 125]. This region presents unique environmental conditions and appliance usage patterns, such as air conditioners for both cooling and heating and electric boilers, constituting significant and flexible loads [9]. This gap highlights the need for NILM approaches tailored to the Mediterranean context to improve accuracy and unlock new opportunities for smart grid integration and flexibility.

To address these gaps, our work proposes an edge-based NILM approach that leverages both unstructured and structured pruning techniques. In addition, we utilize the Plegma dataset [9], a newly established dataset from the Mediterranean area, to develop and evaluate our methods. Our approach aims to reduce the computational complexity of deep learning models, making them suitable for deployment on resource-constrained edge devices while maintaining high accuracy. By focusing on the specific conditions and appliance usage patterns of the Mediterranean region, we aim to provide a robust solution for efficient energy management and enhanced smart grid integration. In summary, the contributions of this work are as follows: (i) Establishment of a comprehensive benchmark for the Plegma dataset and NILM within the Mediterranean context; (ii) Introduction of an optimized structured pruning methodology that exploits unstructured one to identify the optimal sparsity ratio per layer for removing whole units; (iii) Development and evaluation of an edge-based NILM approach utilizing both unstructured and optimized structured magnitude pruning, providing a comparative analysis of their respective strengths and limitations to assess the practical applicability of decentralized NILM approaches in real-world scenarios.

## 7.2 Methodology

### 7.2.1 NILM Problem Formulation

Non-intrusive load monitoring (NILM), initially introduced in [68], involves estimating the appliance level consumption by exclusively relying on the aggregate active power. Formally, for a number of $m = 1, \ldots, M$ appliances and a fixed time window $t = 1, \ldots, T$, NILM is defined as

$$x(t) = \sum_{m=1}^{M} y_m(t) + \varepsilon(t), \tag{7.1}$$

where $y_m(t)$ denotes the power consumption of the $m$-th appliance, $\varepsilon(t)$ denotes the noise signal captured from measurement instruments and from appliances not individually metered during the data collection process [79], and $x(t)$ is the aggregated signal. The aim of energy disaggregation is to estimate the power usage of each appliance $y_m(t)$ at any given time $t$ by relying solely on the aggregated power data $x(t)$. Solving NILM problems with traditional algorithms is often difficult, leading researchers to adopt deep neural networks (DNNs) for this task. While DNNs are effective universal function approximators for NILM, their complexity requires numerous neurons and substantial computational power, highlighting the need for compression.

### 7.2.2 Pruning Strategies

DL models used to approximate eq. (7.1) can be complex, imposing challenges to the computational resources required for their real-time deployment on edge. To tackle such issues, deep neural network (DNN) compression methods are used, with pruning being a popular approach. Pruning entails removing individual weights or entire units, such as filters or channels, based on a criterion like the L1 norm. The removal of individual weights is referred to as *unstructured pruning*, whereas the removal of entire units is known as *structured pruning* Each process is described as follows:

**Unstructured pruning:** For a DNN model with weights represented in an ordered set $\mathbf{W} \in \mathbb{R}^N$, and a sparsity ratio $s$ denoting the proportion of pruned parameters over the total, unstructured pruning sets individual weight values $w_{i,j}$, with connection from $i$ to $j$, to 0 if $||w_{i,j}||_1 < s$, where $||\cdot||_1$ is the $L_1$ norm. As a result, the pruned weights $\mathbf{W}' \in \mathbb{R}^N$ lead to a reduced model trainable parameter size. However, since the weights are replaced with zeros rather than removed, to fully realize compression benefits, we need hardware specifically designed for sparse matrix operations, which increases the cost and complexity of the real-world deployment of such a solution [158]. The number of pruned weights is determined by the sparsity ratio and directly impacts the classification performance of the model; the higher the sparsity ratio, the larger the computational performance drop.

**Structured pruning:** In contrast to the above method, structure pruning defines the sparsity ratio $s$ as the proportion of pruned units like filters, neurons, or channels over the total ones. Specifically, for a a DNN with $C \in \mathbb{N}^k$ channels, the structured pruning process sets $C$ channels to 0 if $||C||_1 < s$. The resulting number of channels, $C' \in \mathbb{N}^{k'}$ not only results in reduced model parameter size but also in reduced computational resources as floating point operations are removed entirely, constituting it a more hardware-efficient and straightforward deployment approach.

### 7.2.3 Proposed Optimized Structured Pruning

Both methods discussed in the previous subsections have their own advantages and disadvantages when real-time edge deployment is considered. Specifically, unstructured pruning makes the weight matrix a sparse one by zeroing individual weights but keeping the floating point operations number the same. On the contrary, structured pruning results in a computationally efficient structure by removing entire units, which can result in significant performance degradation if not applied judiciously.

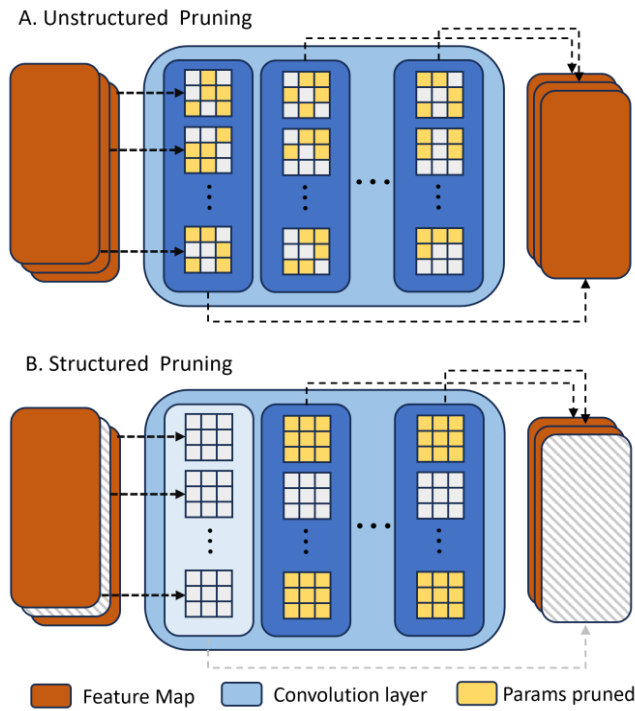

Figure 7.1 Difference between unstructured (top) and structured (bottom) pruning. Unstructured pruning removes individual weights, whereas structured pruning removes whole units (filters and channels).

This work combines the strengths of both approaches by determining the optimal sparsity ratio distribution across different layers in structured pruning of DL models derived from an unstructured approach. The motivation for this work stems from the arbitrary selection of the sparsity ratio of each layer when structured pruning is applied, which often leads to the removal of filters or channels containing crucial features, consequently causing unwanted performance degradation. To tackle this issue, we leverage unstructured pruning to identify the percentage of parameters that should be pruned in each layer based on the desired sparsity ratio. Then, based on this percentage value, we apply structured pruning in each layer to remove an equal number of units, i.e., filters or channels. To give a better intuition behind this, consider the following example. Assume a simple 1D CNN model containing 3 layers, 2 convolutional ones, and 1 linear, along with a desired sparsity ratio of 50%. Using unstructured pruning, an optimal distribution of weights that should be removed can be 15%/, 15%, 30% for the 2 convolutional and the linear layers, respectively. Now that the optimal sparsity ratio per layer is *known*, structured pruning is employed to remove whole units in each layer equal to the percentages identified from the unstructured pruning, i.e., 15%, 15%, 30% for the 2 convolutional and the linear layers respectively. As a result, structured pruning is guided in a sense by the unstructured one to keep filters and channels with the most important features instead of removing them arbitrarily.

## 7.3 Experimental Results

### 7.3.1 Experimental setup: Dataset & Model architecture

The experiments for both the baseline, which is the unpruned trained model and the pruned models were conducted using the Plegma dataset. [9], which provides total and appliance-level electricity consumption data at 10-second intervals. This study focuses on devices commonly used in the Mediterranean region, such as boilers and air conditioners, which are not typically included in other NILM datasets, and thus, their results are presented. The models were trained on data from houses 1-13, excluding house 2, which was used as unseen test data. The chosen model was a sequence-to-sequence 1D Convolutional Neural Network known for its capabilities in edge-NILM [13]. This model offers better computational efficiency than the traditional seq2point CNN architecture, as it generates predictions for entire windows rather than individual time points, thus requiring fewer forward passes.

### 7.3.2 Evaluation metrics

To evaluate the model's performance, the following three metrics were adopted: Mean Absolute Error (MAE), Mean Relational Error (MRE), and the F1 score. MAE and MRE were used to assess the regression performance of the model and its ability to predict the consumption signature of each of the tested appliances, while the F1 score was used to evaluate the model's performance in correctly classifying their 'on/off' states. Additionally, the computational complexity of the model was evaluated using the number of trainable parameters and Floating Point Operations (FLOPs). They measure the number of arithmetic operations required to execute the model, providing insight into its computational efficiency.

To identify the optimal pruning sparsity ratio $\hat{s}$ for each model, we consider a two-step procedure [13]. First, we calculate the Euclidean distance for each one of the selected sparsity ratio percentages with values $s = 0\%, 5\%, \ldots, 90\%, 95\%$ with respect to the 'ideal point' which is that of $s = 100\%$ and $F1 = 100\%$ (Eq. 7.2). Then, we select $\hat{s}$ to be equal to the $s$, resulting in the minimum Euclidean distance as shown in Eq. 7.3. Note that the above metric used accounts for both performance loss in pruned models and savings in parameter reduction.

$$\text{dist}(F1, s) = \sqrt{(100 - F_1)^2 + (100 - s)^2} \tag{7.2}$$

$$\hat{s} = \arg\min_{s \in (0, 0.95)} (\text{dist}(F1, s)) \tag{7.3}$$

### 7.3.3 Evaluation and Insights

A comparative analysis of unstructured and optimized structured magnitude-based pruning methods is shown in Fig. 7.2. Specifically, it evaluates both pruning approaches using the chosen performance metrics across various compression levels ranging from 0-95% sparsity. The 'sparsity vs performance' curves suggest that unstructured pruning achieves higher performance compared to optimized structured pruning when large sparsity levels are selected for all tested appliances. For the boiler, which exhibits a simple consumption signal and low variation across different boiler types, the performance gap between the two pruning methods is smaller. Both methods effectively identify sparser networks, achieving trainable parameter reductions of 75% and 60%, respectively. In contrast, for the air conditioner (A/C), which has a highly variable consumption signal due to the diversity in device types, wattages, and operational modes, the performance gap is more pronounced. Specifically, optimized structured pruning is only able to reduce model parameters by 15%, highlighting its limitations compared to unstructured pruning, which achieves a reduction of 75% in handling devices with complex consumption patterns.

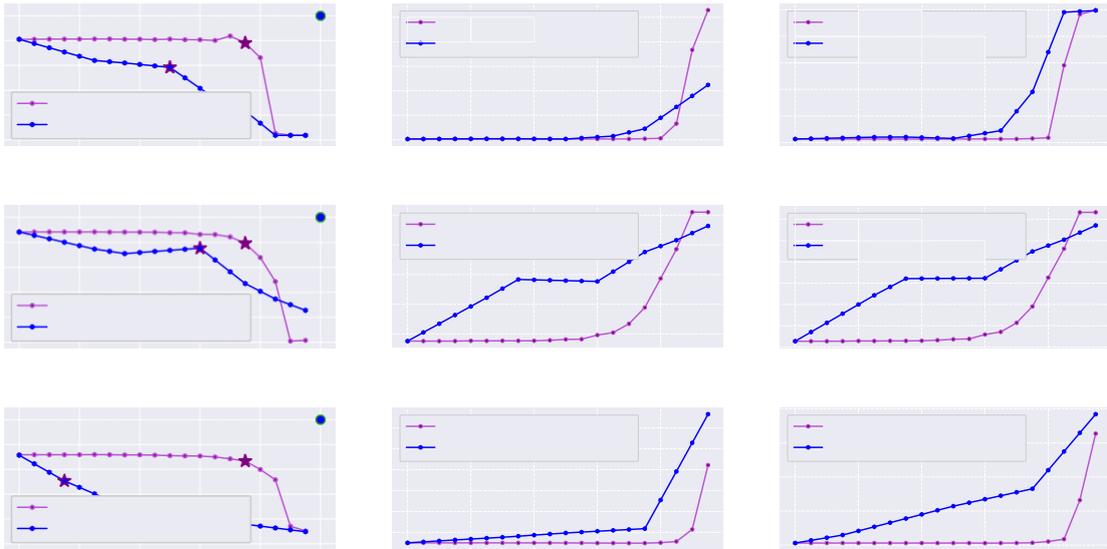

Figure 7.2 Comparison of computational complexity (sparsity %) for different pruning thresholds (0-95%) to performance degradation (F1, MAE, MRE). Stars in the Sparsity-F1 plots mark the optimal pruning ratio $\hat{s}$ for both optimized structured and unstructured magnitude-based pruning, while the blue dot represents the 'ideal' point.

Similar insights are derived from the analysis of the consumption prediction diagrams presented in Figure 7.3. The results indicate that unstructured pruning provides better disaggregation performance compared to optimized structured pruning across all tested appliances. Nevertheless, both pruning methods effectively identify device activations and accurately classify their on-off states.

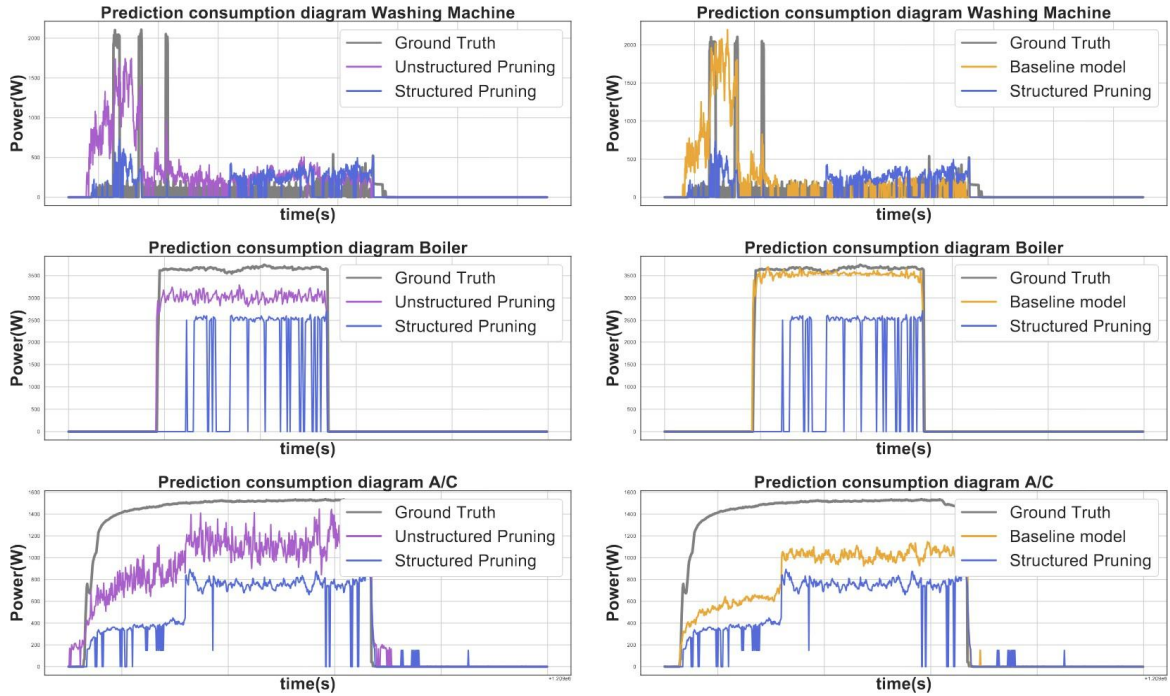

Figure 7.3 Prediction consumption using the optimized structured pruning vs. the unstructured pruning scheme and the optimized structured pruning vs. the baseline model. The pruning thresholds were set equal to the $\hat{s}$ as determined by Eq. (5.4).

An analysis of the compression metrics reveals significant differences between the pruning methods. Although unstructured pruning achieves a better trade-off between pruning percentage and performance, this advantage does not translate into improved MFLOPs, showing no improvement (0%) for all tested appliances. This is due to the inherent execution method of unstructured pruning, where parameters are replaced with zeroes, maintaining the same model dimensions. In contrast, the optimized structured pruning, although not delivering as robust performance, significantly reduces the MFLOPs by removing entire units and filters from the model. This method directly enhances the efficiency of the model's inference phase, resulting in reductions of 39.8%, 48.85%, and 11.75% in MFLOPs for the washing machine, boiler, and air conditioner, respectively.

These findings are particularly relevant for the real-world deployment of edge NILM (Non-Intrusive Load Monitoring) solutions. Optimized structured pruning's ability to reduce the computational load and improve inference efficiency makes it a more practical and deployable option for edge devices, where resource constraints are a critical factor. By lowering the MFLOPs, optimized structured pruning enables more efficient and effective NILM solutions on edge devices, facilitating wider adoption in real-world applications.

Table 7.1 Comparative evaluation results - Disaggregation performance with respect to compression threshold

| Appliance | Approach | Compression metrics | | | Performance metrics | | |
|---|---|---|---|---|---|---|---|
| | | Pruning (%) $s = \hat{s}$ | No Params | MFLOPs | F1 | MAE | MRE |
| Washer | Baseline | 0 | 22147640 | 430.26 | 0.81 | 5.79 | 0.026 |
| | Unstructured Pruning | 75 | 5538140 | 430.26 | 0.78 | 8.42 | 0.031 |
| | Opt. Structured Pruning | 50 | 11085760 | 259.01 | 0.58 | 6.05 | 0.032 |
| Boiler | Baseline | 0 | 22147640 | 430.26 | 0.88 | 17.64 | 0.081 |
| | Unstructured Pruning | 75 | 5538140 | 430.26 | 0.79 | 28.92 | 0.082 |
| | Opt. Structured Pruning | 60 | 8868520 | 220.12 | 0.75 | 37.71 | 0.086 |
| A/C | Baseline | 0 | 22147640 | 430.26 | 0.71 | 50.19 | 0.212 |
| | Unstructured Pruning | 75 | 5538140 | 430.26 | 0.66 | 48.79 | 0.228 |
| | Opt. Structured Pruning | 15 | 18827590 | 379.66 | 0.51 | 64.08 | 0.259 |

## 7.4 Conclusion

This work addresses the critical challenges posed by the resource-intensive nature of deep learning models in NILM, as well as their context-aware characteristics. Unlike existing studies, this research benchmarks devices such as air conditioners and boilers from the Plegma dataset, which are prevalent in the Mediterranean region. It compares the performance between the baseline CNN model, unstructured magnitude-based pruning, and optimized structured pruning. Experimental results demonstrate that all tested methods provided satisfactory results for the selected appliances, offering valuable insights into the applicability of NILM in this geographical area. Furthermore, the proposed structured pruning technique, in contrast to the unstructured approach, presents a viable solution for edge-computed NILM by significantly reducing the MFLOPs of the model by up to 48.85%. Unlike unstructured pruning, which does not achieve a reduction in MFLOPs, structured pruning decreases the computational load, thereby enhancing the practicality of NILM systems for real-world deployment on resource-constrained edge devices. These findings underscore the potential of edge NILM towards flexibility and energy transition in the Mediterranean region, facilitating the broader adoption and implementation of NILM solutions in real-world scenarios.

## Chapter 8

# Edge-Optimized Non-Intrusive Load Monitoring Using Dependency Graph-Based Structural Pruning

## 8.1 Introduction

As the global community strives to achieve net-zero carbon emissions by 2050, transforming traditional power grids into intelligent, adaptive smart grids has become essential [133]. According to the International Energy Agency, the energy sector accounts for three-quarters of total greenhouse gas emissions, with buildings alone responsible for approximately 37% of global energy-related $CO_2$ emissions [7]. Central to this objective is enhancing energy efficiency and optimizing energy consumption within buildings, which requires innovative solutions for monitoring and managing energy use. A crucial component of this transformation is Non-Intrusive Load Monitoring (NILM), a technology that disaggregates total energy consumption into individual appliance-level data without requiring intrusive hardware on the building side. By providing detailed insights into energy usage patterns, NILM supports intelligent demand-side management, enhances energy efficiency, and encourages households to adopt more sustainable energy habits. Indicatively, NILM can achieve significant energy savings of around 15%, making it a key enabler in the transition toward a more flexible and resilient energy ecosystem [201].

With advancements in deep learning, the state of the art in NILM has evolved beyond traditional signal processing methods [194] to approaches leveraging Artificial Neural Networks (ANNs). These deep learning-based techniques offer significant improvements, not only in detecting appliance states (on/off) but also in accurately estimating their power consumption patterns [90]. Cutting-edge NILM models utilize architectures such as Long Short-Term Memory (LSTM) [89], Gated Recurrent Units (GRU) [188], Convolutional Neural Networks (CNNs) [191], Graph Convolutional Networks (GCNs) [14], and Transformer-based architectures [189]. These methods enhance feature extraction, temporal modeling, and contextual understanding, leading to more precise load disaggregation and improved generalization across diverse datasets with different contextual characteristics.

Although deep learning approaches have been recognized for their performance in the domain of NILM, they are inherently limited by the requirement for a centralized data processing scheme due to the high computational demands of DNN-based NILM models [160]. This constraint hinders the real-world adoption of such solutions, as the necessary infrastructure increases costs and raises privacy concerns associated with transferring sensitive data to external sources.

To address these challenges, the scientific community has increasingly focused on simplifying DNN-based NILM approaches to enable their deployment on edge devices, making energy disaggregation more accessible for consumers [13]. This shift from centralized data processing to user-side energy disaggregation has led to the development of various edge-NILM solutions. The primary objective of these solutions is to optimize and compress model architectures, ensuring they operate efficiently with limited computational resources while maintaining performance.

The research landscape of edge-computed NILM is diverse, encompassing multiple approaches, including deep learning model architecture compression [16, 105, 19, 2], feature extraction techniques [163, 185], and federated learning [192]. One widely explored technique in NILM is pruning—a deep learning method that enhances model efficiency by eliminating less significant values in trained weight tensors. In [19], various pruning techniques, including magnitude-based, relative threshold, and entropy-based pruning, are investigated and applied to the NILM CNN sequence-to-point (seq2point) model proposed in [191]. These methods are evaluated using the kettle and dishwasher appliances from the REFIT dataset. Additionally, in [13, 15], a pretraining pruning strategy inspired by the Lottery Ticket Hypothesis is proposed, aiming to remove parameters at the initialization phase of the model reducing computational complexity during both the training and inference phases. Finally in [160], a performance-aware NILM compression scheme is introduced, incorporating unstructured pruning and a combination of unstructured pruning with quantization, which is tested on the UK-DALE dataset.

Existing pruning techniques applied to NILM present several limitations. Unstructured pruning relies on specialized AI accelerators or dedicated software for deployment, making it less practical for widespread adoption. On the other hand, although conventional structured pruning provides a more straightforward deployment, it can disrupt network integrity and lead to significant performance degradation since it removes entire units of the dnn's layers. Mitigating this challenge requires case-by-case dependency analysis, which severely hinders scalability.

To address these issues, the proposed Dependency Graph-based (DGB) pruning approach [57] is adopted to explicitly model the interdependencies between the pruned and its corresponding paired layers. The core idea behind this approach is that structured pruning in one layer will trigger pruning in adjacent layers, in order to maintain consistency in the model architecture. However, this simultaneous pruning complicates the selection of unimportant parameters, as traditional importance criteria are designed for individual layers. To address this, the approach transitions from assigning importance at the single-layer level to a group-level assessment, enabling the learning of consistent sparsity patterns across groups. This ensures that deactivated groups can be safely removed with minimal performance degradation. Our approach was evaluated using the newly established Plegma dataset on three different appliance types—a water boiler, a washing machine, and an air conditioner (A/C)—and was compared against both unstructured pruning and conventional structured pruning. The key contributions of this work are summarized below:

- **Novel Dependency Graph-Based Structured Pruning for NILM:** Proposes a structured pruning technique that leverages dependency graphs to model interdependencies between layers, ensuring structural coherence. By pruning interconnected layers simultaneously, this method mitigates the risk of network-breaking disruptions and maintains model integrity.

- **Optimized Trade-Off Between Pruning Efficiency and Performance degradation:** Introduces a transition from traditional single-layer importance assessment to a group-level sparsity pattern learning approach. This enhancement ensures a more balanced trade-off between computational efficiency and model accuracy, facilitating practical deployment on resource-constrained edge devices.

- **Comprehensive Evaluation on Real-World NILM Data:** Validates the effectiveness of the proposed pruning framework using the newly established Plegma dataset. Through comparative benchmarking against unstructured and conventional structured pruning methods, the study demonstrates the advantages of the proposed approach in improving NILM model efficiency while preserving performance.

## 8.2 Background on Deep Learning Model Pruning for NILM

### 8.2.1 NILM Problem Formulation

NILM is defined as the estimation of the individual appliance consumption levels by relying solely on their aggregated active power consumption. Formally, considering a fixed time-window $t = 1, \ldots, T$, a number of fixed appliances $m = 1, \ldots, M$ along with their consumptions as $y_m(t)$, NILM is defined as

$$x(t) = \sum_{m=1}^{M} y_m(t) + \varepsilon(t), \quad (8.1)$$

where $x(t)$ is the aggregated signal and $\varepsilon(t)$ is undesired noise originating from instrumentation measurements and/or appliances not contributing to the data collection process [79]. Solving eq. (8.1) directly through linear algebra requires matrix inversion [?], yet the process itself can make the matrix corresponding to the appliances power consumption patterns a) ill-conditioned due to correlation of the appliances values and b) under-determined due to unavailability of power measurements corresponding to the appliances.

To overcome the above challenges, DL models are used as their are proven effective in function approximation [107]. For the NILM case and considering a single appliance, a DL model approximates eq. (8.1) as $\hat{y}_m(t) = F(x(t);W)$, where $W \in \mathbb{R}^{L \times K \times N}$ are the models learnable parameters mapping for each layer $l$ a $k$-th dimensional input to an $n$-th dimensional output. According to the DL model architecture used, matrix $W$ can, in fact, be large making its real-world deployment in resource-constrained devices prohibitive. To this end, pruning compression methods are used to reduce their complexity so as to make them compact.

### 8.2.2 Unstructured Magnitude-Based Pruning

Unstructured pruning is a popular approach widely used within the context of NILM [2, 13]. Assuming that the individual weights values $w_{k,n}$ represent connections from $k$ to $n$, unstructured pruning sets their values to zero based on a sparsity ratio $s$; a selective parameter denoting the proportion of pruned parameters over the total ones. This process is achieved by making $L_1$ norm of the weighs being smaller to the sparsity ratio, i.e. $||w_{k,n}||_1 < s$. At first glance, the resulting pruned matrix, $W'$, is expected to be greatly reduced in terms of parameter size given the weights zeroing, yet this is not the case. Rather than being eliminated completely from the matrix, weights values are replaced with zeros and thus to fully exploit the compression benefits, hardware-specific optimizing tools are required tailored to process sparse matrix operations [179]. This further increases the complexity of deployable models making unstructured pruning an attractive, but challenging solution to be applied in real-world scenarios.

### 8.2.3 Structured Magnitude-Based Pruning

Structured pruning operates in a higher level compared to the unstructured on by removing units such as whole filters, neurons or channels, rather than zeroing their individual weight values by adopting the $L_1$ norm approach. This results in both reduced model parameters size and computational resources given that floating points operations are removed completely, thus making structured pruning a more efficient and straightforward to realize hardware-wise compared to its unstructured counterpart [8]. However, selecting the sparsity ratio should be made with caution; arbitrary setting its value to a large percentage can aggressively reduce the models trainable parameters at the cost of large drops in accuracy attributed to the elimination of the computational units underlying inter-dependencies such as residual connections [57].

## 8.3 Dependency Graphs-Based Structured Pruning for NILM

Dependency Graph (DG) is a grouping technique applied to structured pruning to generalize the unit removal process by transitioning it from architecture-specific to architecture-agnostic. Instead of manually selecting and pruning structurally grouped parameters, a varying process according to the DL architecture used, DG generalizes the procedure by modeling the interdependencies between paired layers before removing them [57]. This ensures that computational blocks performing important operations within the network's architecture such as feature extraction, learning stabilization and others, remain intact rather than being indiscriminately removed by constraint-unaware thresholding in structured pruning [? ].

The DG structural pruning is a sequential three-step process, involving 1) *dependency modeling*, 2) *grouping*, and 3) *group-level pruning*, with each one explained in detail below:

**Dependency modeling** is the initial step serving as the foundation of the DG structural pruning process. The function $F(x(t);W)$ describing the NN's architecture is decomposed to a smaller number of sub-functions $f_1, \ldots, f_L$, each one describing operations, both parameter (e.g., convolution) and non-parameter (e.g. residual connection) based, existing in each layer. To model the dependencies, each sub-function $f_l$ is then assigned to an input-output as $f_l = (f_l^-, f_l^+)$ to identify the *intra-layer* dependencies and the *inter-layer* dependencies. The former, corresponds to the dependencies existing between the input $f_l^-$ and output $f_l^+$ of the same sub-function, defined as $f_l^- \Leftrightarrow f_l^+$ and denoting that they share an identical pruning function $g(\cdot)$ as $g(f_l^-) = g(f_l^+)$. The latter, corresponds to the dependencies existing between the input $f_i^-$ and output $f_j^+$ of two different sub-functions, defined as $f_i^- \Leftrightarrow f_j^+$ and denoting that there is a consistent connection between them as $f_i^- \leftrightarrow f_j^+$. Then, their modeling is done by considering both dependencies for all layers, i.e. $i, j \in \{1, \ldots, L\}$, formulated as

$$d_{i,j}(f^-, f^+) = \mathbb{1}_{\{f_i^- \leftrightarrow f_j^+\}} \cup \mathbb{1}_{\{(i=j) \cap (g(f_l^-) = g(f_l^+))\}} \tag{8.2}$$

where $\mathbb{1}_{\{\cdot\}}$ is the indicator function. The elements of eq. (8.2) form a matrix $D = [d_{i,j}] \in [0, 1]^{2L \times 2L}$ corresponding to the adjacency matrix of the architecture's dependency graph $V(F, E)$.

**Grouping** builds upon the dependency graph by establishing the sub-groups to be pruned. It considers $V(F, E)$ and applies on it a graph traversal algorithm based on breadth-first search expansion, achieved by initializing an empty set $Q = \emptyset$ and iterating over a set of candidate nodes $i = 1, \ldots, 2|D|$. For each node $i$, this step initializes a new group $q = \{i\}$; b) finds not previously visited nodes $u = \{1, \ldots, 2|D|\} - q$; c) checks if a dependency exists, $q' = \{j \in v | \exists k \in q, d_{k,j} = 1\}$ and d) adds it as $q = q \cup q'$. Finally, the groups are updated as $Q = Q \cup \{q\}$.

**Group-level pruning** is the final step of the DG-based structured pruning responsible for executing the pruning part. Assuming that there are $C$ prunable dimensions with indexing $W_C$ the following regularization term for sparse training is applied

$$R(g,c) = \sum_{c=1}^{C} \sum_{w \in q} \gamma_c \|W_c\|_2^2 \tag{8.3}$$

where $\gamma_k = 2^{\alpha(I_q^{max} - I_{q,c})/(I_q^{max} - I_q^{min})}$ is a parameter controlling the shrinkage strength applied to the parameters, while $I_q$ is the $L_2$-norm of a weights group $q = \{w_1, \ldots, w_{|q|}\}$. Note that here, $\alpha$ is set to 4. Finally, to identify and remove unnecessary parameters, a relative score considering the max $P$

parameters is used as

$$\hat{I}_{q,c} = \frac{PI_{q,c}}{\sum \max{}_P(I_q)}.  \quad (8.4)$$

The DG-based process described above is a major step towards enhancing the structural pruning used in DL models within the context of NILM. This is mainly attributed to the time-series based inherent nature of NILM necessitating the employment of DL models capable of processing such kind of data, with 1D CNNs favoring the application scenario. Despite these models having much lower computational cost compared to their 2D counterparts as they rely on fewer feature extraction layers, they are in fact more susceptible to conventional structural pruning due to it removing units without taking into account their hierarchical importance nor their subsequent connections and interdependencies. As a result, arbitrary and aggressive selection of the sparsity ratio used to control the percentage of the weights to be pruned reduces the classification performance capabilities of the 1D CNNs dramatically thus hindering its applicability. On the other hand, the DG-enhanced structural pruning identifies and considers the layers importance and their interdependencies before removing them allowing for higher sparsity ratios selection with increased computational performance thereby advancing its adoption.

## 8.4 Experimental Results

### 8.4.1 Experimental Setup

To confirm the efficacy of the proposed compression scheme, we utilized the recently established Plegma dataset [9]. It consists of aggregate and appliance-level consumption measurements from 13 different households in Greece. The dataset was built at a sample rate of 0.1 Hz or a measurement every 10 seconds to be similar to the specified resolution of SMETS2 HAN [167] and to ensure the applicability in the real world and is among the first to represent Mediterranean regions, capturing appliances that are not commonly recorded, such as water boilers and air conditioning systems. These devices account for a significant portion of total energy consumption in these areas and offer substantial flexibility potential, making them strong candidates for energy disaggregation research. For this study, the selected models were trained and validated using data from houses 1–13, excluding house 2; it was reserved as completely unseen data to simulate a real-world scenario.

To evaluate and test the proposed DG-based structured pruning approach, we employ a sequence-to-sequence CNN architecture inspired by [191], which has been widely used in various edge NILM approaches [105, 13, 16, 160], enhancing comparability with existing works. The proposed model consists of five one-dimensional convolutional layers with rectified linear unit (ReLU) activation functions, followed by two fully connected layers with ReLU and sigmoid activations, respectively. The baseline model comprises 22.15 million parameters and has a total size of 84.5 MB.

### 8.4.2 Performance and Compression Metrics

To evaluate the model's disaggregation performance, the Mean Absolute Error (MAE) was used for regression tasks, measuring the difference between predicted and actual appliance consumption as

$$\text{MAE}(\hat{y}_m, y_m) = \frac{1}{T}\sum_{t=1}^{T}|\hat{y}_m(t) - y_m(t)| \quad (8.5)$$

where $\hat{y}$ and $\mathbf{y}$ denote the approximated and the actual power consumption of appliance $m$, while $T$ denotes the total time window.

To assess the model's ability to detect appliance states (ON/OFF), the accuracy and the F1-score metrics are employed. Accuracy measures the proportion of correctly classified states as T where $TP$ and $TN$ represent correctly predicted ON and OFF states, respectively, while $FP$ and $FN$ denote misclassified instances. The F1-score provides a balanced evaluation by considering both precision and recall as

$$\text{F1-score} = 2 \times \frac{\text{Precision} \times \text{Recall}}{\text{Precision} + \text{Recall}}, \quad (8.6)$$

where

$$\text{Precision} = \frac{TP}{TP+FP}, \quad \text{Recall} = \frac{TP}{TP+FN} \quad (8.7)$$

From (8.7), precision quantifies the proportion of correctly predicted ON states among all ON predictions, while recall measures the proportion of correctly identified ON states among all actual ON instances. Note that accuracy on itself can be misleading when imbalanced data or data patterns are exhibited so F1-score is considered as a more representative metric within the context of NILM.

The computational complexity of the model was assessed based on the number of trainable parameters, the model's storage size (MB), and the Multiply-Accumulate Operations (MACs). To further evaluate the model's its computational efficiency, that is before and after its pruning process, the following metric is considered

$$\text{Computational Efficiency} = \frac{\text{MACs}_{(\text{baseline})}}{\text{MACs}_{(\text{pruned})}} \quad (8.8)$$

To determine the optimal pruning threshold, denoted as $P_{\text{opt}}$, for each model, we employ a two-step procedure [13]. First, we select pruning thresholds percentages with increasing values as $P_i$ where $i \in \{5, 10, \ldots, 90, 95\}$ and compute the Euclidean distance for each point as $d(P_i)$ as

$$d(F1, P_i) = \sqrt{(100 - F_1)^2 + (100 - P_i)^2} \quad (8.9)$$

Note we assume that the "theoretical ideal point" is the one in which the F1-score of the pruned model is perfect, i.e. 1 (corresponding to 100%), while its parameters are minimized, i.e. the pruned threshold is 100%. Second, considering (8.9), we select the optimal pruning threshold as

$$P_{opt} = \arg \min_{i \in (5, 95)} d(F_1, P_i). \tag{8.10}$$

This formulation ensures that both model performance degradation and parameter reduction are taken into account, facilitating an optimal trade-off between accuracy and efficiency.

### 8.4.3 Evaluation and Findings

A comparative analysis between the unstructured magnitude pruning, structured magnitude pruning and the proposed depedency graph-based structured pruning is shown in Fig. 8.1. The diagram analyzes these pruning methods across thresholds ranging from 5% to 95% in 5% increments, assessing their impact on both model performance and size. The F1 score is used to assess classification performance, MAE to evaluate regression accuracy, and model size in MB to illustrate the effect of pruning on model compression.

The 'pruning threshold vs performance' diagrams indicate that the proposed dependency graph-based structured pruning maintains performance comparable to unstructured magnitude pruning, even for high pruning percentages, while consistently outperforming conventional structured pruning across all devices.

The "Pruning Threshold vs. Model Size" diagrams highlight a fundamental distinction between unstructured and structured pruning approaches, reinforcing their differing mechanisms as outlined in the Methodology section. Specifically, structured pruning and DGP-based structured pruning reduce model size by physically removing entire units from the network, whereas unstructured pruning retains the original model size since it uses a binary masking technique that sets pruned weights to zero without altering the total number of parameters. For the boiler and A/C appliances, both DGB structured pruning and magnitude-based unstructured pruning identify the optimal pruning threshold at 90%, while for the washing machine, these thresholds are 95% and 85%, respectively, achieving substantial parameter reduction with minimal performance loss. In contrast, conventional structured pruning struggles to maintain a favorable balance between pruning threshold and performance degradation, reaching significantly lower optimal pruning thresholds of 25%, 50%, and 60% for the A/C, washing machine, and boiler, respectively.

Referring to the performance metrics presented in Table 8.1, we can validate that the unstructured and DGB structured pruned models effectively preserve both their classification and regression capabilities compared to the baseline model. In contrast, conventionally structured pruned models experience a greater impact on performance. For the air conditioner (A/C) appliance, the F1 score decreases slightly from 0.72 to 0.67 with unstructured pruning and to 0.71 with the proposed DGB structured pruning. In contrast, conventional structured pruning shows a more significant drop to 0.58, indicating a greater impact on classification performance. A similar pattern is observed in the regression performance where both unstructured and DGB structured pruning result in minimal change of less than 1%, whereas structured pruning leads to a substantial increase in MAE from 50.20 to 88.93. For the washing machine appliance, the classification and regression performance of the model remain largely unaffected, with only a minor decline in the F1 score from 0.81 to 0.79 with unstructured pruning and to 0.72 with DGB structured pruning. The MAE increases slightly from 6.18 to 7.12 and 8.11, respectively. However, structured pruning once again has a more pronounced impact on performance, with the F1 score dropping to 0.69 and the MAE increasing sharply to 22.15. Finally, for the boiler appliance, which exhibits a relatively simple consumption pattern and lower variability across different boiler types, the impact of pruning on classification performance is minimal. The F1 score experiences only a slight decrease from 0.88 to 0.85 with unstructured pruning and to 0.84 with the proposed DGB structured pruning. In contrast, the MAE shows a more pronounced increase, rising from 17.64 to 27.98 for unstructured pruning and to 25.12 for DGB structured pruning. Conventional structured pruning follows the same trend observed in other devices, with the F1 score dropping to 0.77 and the MAE rising to 38, underscoring its more significant impact on performance degradation.

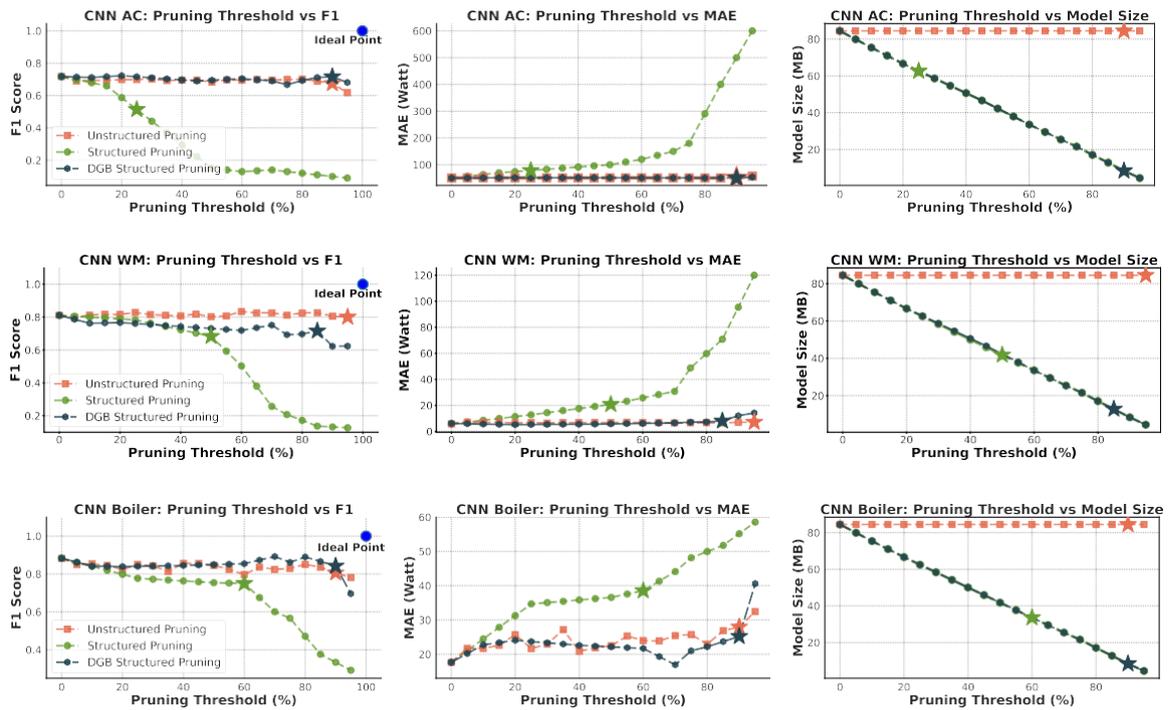

Figure 8.1 Diagrams compare computational complexity for different pruning thresholds (5-95%) to performance degradation and model size (F1, MAE, MB). Stars in the Pruning Threshold-F1 plots mark the optimal pruning threshold for DGB, convetional structured and unstructured pruning, as determined by Eq. (8.10) while blue dot represents the 'ideal' point.

Table 8.1 Comparative Evaluation Results - Performance with Respect To Pruning Ratio

| Appliance | Approach | Pruning (%) $P_{thr} = P_{opt}$ | Compression metrics | | | Performance metrics | | |
|---|---|---|---|---|---|---|---|---|
| | | | Number of params (M)($\times 10^6$) | Model size on disk (MB) | MACs (G)($\times 10^9$) | F1 Score | MAE | Accuracy |
| AC | Baseline Model | 0 (%) | 22.15 | 84.5 | 39.27 | 0.72 | 50.20 | 0.96 |
| | Unstructured Pruning | 90 (%) | 22.15 | 84.5 | 39.27 | 0.67 | 50.50 | 0.95 |
| | Structured Pruning | 25 (%) | 16.61 | 62.5 | 29.56 | 0.58 | 88.93 | 0.94 |
| | **DGB Structured Pruning** | **90 (%)** | **2.20** | **8.4** | **3.93** | **0.71** | **50.30** | **0.96** |
| Washing Machine | Baseline Model | 0 (%) | 22.15 | 84.5 | 39.27 | 0.81 | 6.18 | 0.98 |
| | Unstructured Pruning | 95 (%) | 22.15 | 84.5 | 39.27 | 0.79 | 7.12 | 0.97 |
| | Structured Pruning | 50 (%) | 11.53 | 41.8 | 19.69 | 0.69 | 22.15 | 0.96 |
| | **DGB Structured Pruning** | **85 (%)** | **3.36** | **12.8** | **6.06** | **0.72** | **8.11** | **0.97** |
| Boiler | Baseline Model | 0 (%) | 22.15 | 84.5 | 39.27 | 0.88 | 17.64 | 0.98 |
| | Unstructured Pruning | 90 (%) | 22.15 | 84.5 | 39.27 | 0.85 | 27.98 | 0.98 |
| | Structured Pruning | 60 (%) | 8.72 | 33.5 | 15.7 | 0.77 | 38 | 0.97 |
| | **DGB Structured Pruning** | **90 (%)** | **2.20** | **8.4** | **3.88** | **0.84** | **25.21** | **0.98** |

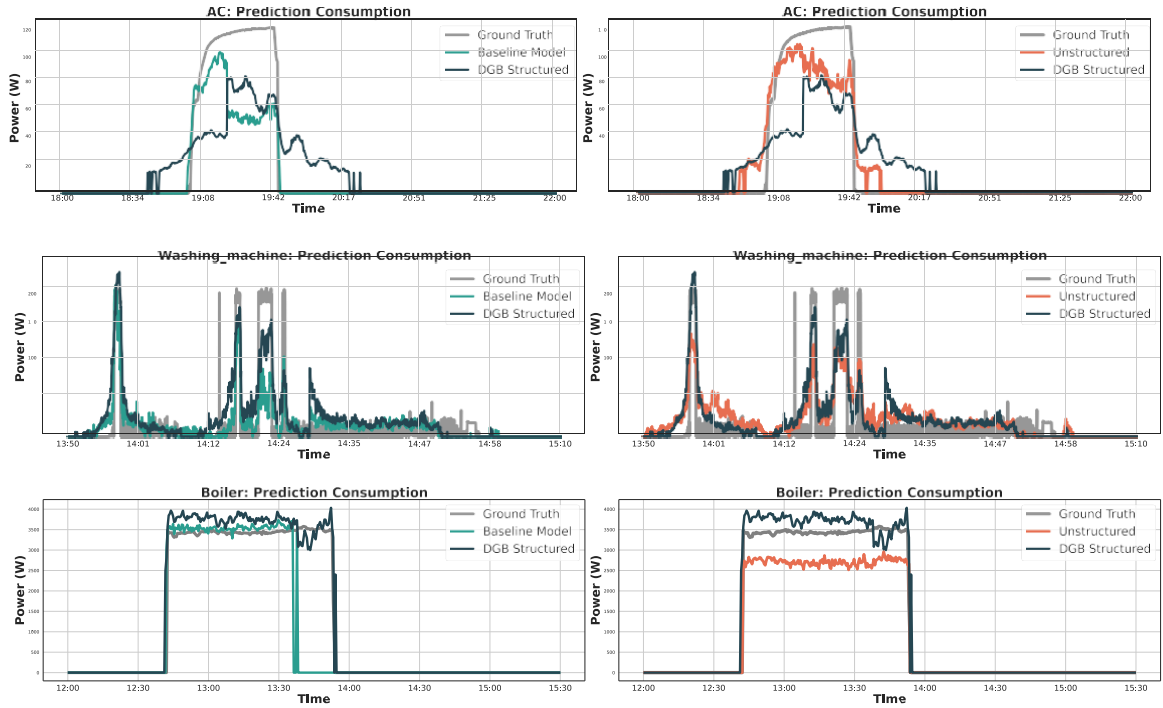

Figure 8.2 Comparison of predicted consumption diagrams using the proposed Depedency Graph-based Structured Pruning against Unstructured Pruning and the Baseline Model, with pruning thresholds set to $P_{\text{thr}} = P_{\text{opt}}$.

The results can be visually validated through the prediction consumption diagrams in Fig. 8.2, which illustrate the predicted appliance consumption for the baseline model, unstructured pruned model, and DGB structured pruned model, alongside the ground truth consumption data. These pruning approaches were included as they achieve an optimal balance between compression level and performance degradation. According to the diagrams, the proposed DGB structured pruning achieves a performance comparable to that of the baseline and unstructured pruning approaches. It successfully infers the activations of all three tested devices and accurately classifies their on-off states.

Examining the compression metrics highlights significant differences between the three pruning approaches. While both DGB structured pruning and unstructured pruning maintain a favorable trade-off between pruning threshold and classification-regression performance degradation, this does not necessarily translate into improved computational efficiency for both methods. As previously discussed, unstructured pruning retains the full model size since it relies on a binary masking technique rather than physically removing parameters. In contrast, structured pruning and the proposed DGB structured pruning physically eliminate entire units from the model architecture, leading to a notable reduction in memory footprint.

Table 8.2 Improvement in compression metrics

| Appliance | Approach | Improvement in Compression Metrics | | |
|---|---|---|---|---|
| | | Reduction Number of params (%) | Reduction Model size on disk (%) | Computational Efficiency ($\times \frac{MACs_{base}}{MACs_{pruned}}$) |
| AC | Unstructured Pruning | 0.00% | 0.00% | 1.00$\times$ |
| | Structured Pruning | 25.01% | 26.04% | 1.33$\times$ |
| | DGB Structured Pruning | 90.07% | 90.04% | 9.99$\times$ |
| Washing Machine | Unstructured Pruning | 0.00% | 0.00% | 1.00$\times$ |
| | Structured Pruning | 47.95% | 50.53% | 2.00$\times$ |
| | DGB Structured Pruning | 84.83% | 84.85% | 6.48$\times$ |
| Boiler | Unstructured Pruning | 0.00% | 0.00% | 1.00$\times$ |
| | Structured Pruning | 60.63 | 60.36 | 2.50$\times$ |
| | DGB Structured Pruning | 90.07 | 90.06 | 10.12$\times$ |

Beyond model size, this reduction extends to other key compression metrics, including trainable parameters and multiply-accumulate operations (MACs), which remain unchanged for unstructured pruning but are significantly reduced through structured methods. Conventional structured pruning achieves moderate improvements, reducing the number of parameters by 25% to 60%, model size by 26% to 60%, and computational efficiency by up to 2.5$\times$, depending on the appliance. However, DGB structured pruning offers far greater efficiency gains, exceeding 90% parameter reduction, reducing model size by 85–90%, and significantly enhancing theoretical computational efficiency—with a 10$\times$ speed-up for the air conditioner and boiler, and 6.5$\times$ for the washing machine (as shown in Table 8.2).

These results highlight that while structured pruning enhances computational efficiency, DGB structured pruning emerges as the most effective approach, striking an optimal balance between compression and performance retention. Its ability to significantly reduce model size and computational cost while maintaining accuracy makes it a highly viable solution for deployment in resource-constrained environments.

## 8.5 Conclusion

In this study, we have proposed a dependency graph-based structured pruning method for NILM DNN-based approaches, aimed at reducing both computational complexity and storage requirements. Unlike conventional unstructured magnitude-based pruning, which applies binary masking without reducing model size, the proposed structured pruning physically removes entire units from the network, significantly decreasing model size and computational demands. Furthermore, when comparing DGB structured pruning with conventional structured pruning, we observe that our method achieves a far superior balance between compression level and performance degradation.

Experimental results show that our approach maintains performance comparable to unstructured pruning across various pruning thresholds while delivering substantial gains in storage efficiency and computational cost. Specifically, DGB structured pruning reduces the number of multiply-accumulate operations by up to 90% and improves inference efficiency by up to 10×, making it a viable solution for deployment on resource-constrained edge devices.

These findings highlight DGB structured pruning as a practical model compression strategy, facilitating deep learning NILM models deployment in real-world applications with limited computational resources. Future work will explore integrating our approach with hardware-aware optimizations to further enhance model efficiency and scalability.

# Part V

# Conclusions & Future Work

# Chapter 9

# Conclusion of the Thesis

## 9.1 Summary

*This thesis was concerned with the development of novel methodologies and frameworks aimed at advancing Non-Intrusive Load Monitoring (NILM) and energy behavior analysis in edge-computed environments. It specifically addresses key challenges related to data acquisition and dataset diversity, computational efficiency and deployment on resource-constrained devices, as well as the practical applicability of these approaches in real-world scenarios.*

- Part I established the theoretical foundation and contextual framework guiding this research towards advancing energy monitoring and Non-Intrusive Load Monitoring (NILM). It outlined critical research gaps, including challenges in data acquisition, the limited diversity of available datasets, and the context-aware nature of NILM models. A particular emphasis was placed on the lack of datasets capturing Mediterranean-specific energy consumption patterns, highlighting the need to address this regional gap. This section also identified practical challenges in deploying NILM solutions, particularly the reliance on resource-intensive centralized systems. It underscored the necessity of developing edge-computed NILM approaches to enhance scalability, privacy, and real-world applicability. Additionally, Part I articulated the main research questions driving this work, shaped by the challenges and motivations identified. These questions aimed to bridge existing gaps, proposing innovative methodologies to address data limitations, computational constraints, and region-specific applicability. This part also introduced the primary contributions of this thesis, setting the stage for the subsequent chapters where these contributions are detailed and validated.

- Part II concentrated on bridging the dataset gaps in NILM research by developing an interoperable data collection framework utilizing advanced IoT technologies and communication protocols. The focus was on ensuring robust, diverse, and region-specific data collection to support the development and deployment of NILM solutions. In particular, Chapter 2 intro-duced an Internet of Things (IoT)-based framework designed for household energy monitoring and analysis. This framework, built on the Z-wave communication protocol, emphasizes interoperability and cost-effectiveness while addressing the needs of smart-home users. Its robust architecture ensures real-time data collection and seamless integration with diverse devices. Chapter 3 presented the Plegma dataset, created using the developed energy monitoring framework. This dataset addresses the critical data gap for NILM in the Mediterranean region. Collected over a year from 13 households in Greece, the Plegma dataset includes whole-house energy loads, appliance-level consumption data at 10-second intervals, and environmental metrics like temperature and humidity. Additionally, it incorporates metadata on building characteristics, demographics, and user practices. By capturing Mediterranean-specific consumption patterns—particularly for underrepresented appliances such as air conditioners and electric water boilers with high flexibility potential—the Plegma dataset is a vital resource for advancing NILM research and applications.

- Part III addressed the second research question by exploring advanced DNN-based NILM techniques and innovative compression strategies to improve computational efficiency and enable practical, real-world deployment on resource-constrained devices. Chapter 4 introduced a novel seq2seq approach utilizing Graph Neural Networks (GNNs) as an encoder and a Transformer-based decoder, marking the first attempt to model NILM as a graph problem. This method captured time-invariant dependencies and appliance-specific operational states, achieving competitive or superior performance for multi-state appliances like washing machines. While currently limited to certain device types, this graph-based perspective offers significant potential for future NILM research. Chapter 5 presented a pre-training model compression strategy, employing iterative magnitude pruning to create optimized, lightweight DNNs. This approach achieved up to 95% reduction in training computational costs while maintaining comparable performance to full models, demonstrating suitability for edge IoT deployment. Validation on the UK-DALE dataset highlighted the effectiveness of this technique in balancing efficiency and performance. Chapter 6 introduced OPT-NILM, a novel pruning strategy that optimizes NILM models before full training, significantly reducing computational complexity while maintaining or enhancing disaggregation accuracy. This approach employed a systematic metric for determining pruning thresholds and demonstrated real-world feasibility through deployment on a Raspberry Pi. OPT-NILM showcased the ability to run NILM applications on low-resource edge devices, broadening the accessibility and adoption of NILM technologies.

- Part IV investigated the application of compression techniques to NILM models, with a particular focus on Mediterranean-specific energy consumption scenarios. This section addressed the third research question by developing methodologies to optimize NILM models for efficient edge deployment while preserving accuracy and aligning with the unique energy patterns of the Mediterranean region. The findings highlighted the potential of these techniques to identify flexible loads, such as air conditioners and water boilers, which represent a significant portion of energy consumption in this area, showcasing their importance for advancing energy efficiency and grid flexibility. Chapter 7 introduced an optimized structured pruning methodology that combines unstructured pruning for determining sparsity ratios with structured pruning to remove entire network units. This dual-pruning approach preserves critical feature information, avoids arbitrary thresholds, and enhances model performance. The methodology, evaluated using the Plegma dataset, demonstrated a 48.85% reduction in computational costs (MFLOPs) while maintaining robust disaggregation performance. These results highlight the potential of edge-based NILM applications for promoting energy efficiency and real-world adoption in the Mediterranean region. Chapter 8 presented a Dependency Graph (DG) structural pruning approach to optimize Deep Learning models for edge-based Non-Intrusive Load Monitoring (NILM). By preserving inter-layer dependencies during pruning, DG ensures structural integrity and supports more aggressive compression compared to conventional methods. Applied to the Plegma dataset, the method achieved up to 90% model size reduction and 10× computational efficiency gains with minimal performance loss. These results demonstrate the potential of DG pruning for scalable and energy-efficient NILM in resource-constrained Mediterranean settings.

## 9.2  Innovation and Originality

The work presented in the previous chapters was conducted with the overarching aim of emphasizing the pivotal role of Non-Intrusive Load Monitoring (NILM) in advancing energy efficiency, leveraging cutting-edge machine learning techniques, and implementing state-of-the-art optimization and data collection frameworks. The primary contributions of this thesis are summarized as follows.

1. **Development of an interoperable and cost-effective data collection framework** (see Section 2.3). The primary goal of this approach is to improve dataset development, thereby expanding the applicability of AI-driven energy solutions, such as AI-enabled NILM.

2. **Development of the Plegma Dataset** (see Sections 3.1, 3.2, 3.3, 3.4). The primary objective of creating this dataset was to address the lack of NILM dataset coverage in the Mediterranean region, fostering the application of NILM and energy-related technologies in this area. By incorporating region-specific energy patterns and appliances, such as air conditioners and water boilers, along with environmental, sociodemographic, and building characteristics, the Plegma dataset significantly enhances its potential applications and usability for advancing NILM methods and energy analysis tailored to the unique dynamics of the Mediterranean region.

3. **Introduction of a seq2seq NILM Approach with Graph Neural Networks (GNNs)** (see Sections 4.3, 4.4). This method models NILM as a graph problem, using GNNs to encode aggregated signals based on their entropy level and a Transformer-based decoder for disaggre-gation. It achieves competitive performance for multi-state appliances, introducing a novel graph-based perspective to NILM.

4. **Introduction of a Pre-training Compression Strategy for DNN-Based NILM** (see Sections 5.2, 5.3). This approach employs iterative magnitude pruning applied prior to full model training, creating optimized and lightweight DNNs. It reduces training costs by up to 95% with minimal performance degradation. Validated on the UK-DALE dataset, this strategy facilitates efficient edge IoT deployment for NILM applications.

5. **Introduction of OPT-NILM Pruning Strategy for NILM Optimization** (see Sections 6.4, 6.5, 6.6). OPT-NILM optimizes NILM models by identifying efficient network structures before full training, reducing computational demands while maintaining performance. Its deployment scenario on a Raspberry Pi demonstrates the feasibility of running NILM applications on resource-constrained edge devices, enabling practical real-world implementations and expanding accessibility in the energy sector.

6. **Development of a Structured Pruning Methodology for NILM Deployment in the Mediterranean Region** (see Sections 7.2, 7.3). This methodology integrates structured and unstructured pruning to optimize NILM models for efficient edge deployment. Leveraging the Plegma dataset, which captures energy consumption patterns specific to the Mediterranean, it showcases the practical feasibility of deploying NILM solutions in the region, enabling energy-efficient applications tailored to local needs.

7. **Development of a Dependency Graph-Based Structural Pruning Methodology** for Edge-Deployable NILM Models in the Mediterranean Region (see Section 8.3). This methodology introduces Dependency Graph (DG) structural pruning, which preserves inter-layer dependencies while removing redundant parameters, ensuring architectural consistency and enabling more aggressive compression. Applied to the Plegma dataset—capturing Mediterranean appliance usage patterns—the approach achieves up to 90% model size reduction and a 10× gain in computational efficiency, with minimal performance degradation. These results highlight the method's effectiveness in delivering scalable, energy-efficient NILM solutions tailored to the constraints of edge devices and the needs of the region's energy transition.

## 9.3 Future Research Directions

This research has successfully bridged the gap between academic advancements and the real-world adoption of NILM technologies. The creation of the Plegma dataset has significantly expanded the applicability and research on NILM within the Mediterranean region, addressing the lack of region-specific datasets. Additionally, the development of DNN-based and edge-computed NILM approaches has paved the way for practical deployment, overcoming critical challenges such as privacy concerns, high storage costs, and the computational demands of centralized systems. By applying novel compression techniques tailored to the Mediterranean use case, utilizing the Plegma dataset, we demonstrated that edge-based NILM is not only feasible but also highly effective in this region. This has unlocked significant potential for building energy flexibility, particularly through the disaggregation of high-energy-consuming appliances like air conditioners and water boilers.However, the journey of research and innovation does not end here. As NILM technologies continue to evolve, there are numerous opportunities to overcome existing challenges and explore new approaches to further enhance their usability, performance, and real-world impact. Some promising future research directions to advance NILM adaptability and effectiveness in practical scenarios include the following:

**Enhancing Explainability of NILM Models** A key area for future work is the development of explainable NILM models [127, 21, 117]. Incorporating explainable AI techniques can provide users with clear insights into how these models function, fostering trust and driving adoption. Transparent and interpretable results will also enable more informed energy management decisions, empowering users to engage with their energy consumption behaviors more effectively.

**Developing Multimodal NILM Models** Another exciting direction involves the creation of multimodal NILM models that utilize additional features beyond aggregate energy consumption. The growing adoption of Smart Home technologies and IoT has made a wealth of data from buildings accessible, offering new opportunities to enhance disaggregation performance. For instance, the Plegma dataset includes valuable contextual information such as environmental conditions, sociodemographic data, occupant routines, and building characteristics. Integrating these diverse data sources into NILM models can significantly boost disaggregation accuracy, broaden their applicability, and provide more actionable insights, ultimately fostering greater user adoption.

**Exploring Adaptive NILM Systems Using Active Learning** Adaptive NILM systems represent another promising frontier, particularly through the application of active learning techniques [165, 110]. By allowing users to actively participate in the model training and fine-tuning process, these systems can iteratively improve their accuracy while fostering user engagement [123]. This interaction not only boosts the model's adaptability to changing usage patterns but also promotes user trust and ownership, further encouraging the widespread adoption of NILM technologies.

With these advancements, NILM technologies are set to play a pivotal role in addressing global energy efficiency challenges. By enabling precise energy monitoring, fostering sustainable consumption, and involving users in this process, NILM can drive the transition to a greener, more sustainable, and energy-resilient future.